\def\isarxiv{1} 

\ifdefined\isarxiv
\documentclass[11pt]{article}

\usepackage[numbers]{natbib}

\else

\documentclass[twoside]{article}
\usepackage[round]{natbib}
\usepackage[accepted]{aistats2024}
\fi

\usepackage{amsmath}
\usepackage{amsthm}
\usepackage{amssymb}
\usepackage{algorithm}
\usepackage{subfig}
\usepackage{algpseudocode}
\usepackage{graphicx}
\usepackage{grffile}
\usepackage{wrapfig,epsfig}
\usepackage{url}
\usepackage{xcolor}
\usepackage{epstopdf}

\usepackage{bbm}
\usepackage{dsfont}

\allowdisplaybreaks

\ifdefined\isarxiv

\usepackage{tikz}
\usepackage{hyperref}  
\hypersetup{colorlinks=true,citecolor=blue,linkcolor=blue} 
\usetikzlibrary{arrows}
\usepackage[margin=1in]{geometry}

\else

\usepackage{microtype}
\usepackage{hyperref}
\definecolor{mydarkblue}{rgb}{0,0.08,0.45}
\hypersetup{colorlinks=true, citecolor=mydarkblue,linkcolor=mydarkblue}

\fi

\graphicspath{{./figs/}}

\newtheorem{theorem}{Theorem}[section]
\newtheorem{lemma}[theorem]{Lemma}
\newtheorem{definition}[theorem]{Definition}

\newtheorem{fact}[theorem]{Fact}
\newtheorem{remark}[theorem]{Remark}

\newcommand{\wh}{\widehat}
\newcommand{\wt}{\widetilde}

\newcommand{\R}{\mathbb{R}}

\newcommand{\ose}{\mathrm{ose}}
\newcommand{\final}{\mathrm{final}}
\newcommand{\odd}{\mathrm{odd}}
\newcommand{\even}{\mathrm{even}}

\DeclareMathOperator*{\E}{{\mathbb{E}}}

\DeclareMathOperator*{\Z}{\mathbb{Z}}

\DeclareMathOperator{\OPT}{OPT}

\DeclareMathOperator{\poly}{poly}

\DeclareMathOperator{\nnz}{nnz}

\DeclareMathOperator{\rank}{rank}
\DeclareMathOperator{\diag}{diag}
\DeclareMathOperator{\dist}{dist}

\DeclareMathOperator{\tr}{tr}

\makeatletter
\newcommand*{\RN}[1]{\expandafter\@slowromancap\romannumeral #1@}
\makeatother

\usepackage{lineno}

\title{Solving Attention Kernel Regression Problem via Pre-conditioner}

\begin{document}

\ifdefined\isarxiv

\date{}

\title{Solving Attention Kernel Regression Problem via Pre-conditioner}
\author{
Zhao Song\thanks{\texttt{zsong@adobe.com}. Adobe Research.}
\and
Junze Yin\thanks{\texttt{junze@bu.edu}. Boston University.}
\and
Lichen Zhang\thanks{\texttt{lichenz@mit.edu}. MIT.}
}

\else

\twocolumn[
\aistatstitle{Solving Attention Kernel Regression Problem via Pre-conditioner}
\aistatsauthor{Zhao Song \And Junze Yin \And Lichen Zhang}
\aistatsaddress{Adobe Research \And Boston University \And Massachusetts Institute of Technology}
]

\fi

\ifdefined\isarxiv
\begin{titlepage}
  \maketitle
  \begin{abstract}

The attention mechanism is the key to large language models, and the attention matrix serves as an algorithmic and computational bottleneck for such a scheme. In this paper, we define two problems, motivated by designing fast algorithms for \emph{proxy} of attention matrix and solving regressions against them. Given an input matrix $A\in \mathbb{R}^{n\times d}$ and a response vector $b$, the first proxy, for $n\gg d$, uses powers of the matrix $A^\top A$, and we consider two types of regression problems: $\min_{x\in \mathbb{R}^d}\|(A^\top A)^jx-b\|_2$ and $\min_{x\in \mathbb{R}^d}\|A(A^\top A)^jx-b\|_2$ for any positive integer $j$. We design fast algorithms for these power regressions, based on sketching and preconditioning. The second proxy is applying exponential entrywise to the Gram matrix, denoted by $\exp(AA^\top)$ and solving the regression $\min_{x\in \mathbb{R}^n}\|\exp(AA^\top)x-b \|_2$. We call this problem the \emph{attention kernel regression} problem, as the matrix $\exp(AA^\top)$ could be viewed as a kernel function with respect to $A$. For this problem, we show that the sketching and preconditioning approach known for Gaussian kernels gives error $\epsilon\|b\|_2$ for every $b$, with a sketch dimension that is, up to logarithmic factors, $\epsilon^{-2}n$ times the condition number of the kernel matrix, which is at least $n/e$ when the rows of $A$ have norm at most $1$. For rows of norm at most $1$, we also give an algorithm whose sketch dimension, up to logarithmic factors, does not depend on $n$. It sketches the kernel plus a ridge term, and for a given $K\geq 1$ its error is at most $\epsilon\|b\|_2$ plus $(1+\sqrt{\epsilon})$ times the best residual among solutions of norm at most $(K/n)\|b\|_2$. We hope these efforts will provide an alternative perspective of studying efficient approximation of attention matrices.

  \end{abstract}
  \thispagestyle{empty}
\end{titlepage}

{
}
\newpage

\else

\begin{abstract}

The attention mechanism is the key to large language models, and the attention matrix serves as an algorithmic and computational bottleneck for such a scheme. In this paper, we define two problems, motivated by designing fast algorithms for \emph{proxy} of attention matrix and solving regressions against them. Given an input matrix $A\in \mathbb{R}^{n\times d}$ and a response vector $b$, the first proxy, for $n\gg d$, uses powers of the matrix $A^\top A$, and we consider two types of regression problems: $\min_{x\in \mathbb{R}^d}\|(A^\top A)^jx-b\|_2$ and $\min_{x\in \mathbb{R}^d}\|A(A^\top A)^jx-b\|_2$ for any positive integer $j$. We design fast algorithms for these power regressions, based on sketching and preconditioning. The second proxy is applying exponential entrywise to the Gram matrix, denoted by $\exp(AA^\top)$ and solving the regression $\min_{x\in \mathbb{R}^n}\|\exp(AA^\top)x-b \|_2$. We call this problem the \emph{attention kernel regression} problem, as the matrix $\exp(AA^\top)$ could be viewed as a kernel function with respect to $A$. For this problem, we show that the sketching and preconditioning approach known for Gaussian kernels gives error $\epsilon\|b\|_2$ for every $b$, with a sketch dimension that is, up to logarithmic factors, $\epsilon^{-2}n$ times the condition number of the kernel matrix, which is at least $n/e$ when the rows of $A$ have norm at most $1$. For rows of norm at most $1$, we also give an algorithm whose sketch dimension, up to logarithmic factors, does not depend on $n$. It sketches the kernel plus a ridge term, and for a given $K\geq 1$ its error is at most $\epsilon\|b\|_2$ plus $(1+\sqrt{\epsilon})$ times the best residual among solutions of norm at most $(K/n)\|b\|_2$. We hope these efforts will provide an alternative perspective of studying efficient approximation of attention matrices.

\end{abstract}

\fi

\section{Introduction}

Many numerical linear algebra tasks admit efficient, randomized solutions~\citep{wlrt08,kpst94,we06,afk+01}. These problems include the approximating leverage scores, least squares regression, and low rank approximation, and they have numerous applications in areas such as recommendation systems \citep{dkr02}, data mining \citep{afk+01}, web search \citep{afkm01,k99}, information retrieval \citep{ptrv98}, learning mixtures of distributions \citep{am05,ksv08}, and clustering \citep{dfk+04,m01}. The deployment of randomization and approximation enables these problems to be solved much more rapidly than their deterministic and exact counterpart.

In the modern big data era in which the size of the matrices grows rapidly, these randomized approaches become even more appealing as most of them could leverage the structure of the input matrix, and subsequently solve the problem in time nearly linear of the data. A prominent randomized paradigm for numerical linear algebra is the \emph{sketching} approach~\citep{s06,cw13}. Roughly speaking, given a tall and skinny matrix $A\in \R^{n\times d}$ with $n\gg d$, one draws a random matrix $S\in \R^{m\times n}$ from a structured family and computes the product $SA$. Based on the choice of $m$, the random matrix $S$ can either preserve the column norms of $A$~\citep{jl84} or the entire subspace spanned by $A$~\citep{s06}. Moreover, the structure of $S$ oftentimes enables the matrix product $SA$ to be computed very efficiently~\citep{ac06,cw13,ldfu13,nn13,kn14}. Such an approach has found many applications including linear regression~\citep{cw13}, low rank approximation~\citep{cw13,bdm14} and kernel approximation~\citep{anw14,akk+20,swyz21}.

In this paper, we study the efficient computation and approximation of \emph{attention matrices}. These matrices are fundamental objects of deep learning models utilized in a wide range of domains, including natural language processing \citep{jys+19}, computer vision \citep{gxl+22}, speech recognition \citep{cbs+15,wdcx21}, and robotics \citep{lew+22}. Attention mechanisms enable models to focus selectively on specific portions of input data and dynamically adjust weights for distinct features and context information. The attention matrix is a crucial component of attention mechanisms, capturing the relationships between input elements and the query vector. By computing the attention matrix, models can learn to attend to pertinent information and disregard irrelevant information, leading to improved performance and interpretability. Furthermore, the attention matrix provides valuable insights into how models make decisions and reason about input data, aiding in debugging and enhancing models. Therefore, understanding the attention matrix is crucial for comprehending the behavior and limitations of deep learning models and developing more potent attention mechanisms. Recent studies have shown that attention mechanisms can be applied to other problems beyond traditional deep learning, such as graph neural networks \citep{zwg+20,gqsw22}, reinforcement learning \citep{bc22}, and meta-learning \citep{smp21}. 

As indicated by~\cite{as23}, efficient approximation of attention matrices requires the structural condition that the entries are bounded given standard complexity theoretical assumptions. On the flip side, they also show these matrices can be approximated in $O(n^{1+o(1)})$ time if certain parameters are desirable. While this seems to close the algorithmic study of approximating attention matrices, it doesn't bar the study of \emph{proxy} for these objects, i.e., components of attention matrices that either admit different structural assumptions, bypassing bounded entries; or different type of matrices which serve as a replacement for the standard attention. We hence introduce two kinds of proxy: the first uses powers of the matrix $A^\top A$, and the second applies exponential function entrywise to a Gram matrix. We then consider solving regressions against these proxies.

\begin{definition}\label{def:3_A}
Given an input matrix $A \in \R^{n \times d}$, a response $b \in \R^n$, the goal is to solve
\begin{align*}
\min_{x \in \R^d} \| A A^\top A x - b \|_2^2
\end{align*}
\end{definition}

\begin{definition}\label{def:4_A}
Given an input matrix $A \in \R^{n \times d}$, a response $b \in \R^d$, the goal is to solve
\begin{align*}
\min_{x \in \R^d} \| A^\top A A^\top A x - b \|_2^2
\end{align*}
\end{definition}

As we will see later, by induction, algorithms for these problems naturally extend to higher even and odd powers:

\begin{align}\label{eq:informal_even}
    \min_{x \in \R^d} \| (A^\top A)^j x - b \|_2
\end{align}
and
\begin{align}\label{eq:informal_odd}
    \min_{x \in \R^d} \| A(A^\top A)^j x - b \|_2,
\end{align}
where $d$ and $j$ are arbitrary natural numbers. We study these regressions against the powers $(A^\top A)^j$ and $A(A^\top A)^j$ as problems of their own interest.

The next problem concerns applying entrywise exponential to a Gram matrix. As it could be interpreted as applying an exponential kernel function to the data $A$, we call it \emph{attention kernel regression} or \emph{exponential regression}.

\begin{definition}[Attention Kernel Regression (or Exponential Regression)]\label{def:exp}
Given an input matrix $A \in \R^{n \times d}$, a response $b \in \R^n$, the goal is to solve
\begin{align*}
    \min_{x \in \R^n} \| \exp(AA^\top) x - b \|_2^2
\end{align*}
where $\exp(\cdot)$ is applied entrywise to the matrix $AA^\top$.
\end{definition}

The attention kernel regression problem owing its name to the attention mechanism, as it could be viewed as a simplification of approximating one critical component of the attention matrix. 

\subsection{Our Result}

We present the informal version of our main result below.

\begin{theorem}[Informal Version of Theorem~\ref{thm:even}]\label{thm:informal_even}
Let $A \in \R^{n \times d}$ have full column rank, let $b \in \R^d$, and let $\kappa \geq \kappa(A)$ be an upper bound on the condition number of $A$, given as input. Let $\epsilon_{\mathrm{final}}, \delta_{\mathrm{final}} \in (0,0.1)$.

For the regression problem shown in Eq.~\eqref{eq:informal_even}, there exists an algorithm (Algorithm~\ref{alg:even}) that, with probability at least $1-\delta_{\mathrm{final}}$, stops within time
\begin{align*}
O ( (n d + d^3) \cdot ( j^2 \log \kappa + j \log(j / \epsilon_{\mathrm{final}}) ) \cdot \log^2(jn/\delta_{\mathrm{final}}) )
\end{align*}
and outputs a vector $x' \in \R^d$ such that $\| (A^\top A)^j x' - b \|_2 \leq  \epsilon_{\mathrm{final}} \| b \|_2$.
\end{theorem}

\begin{theorem}[Informal Version of Theorem~\ref{thm:odd}]\label{thm:informal_odd}
    Let $A \in \R^{n \times d}$ have full column rank, let $b \in \R^n$, and let $\kappa \geq \kappa(A)$ be an upper bound on the condition number of $A$, given as input. Let $\epsilon_{\mathrm{final}}, \delta_{\mathrm{final}} \in (0,0.1)$, and let $\OPT := \min_{x \in \R^d} \| A(A^\top A)^j x - b \|_2$.

For the regression problem shown in Eq.~\eqref{eq:informal_odd}, there exists an algorithm (Algorithm~\ref{alg:odd}) that, with probability at least $1-\delta_{\mathrm{final}}$, stops within time
\begin{align*}
O ( (n d + d^3) \cdot ( j^2 \log \kappa + j \log(j / \epsilon_{\mathrm{final}}) ) \cdot \log^2(jn/\delta_{\mathrm{final}}) )
\end{align*}
and outputs a vector $x' \in \R^d$ such that $\| A(A^\top A)^j x' - b \|_2 \leq (1+\epsilon_{\mathrm{final}}) \OPT + \epsilon_{\mathrm{final}} \| b \|_2$.
\end{theorem}

The term $\OPT$ in Theorem~\ref{thm:informal_odd} cannot be avoided. Since $A$ has full column rank, the vectors $A(A^\top A)^j x$ with $x \in \R^d$ form exactly the column space of $A$, so $\OPT = \|(I - AA^\dagger) b\|_2$ is the distance from $b$ to that space, and no output has a smaller residual. When $b$ lies in the column space of $A$, we have $\OPT = 0$, and the bound reads $\epsilon_{\mathrm{final}} \| b \|_2$.

Before proceeding, we compare our running time with exact methods. An exact method never forms the product $(A^\top A)^j$. It forms $A^\top A$ once in $O(nd^2)$ time, or in $O(nd^{\omega-1})$ time with fast matrix multiplication, where $\omega$ denotes the exponent of matrix multiplication. It factors $A^\top A$ once in $O(d^3)$ time, and then applies $(A^\top A)^{-1}$ $j$ times with pairs of triangular solves in $O(jd^2)$ time. This gives $x = (A^\top A)^{-j} b$ in the even case and $x = (A^\top A)^{-(j+1)} A^\top b$ in the odd case, so the exact cost is $O(nd^2 + d^3 + jd^2)$. Let $J := j^2 \log \kappa + j \log(j/\epsilon_{\mathrm{final}})$. Our algorithms run in time $O((nd + d^3) \cdot J \cdot \log^2(jn/\delta_{\mathrm{final}}))$. Up to the $\log^2(jn/\delta_{\mathrm{final}})$ factor, this is faster when $J \ll d$ and $n \gg dJ$. Against fast matrix multiplication, the two conditions become $J \ll d^{\omega-2}$ and $n \gg d^{4-\omega} J$.

Our first result regarding the attention kernel regression requires some extra assumptions. We assume that the rows of $A$ have norm at most $1$ and that the kernel matrix $\exp(AA^\top)$ is positive definite, and the algorithm is given an upper bound $\kappa$ on the condition number of a factor $Z$ in $\exp(AA^\top) = Z^\top Z$. The result follows the sketching and preconditioning approach of~\cite{swyz21} for Gaussian kernel regression. Unlike for the power regressions (Section~\ref{sec:fast_attention}), we do not give a procedure that computes such a bound faster than forming $\exp(AA^\top)$, so the comparison below assumes that $\kappa$ is known. Now we are ready to state our first result.
\begin{theorem}[Informal Version of Theorem~\ref{thm:formal_exp}]\label{thm:informal_exp}
Let $A \in \R^{n \times d}$ have rows of norm at most $1$, and let $b \in \R^n$. Suppose that $G := \exp(AA^\top)$ is positive definite, which holds whenever the rows of $A$ are distinct. Factor $G = Z^\top Z$. The condition number $\kappa(Z) = \sqrt{\lambda_{\max}(G)/\lambda_{\min}(G)}$ does not depend on the choice of the factor $Z$. Let $\kappa \geq \kappa(Z)$ be an upper bound given as input. Let $\epsilon_{\mathrm{final}}, \delta_{\mathrm{final}} \in (0,0.1)$, and let
\begin{align*}
m = O(\epsilon_{\mathrm{final}}^{-2} \kappa^2 n \log^3(n\kappa/\epsilon_{\rm final}) \log^3(nd\kappa/(\epsilon_{\rm final}\delta_{\rm final})))
\end{align*}
be the sketching dimension. There is an algorithm that, with probability at least $1-\delta_{\mathrm{final}}$, stops within time
\begin{align*}
O((mn + n^3) \cdot \log(\kappa/\epsilon_{\rm final}) \cdot \log^2(m/\delta_{\rm final}) + nd\log(n\kappa/\epsilon_{\rm final})\log(nd\kappa/(\epsilon_{\rm final}\delta_{\rm final})))
\end{align*}
and outputs a vector $x' \in \R^n$ such that
\begin{align*}
\| \exp(AA^\top) x' - b \|_2 \leq \epsilon_{\mathrm{final}} \cdot \| b \|_2.
\end{align*}
\end{theorem}
We compare our running time with the exact method. For $d \geq n$, computing $AA^\top$ takes $O(n^{\omega-1} d)$ time, by splitting the columns of $A$ into $d/n$ blocks of $n$ columns each. Applying the entrywise exponential takes $O(n^2)$ time, and solving the linear system takes $O(n^\omega)$ time, so the exact cost is $O(n^{\omega-1} d + n^\omega) = O(n^{\omega-1} d)$. Since $m = \wt{O}(\epsilon_{\rm final}^{-2} \kappa^2 n)$, where $\wt{O}$ hides polylogarithmic factors, our running time is $\wt{O}(\epsilon_{\rm final}^{-2} \kappa^2 n^2 + n^3) + O(nd \log(n\kappa/\epsilon_{\rm final}) \log(nd\kappa/(\epsilon_{\rm final}\delta_{\rm final})))$. The last term is at most the exact cost up to polylogarithmic factors, since $\omega \geq 2$. So, up to polylogarithmic factors, our algorithm is faster when $n^{\omega-1} d$ is larger than $\epsilon_{\rm final}^{-2} \kappa^2 n^2 + n^3$. Dividing by $n^{\omega-1}$, this means that $d$ is larger than $\epsilon_{\rm final}^{-2} \kappa^2 n^{3-\omega} + n^{4-\omega}$. With naive matrix multiplication, the exact cost is $O(n^2 d)$, which is the case $\omega = 3$, and the condition becomes $d \gg \epsilon_{\rm final}^{-2} \kappa^2 + n$, up to polylogarithmic factors. The bound $\kappa$ is never small. The $l = 0$ term of the Taylor series of $G$ (see the proof of Theorem~\ref{thm:gaussian_kernel}) gives $G \succeq {\bf 1}_n {\bf 1}_n^\top$, so $\lambda_{\max}(G) \geq n$, and $\lambda_{\min}(G) \leq G_{1,1} = \exp(\|a_1\|_2^2) \leq e$, where $a_1$ is the first row of $A$. Hence $\kappa^2 \geq \kappa(Z)^2 = \lambda_{\max}(G)/\lambda_{\min}(G) \geq n/e$. The sketch dimension used by the algorithm, $m = \wt{\Theta}(\epsilon_{\rm final}^{-2}\kappa^2 n)$, is therefore $\wt{\Omega}(\epsilon_{\rm final}^{-2} n^2)$, the running time is $\wt{\Omega}(\epsilon_{\rm final}^{-2} n^3)$, and the condition above requires $d \gg \epsilon_{\rm final}^{-2} n^{4-\omega}$. In particular, the term $n^{4-\omega}$ (and $n$ for $\omega = 3$) is always dominated.

We also give a second algorithm for attention kernel regression, whose sketch dimension, up to logarithmic factors, does not depend on $n$. It needs neither that $\exp(AA^\top)$ is positive definite nor a bound on a condition number. Instead, it takes a parameter $K\geq 1$ and measures its error against the best residual among solutions of norm at most $(K/n)\|b\|_2$.
\begin{theorem}[Informal Version of Theorem~\ref{thm:below_n}]\label{thm:informal_below_n}
Let $A\in\R^{n\times d}$ have rows $a_1,\ldots,a_n$ of norm at most $1$, let $b\in\R^n$, let $\epsilon,\delta\in(0,0.1)$, and let $K\geq 1$ be given as input. Let $\beta\in[1,n]$ be an upper bound on the ranks of the matrices $A^{\otimes l}\in\R^{n\times d^l}$ whose $i$-th row is $a_i^{\otimes l}$, for $1\leq l\leq q$, where $q=\lceil\log_2(212K^{3/2}/\epsilon^2)\rceil-1=\Theta(\log(K/\epsilon))$ (the bound $\beta=n$ is allowed). Let $\OPT_K(b)$ be the smallest residual $\|\exp(AA^\top)x-b\|_2$ over all $x\in\R^n$ with $\|x\|_2\leq(K/n)\|b\|_2$. There is an algorithm with sketch dimension $m=\wt O(\min\{\epsilon^{-2}K\beta,\ \epsilon^{-3}K^2\})$ that, with probability at least $1-\delta-O(\log(K/\epsilon))/\poly(n)$, outputs $x'\in\R^n$ with
\begin{align*}
\|\exp(AA^\top)x'-b\|_2\leq\epsilon\|b\|_2+(1+\sqrt{\epsilon})\cdot\OPT_K(b).
\end{align*}
When $m<n$, on the same event it stops within time $\wt O(nm+m^3+\epsilon^{-2}K\cdot\nnz(A))$, where $\wt O$ hides polylogarithmic factors in $ndK/(\epsilon\delta)$. In particular, $\|\exp(AA^\top)x'-b\|_2\leq\epsilon\|b\|_2$ whenever $b=\exp(AA^\top)x^*$ for some $x^*$ with $\|x^*\|_2\leq(K/n)\|b\|_2$.
\end{theorem}
A guarantee of $\epsilon\|b\|_2$ for every $b$ is out of reach for a natural class of algorithms with a small sketch. An algorithm whose output lies in a subspace of dimension $m\leq n/2$, chosen before $b$ is seen, cannot guarantee $\|\exp(AA^\top)x'-b\|_2\leq\epsilon\|b\|_2$ for every $b$, even when $\beta=1$ (Remark~\ref{rem:below_n_needed}). Our algorithm is not of this form, since its output contains $b$ itself, but when $m<n$ it does not give $\epsilon\|b\|_2$ for every $b$ either (Remark~\ref{rem:below_n_needed}). What changes is the guarantee. Its guarantee has the same form as that of Theorem~\ref{thm:informal_odd}: an additive $\epsilon\|b\|_2$ plus a multiple of an optimal residual, where the optimum is now taken over solutions of norm at most $(K/n)\|b\|_2$. The term $\OPT_K(b)$ is small for $b$ close to the subspace $U$ below, and only for $b$ close to a subspace of dimension at most $\min\{eK/\epsilon,1+q\beta\}$. Let $U$ be the span of the eigenvectors of $\exp(AA^\top)$ with eigenvalue at least $n/K$. Since $\tr(\exp(AA^\top))\leq en$, there are at most $eK$ such eigenvalues, and in fact $\dim U\leq\min\{eK,1+q\beta\}$. Moreover, $\OPT_K(b)\leq\dist(b,U)$ (Theorem~\ref{thm:below_n}). So the error is at most $\epsilon\|b\|_2$ for every $b\in U$, and in general it exceeds $\epsilon\|b\|_2$ by at most $(1+\sqrt{\epsilon})\cdot\dist(b,U)$. Conversely, let $U_\epsilon$ be the span of the eigenvectors of $\exp(AA^\top)$ with eigenvalue at least $\epsilon n/K$. Then $\dim U_\epsilon\leq\min\{eK/\epsilon,1+q\beta\}$ and $\OPT_K(b)\geq\dist(b,U_\epsilon)-\epsilon\|b\|_2$ for every $b$ (Theorem~\ref{thm:below_n}). So when $n$ is much larger than $\min\{eK/\epsilon,1+q\beta\}$, for most $b$ (for example $b$ uniform on a sphere) $\OPT_K(b)$ is close to $\|b\|_2$, and the bound above is larger than $\|b\|_2$, the residual of $x'=0$. Unlike $\OPT$ in Theorem~\ref{thm:informal_odd}, which every algorithm must pay, $\OPT_K(b)$ can be close to $\|b\|_2$ even when $\exp(AA^\top)x=b$ has an exact solution. If $\exp(AA^\top)$ is positive definite, taking $K\geq n/\lambda_{\min}(\exp(AA^\top))$ makes $\OPT_K(b)=0$ for every $b$. Since $\lambda_{\min}(\exp(AA^\top))\leq e$, such $K$ are at least $n/e$, and then the sketch has $m\geq n$ rows whenever $q/\poly(n)<1$, where $1/\poly(n)$ is the failure probability of the sketch of~\cite{akk+20}. With $K=\kappa^2$ for an upper bound $\kappa\geq\kappa(Z)$ as in Theorem~\ref{thm:informal_exp}, which satisfies $K\geq\lambda_{\max}(\exp(AA^\top))/\lambda_{\min}(\exp(AA^\top))\geq n/\lambda_{\min}(\exp(AA^\top))$, the sketch dimension is $\wt O(\epsilon^{-2}\kappa^2n)$, the order of Theorem~\ref{thm:informal_exp}. For the running time, recall that for $d\geq n$ the exact method takes $O(n^{\omega-1}d)$ time. Write $s\ll t$ if $s$ is smaller than $t$ by a large enough polylogarithmic factor. If $d\geq n$, $m\ll n^{\omega/3}$ and $\epsilon^{-2}K\ll n^{\omega-2}$, then $m<n$, $m^3\ll n^\omega\leq n^{\omega-1}d$, $nm\ll n^{1+\omega/3}\leq n^{\omega-1}d$ and $\epsilon^{-2}K\cdot\nnz(A)\leq\epsilon^{-2}Knd\ll n^{\omega-1}d$. So in this case the algorithm of Theorem~\ref{thm:informal_below_n} is faster than the exact method, for a weaker guarantee: the exact method reaches residual $0$ when $\exp(AA^\top)$ is positive definite, and computing the exact ridge solution $(\exp(AA^\top)+\lambda I_n)^{-1}b$ with $\lambda=\epsilon n/(4K)$, which meets the bound of Theorem~\ref{thm:informal_below_n} with $\epsilon/4$ in place of $\epsilon$ and $1$ in place of $1+\sqrt{\epsilon}$, costs the same $O(n^{\omega-1}d)$ time.

\subsection{Related Work}

\paragraph{Total Least Squares.}

The fitting method referred to as ``total least squares" has only recently been named as such in literature \citep{gv80}. However, it is not a new method and has been extensively studied in the statistical literature for a long time under various names, such as ``orthogonal regression," ``errors-in-variables," and ``measurement errors." In fact, the problem of univariate fitting, $n = 1, d = 1$, was first discussed in 1877 by Adcock \citep{a77}, and subsequent contributions were made by Pearson \citep{p01}, Koopmans \citep{k37}, and York \citep{y66}. The method has been rediscovered several times, often independently, and around 50 years ago, it was extended by Gleser \citep{g81} to multivariate problems of dimension $n>1$ and $d>1$.

In more recent times, the total least squares method has gained attention beyond the field of statistics. Golub and Van Loan \citep{gv80} were the first to study this problem in the field of numerical analysis, and they developed an algorithm based on the singular value decomposition. Staar \citep{s82} independently arrived at the same concept through geometrical insight into the properties of the singular value decomposition. Van Huffel and Vandewalle \citep{vv88} extended Golub and Van Loan's algorithm to cover all cases where their algorithm fails to produce a solution. They described the properties of these non-generic total least squares problems and proved that their proposed generalization still satisfies the total least squares criteria if additional constraints are imposed on the solution space. This approach, which appears to be different from the multivariate EIV regression analysis method studied by Gleser, is actually equivalent to it. Gleser's method is based on an eigenvalue-eigenvector analysis, while the total least squares method uses the singular value decomposition, which is more robust numerically in terms of algorithmic implementation. Moreover, the total least squares algorithm can compute the minimum norm solution whenever the total least squares solution is not unique. 

\paragraph{Attention Matrix.}

The attention matrix is a square matrix that represents correlations between words or tokens in natural language text. It has rows and columns that correspond to each token, and its entries denote the degree of correlation between them. This matrix is employed to determine the significance of each input token in a sequence when generating an output. In an attention mechanism, each input token is assigned a weight or score, which indicates its relevance or importance to the current output being produced. These scores are calculated based on a similarity function that compares the current output state to the input states.

There are several methods that attempt to estimate the heavy entries of the attention matrix by constraining the attention to local neighbors of queries using techniques such as \emph{locality-sensitive hashing} (LSH) \citep{kkl20,clp+21,syy21} or $k$-means clustering \citep{dkod20}. Another approach is to use random feature maps of Gaussian or exponential kernels to approximate the attention matrix \citep{cld+20}. Recently, Chen et al. \citep{cdw+21} demonstrated that combining LSH-based and random feature-based methods is more effective at approximating the attention matrix.

The computation of inner product attention \citep{zhdk23,as23,gms23,gsy23_hyper,bsz23,dls23,lsz23,gsy23_dp} is also a crucial task in contemporary machine learning. It is necessary for training large language models (LLMs) such as Transformer \citep{vsp+17}, GPT-1 \citep{rns+18}, BERT \citep{dclt18}, GPT-2 \citep{rwc+19}, GPT-3 \citep{bmr+20}, and ChatGPT, which are capable to handle natural language more effectively than conventional algorithms or smaller models. \cite{as23} provides both algorithm and hardness for static attention computation. \cite{bsz23} provides both algorithm and hardness for dynamically maintaining the attention matrix. \cite{gsy23_dp} shows how to compute the attention matrix with differential privacy guarantees. \cite{lsz23} studies the regression for matrix hyperbolic functions. \cite{gms23,dls23} consider algorithms for softmax attention.

\paragraph{Sketching.}

Sketching techniques are powerful tools for speeding up machine learning algorithms and optimization. The central idea is to break down a large input matrix into a much smaller matrix while preserving important characteristics. This enables downstream tasks to be performed on top of the smaller matrix. Many previous works have developed sketching algorithms with strong theoretical guarantees. For example, the Johnson-Lindenstrauss lemma in \citep{jl84} shows that under a certain high-dimensional space, the projecting points onto the lower-dimensional subspace may preserve the pairwise distances between the points. This property supports the development of faster algorithms for problems, such as approximate nearest neighbor search. 

Typically, there are two methods for employing sketching matrices. The first one is called sketch-and-solve, which applies sketching and then subsequently solves the problem in a one-shot manner. The second one is called iterate-and-sketch: sketching can be employed during each iteration of the algorithm to speed up the iteration.

Sketch-and-solve has led to faster algorithms in several domains: for low-rank approximation and linear regression \citep{nn13,cw13}, sketches are applied to reduce the problem size, so it is much easier to solve the smaller regression problem to get an approximated solution to the original problem; for approximating kernel matrices, as in \cite{lgtc15}; for fast computation of vector and matrix tensors, a long line of works~\citep{anw14,p13,pp13,dssw18} present a technique, called $\mathsf{TensorSketch}$ which can compress tensors down to much smaller core tensors, enabling faster algorithms for problems such as tensor regression~\citep{rsz22,djs+19,dssw18,swyz21}, CP decomposition \citep{ms21}; for column subset selection, sketching the data matrix speeds up column selection with provable approximation guarantees \citep{sg22,swz19b,jll+20,jll+21}. Moreover, it can be used for finding an approximate solution with $\ell_\infty$ guarantees \citep{psw17,syyz23_linf}, designing an efficient neural network training method \citep{qsy23}.

Beyond the classic sketch-and-solve paradigm, sketching has also been adapted to many iterative optimization algorithms. Notable examples include but not limited to non-convex optimization \citep{z22,als+22,szz21,syz21}, discrepancy minimization \citep{sxz22,dsw22}, John Ellipsoid computation \citep{syyz22}, Frank-Wolfe algorithm \citep{sxyz22,xss21}, linear programming \citep{jswz21,cls19,sy21,lsz+23}, reinforcement learning~\citep{ssx23}, dynamic kernel estimation \citep{qrs+22}, empirical risk minimization \citep{qszz23,lsz19}, federated learning \citep{swyz23}, semidefinite programming \citep{gs22}, rational database \citep{qjs+22}, matrix completion~\citep{gsyz23_completion}, matrix sensing \citep{qsz23}, submodular
maximization \citep{qsw23}, trace estimation \citep{jpwz21}, and projection maintenance \citep{syyz23_dp}.

\paragraph{Subspace Embedding.}

Given a matrix $A\in \R^{n\times d}$, we say a family of random matrices is an $(n, d, \epsilon,\delta)$ subspace embedding if, for any fixed orthonormal basis $U$ of $A$, with probability at $1-\delta$, any matrix $S\in \R^{m\times n}$ where $m$ is a function of $n, d, \epsilon,\delta$, from this distribution has the property that
\begin{align*}
    (1-\epsilon) I \preceq U^\top S^\top SU \preceq (1+\epsilon) I
\end{align*}
where $A\preceq B$ denotes the matrix $B-A$ is PSD. Equivalently, the singular values of $SU$ lie in $[\sqrt{1-\epsilon},\sqrt{1+\epsilon}] \subseteq [1-\epsilon,1+\epsilon]$. In the context of least-squares regression, it is usually more convenient to state in the following way: for any vector $x\in \R^d$, $\|SAx\|_2=(1\pm\epsilon) \|Ax\|_2$. Subspace embedding is first introduced in~\cite{s06}. Subsequently, structured family of random matrices that can be applied to $A$ efficiently was proposed~\citep{cw13,nn13,ldfu13,mm13}. For a more comprehensive review of subspace embedding, we refer readers to the book by Woodruff~\citep{w14}.

\paragraph{Roadmap.}

In Section~\ref{sec:short_preli}, we introduce the basic mathematical notations. 
In Section~\ref{sec:tech}, we give an overview of the techniques that we use in this paper.  
In Section~\ref{sec:main_result}, we present our algorithms and main results.
In Section~\ref{sec:conclusion}, we give a conclusion of this paper.


\section{Preliminary}
\label{sec:short_preli}

In this section, we introduce notations used throughout the paper.

We use $\R$ to denote the set of all real numbers. We use $\mathbb{Z}$ to denote the set of all integers and use $\mathbb{Z}_+$ to denote the set containing all positive integers. For any $n \in \mathbb{Z}_+$, we define $[n] := \{1, 2, 3, \dots, n \}$. 

For all $d \in \mathbb{Z}_+$, we use $\R^d$ to denote the set of all vectors with length $d$ and real entries and use $\mathbb{Z}_+^d$ to denote the set containing all vectors with length $d$ and entries of positive integers. For a vector $x \in \R^n$, we use $\|x\|_1$ to denote the $\ell_1$ norm, use $\|x\|_2$ to denote the $\ell_2$ norm, i.e., $\|x\|_2 = \left( \sum_{i = 1}^n x_i^2 \right)^\frac{1}{2}$, and use $\|x\|_\infty$ to denote the $\ell_\infty$ norm. 

For all $m,n \in \mathbb{Z}_+$, we use $\R^{m \times n}$ to denote the set containing all matrices with $m$ rows, $n$ columns, and real entries. For a matrix $A \in \R^{m \times n}$, we use $\|A\|$ to denote the spectral norm of $A$, i.e., $\|A\| = \max_{\|x\|_2 = 1} \|Ax\|_2$. We use $A^\top$ to denote the transpose of $A$. We use $\sigma_{\min}(A)$ to denote the minimum singular value of $A$, i.e., $\sigma_{\min}(A) = \min_{\|x\|_2 = 1} \|Ax\|_2$. Accordingly, we use $\sigma_{\max}(A)$ to denote the maximum singular value of $A$, so $\sigma_{\max}(A) = \|A\|$. Furthermore, we use $A^\dagger$ to denote the pseudo-inverse of $A$ and use $A^{-1}$ to denote the true inverse of $A$. The true inverse exists if $m = n$ and $\rank{(A)} = m$. For all $m_1, m_2, n \in \Z_+$, for all the matrices $A \in \R^{m_1 \times n}$ and $B \in \R^{m_2 \times n}$, we use $A \oplus B$ to denote the matrix $\begin{bmatrix} A \\ B \end{bmatrix}$. Correspondingly, for all $m_1, m_2, \ldots, m_p, p, n \in \Z_+$, for all the matrices $A_1 \in \R^{m_1 \times n}$, $A_2 \in \R^{m_2 \times n}$, $\cdots$, $A_p \in \R^{m_p \times n}$, we use $\oplus_{i = 1}^p A_i$ to denote $A_1 \oplus A_2 \oplus \dots \oplus A_p$.

We write $x = y\pm\epsilon$ if $x\in [y - \epsilon, y + \epsilon]$. ${\bf 1}_n$ represents an $n$-dimensional vector, whose entries are all $1$, and ${\bf 0}_n$ represents a $n$-dimensional vector, whose entries are all $0$. For a symmetric matrix $B\in\R^{n \times n}$, we say $B$ is positive semidefinite (denoted as $B \succeq 0$) if, for any vectors $x$ in $\R^n$, the inequality $x^\top B x \geq 0$ always holds. We also call $B$ a PSD matrix for simplicity.

\section{Technique Overview}
\label{sec:tech}

In Section~\ref{sub:tech:particular_odd}, we present the techniques we use to show the properties of a particular case of the odd power algorithm. In Section~\ref{sub:tech:particular_even}, we show the techniques of showing the correctness and the runtime of a particular case of the even power algorithm, and we recall the standard bound on the forward error for PSD matrices. In Section~\ref{sub:tech:general_even}, we offer the methods to generalize the particular case to all the even power cases. In Section~\ref{sub:tech:general_odd}, we elucidate the methods of generalizing the particular case to all the odd power cases. In Section~\ref{sub:tech:kernel}, we introduce the techniques which are used for analyzing the exponential regression. 

To give an algorithm for Theorem~\ref{thm:informal_even} and~\ref{thm:informal_odd}, we need to analyze the simpler case with only 3 and 4 matrices involved: 
\begin{align}\label{eq:simplied_odd}
    \min_{x \in \R^d} \| A A^\top A x - b \|_2
\end{align}
and
\begin{align}\label{eq:simplied_even}
    \min_{x \in \R^d} \| A^\top A A^\top A x - b \|_2.
\end{align}
Suppose we are provided with algorithms for these smaller problems, we can complete the picture for Theorem~\ref{thm:informal_even} and~\ref{thm:informal_odd} via induction. It remains to show how to solve the simpler cases.

\subsection{A Particular Case for Odd Power Algorithm}
\label{sub:tech:particular_odd}

To solve Eq.~\eqref{eq:simplied_odd}, our approach can be briefly described as follows: we first solve the regression $\min_{y\in \R^d}\|Ay-b\|_2$ then solve $\|A^\top Ax-y\|_2$. To understand the error, suppose $x'$ is the final approximate solution to $\min_{x\in \R^d}\|A^\top Ax-y'\|_2$, we can decompose the error as follows:
\begin{align*}
    \|AA^\top Ax'-b\|_2 = & ~ \|A(A^\top Ax'-y')+Ay'-b\|_2 \\
    \leq & ~ \|A\| \|A^\top Ax'-y' \|_2+\|Ay'-b\|_2.
\end{align*}
For the second term, it can be bounded as long as we utilize a fast solver with relative error dependence in terms of the optimal cost, the value of Eq.~\eqref{eq:simplied_odd}. For the first term, we deploy a fast, high accuracy solver for PSD regression, with the guarantee that
\begin{align*}
    \|A^\top Ax'-y'\|_2 \leq & ~ \epsilon\cdot \|y'\|_2,
\end{align*}
so it remains to get a good handle on $\|y'\|_2$. Note that it is the solution to the regression $\|Ay'-b\|_2$, and we can convert the error on the regression cost to the error on the quality of $y'$. More specifically, we have
\begin{align*}
    \|y'-y_*\|_2 \leq & ~ O(\sqrt{\epsilon})\cdot \frac{1}{\sigma_{\min}(A)}\cdot \|Ay_*-b\|_2
\end{align*}
for $y_*$ being the optimal solution to the regression. Finally, note that $y_*=A^\dagger b$, so we can simply bound its norm as $\|y_*\|_2\leq \frac{1}{\sigma_{\min}(A)}\cdot \|b\|_2$. Put it all together, we have
\begin{align*}
    \|y'\|_2 \leq & ~ \|y'-y_*\|_2+\|y_*\|_2 \\
    \leq & ~ \frac{1}{\sigma_{\min}(A)}(\sqrt{\epsilon}\cdot \|Ay_*-b\|_2+\|b\|_2).
\end{align*}
By setting $\epsilon$ properly, we can conclude
\begin{align*}
    \|AA^\top Ax'-b\|_2 \leq & ~ (1+\epsilon)\cdot \|AA^\top Ax_*-b\|_2+\epsilon \cdot \|b\|_2.
\end{align*}

For the running time, since we would need to set $\epsilon$ perhaps polynomially small to account for the error induced by backward-to-forward conversion and minimum singular values of $A$, fast solvers with runtime depends inverse polynomially on $\epsilon$ is infeasible. We resort to solvers with runtime depends on $\log(1/\epsilon)$~\citep{amt10,bpsw21}, which in turn provides the following final runtime bound:
\begin{align*}
    O((nd+d^3)\log(\kappa/\epsilon)\log^2(n/\delta)),
\end{align*}
where $\kappa \geq \kappa(A)$ is the given upper bound on the condition number of $A$ and $\delta$ is the failure probability of the algorithm.

\subsection{A Particular Case for Even Power Algorithm}
\label{sub:tech:particular_even}

To solve Eq.~\eqref{eq:simplied_even}, our algorithm is similar, except that we need to perform two rounds of regressions on PSD matrices: $\min_y\|A^\top Ay-b\|_2$ and $\min_x\|A^\top Ax-y\|_2$. It is more complicated to analyze this than to analyze the prior algorithm, because the first round is also a PSD regression, whose solver only guarantees a small residual. The analysis turns this residual guarantee of the PSD solver into an error bound with the standard bound $\|x'-x_*\|_2 \leq \|(A^\top A)^{-1}\| \cdot \|A^\top A x'-b\|_2$, where $x'$ is any approximate solution and $x_*$ is the exact solution of $A^\top A x = b$ (see~\cite[Chapter 7]{h02}).

We now recall how this bound gives the forward error for PSD matrices. Let us consider the regression problem
\begin{align}\label{eq:A^topAx_b_2}
    \min_{x \in \R^d} \| A^\top A x - b \|_2,
\end{align}
and we will use $x_*$ to denote the exact solution to the above problem and $x'$ to denote an approximate solution, i.e.,
\begin{align*}
    \|A^\top Ax'-b\|_2 \leq & ~ \epsilon\cdot \|b\|_2.
\end{align*}

We can bound the gap $\|x'-x^*\|_2$ as follows:
\begin{align*}
    \|x'-x_*\|_2 \leq \|  A^\top A ( x' - x_* ) \|_2 \cdot \| (A^\top A)^\dagger \|.
\end{align*}
The second term is simply $\sigma_{\min}^{-2}(A)$, and since $A^\top A$ is full rank, $A^\top Ax_*=b$, so the first term is $\epsilon \cdot\|b\|_2$. This gives us a final bound of
\begin{align*}
    \|x'-x_*\|_2 \leq & ~ \epsilon\cdot \frac{1}{\sigma^2_{\min}(A)}\cdot \|b\|_2,
\end{align*}
coupling with the analysis of error and runtime in the preceding section, we obtain the desired results.

\subsection{General Case for Even Power Algorithm}
\label{sub:tech:general_even}

Given an algorithm and analysis for the base case, we can now generalize to any even powers. Algorithm~\ref{alg:even} sets $b_0 = b$, and its step $i \in [j]$ computes $b_i$ by solving $\min_{x \in \R^d} \| A^\top A x - b_{i-1} \|_2$ with relative accuracy $\eta_i = \epsilon_{\mathrm{final}}/(4 j \kappa^{2(i-1)})$ and failure probability $\delta_{\mathrm{final}}/j$, where $\kappa \geq \kappa(A)$ is the given upper bound on the condition number of $A$. We track the accuracies $\epsilon_i := i \epsilon_{\mathrm{final}}/(2j)$ for $i \in \{0, 1, \ldots, j\}$, so $\epsilon_0 = 0$, $\epsilon_{i-1} \leq \epsilon_i$ and $\epsilon_j = \epsilon_{\mathrm{final}}/2$. We start by setting up the induction hypothesis, for which we assume for some $k \in \{0, 1, \ldots, j-1\}$ and all $i \in \{0, 1, \ldots, k\}$, we have
\begin{enumerate}
    \item $\| (A^\top A)^i b_i - b_0 \|_2 \leq \epsilon_i \| b_0 \|_2$,
    \item $\| b_i \|_2 \leq 2\sigma_{\min}^{-2i}(A) \| b_0 \|_2$,
    \item The first $k$ steps take total time at most $C \cdot (nd + d^3) \cdot \sum_{i=1}^{k} ( (2i-1) \log \kappa + \log(4j/\epsilon_{\mathrm{final}}) ) \cdot \log^2(jn/\delta_{\mathrm{final}})$ for a constant $C$,
    \item The probability that statements 1, 2 and 3 fail for $k$ is at most $k \delta_{\mathrm{final}}/j$.
\end{enumerate}

We want to show that this probability bound also holds for $k + 1$, and that statements 1, 2 and 3 for $k+1$ follow from statements 1, 2 and 3 for $k$ whenever step $k+1$ succeeds. To prove the first statement, we need to analyze $\| (A^\top A)^{k + 1} b_{k+1} - b_0 \|_2$:
\begin{align}\label{eq:A^topA^k+1b_k+1_b_0}
    & ~ \| (A^\top A)^{k + 1} b_{k+1} - b_0 \|_2 \notag\\
    \leq & ~ \| (A^\top A)^{k}\| \cdot \| A^\top A b_{k+1} - b_k\|_2 + \|(A^\top A)^{k} b_k - b_0 \|_2.
\end{align}
The two terms, and the two factors of the first term, can be bounded as follows:
\begin{itemize}
    \item $\|(A^\top A)^{k} b_k - b_0 \|_2$ can be bounded by $\epsilon_k \| b_0 \|_2$ based on induction hypothesis for $i=k$.
    \item $\| (A^\top A)^{k}\|$ can be bounded by $\sigma_{\max}^{2k}(A)$ based on Fact~\ref{fac:norm}.
    \item $\| A^\top A b_{k+1} - b_k \|_2$ can be bounded by $\eta_{k+1} \| b_k \|_2$ with $\eta_{k+1} = \epsilon_{\mathrm{final}}/(4 j \kappa^{2k})$ based on our two matrices version of PSD regression (see Lemma~\ref{lem:regression_solver}).
\end{itemize}
With the second statement for $i = k$, the first term is at most
\begin{align*}
    \sigma_{\max}^{2k}(A) \cdot \eta_{k+1} \cdot 2\sigma_{\min}^{-2k}(A) \| b_0 \|_2 = \Big( \frac{\kappa(A)}{\kappa} \Big)^{2k} \cdot \frac{\epsilon_{\mathrm{final}}}{2j} \cdot \| b_0 \|_2 \leq \frac{\epsilon_{\mathrm{final}}}{2j} \cdot \| b_0 \|_2.
\end{align*}
Adding the two terms gives $(\epsilon_k + \epsilon_{\mathrm{final}}/(2j)) \| b_0 \|_2 = \epsilon_{k+1} \| b_0 \|_2$.

For the second statement, the goal is to bound $\| b_{k + 1} \|_2$. By using the property of the spectral norm, we can get
\begin{align}\label{eq:b_k+1}
    \| b_{k + 1} \|_2 \leq \| ((A^\top A)^{k+1})^{-1} \| \cdot \| (A^\top A)^{k+1}  b_{k+1} \|_2.
\end{align}
These two terms can be bounded similarly, we include the reasoning here:
\begin{itemize}
    \item $\| ((A^\top A)^{k+1})^{-1} \|$ can be bounded by $\sigma^{-2(k+1)}_{\min}(A)$ based on Fact~\ref{fac:norm}.
    \item $\| (A^\top A)^{k+1}  b_{k+1} \|_2$ can be bounded $2 \| b_0 \|_2$ based on the first statement and triangle inequality.
\end{itemize}

The remaining statements are rather straightforward. By Lemma~\ref{lem:regression_solver}, step $k+1$ takes $O((nd + d^3) \log(\kappa/\eta_{k+1}) \log^2(jn/\delta_{\mathrm{final}}))$ time, and $\log(\kappa/\eta_{k+1}) = (2k+1) \log \kappa + \log(4j/\epsilon_{\mathrm{final}})$. Adding this to the time of the first $k$ steps gives the third statement. The failure probability follows from a union bound, since each step fails with probability at most $\delta_{\mathrm{final}}/j$. Summing over the $j$ steps and using $\sum_{i=1}^{j} (2i-1) = j^2$, the total time is $O((nd + d^3) \cdot (j^2 \log \kappa + j \log(j/\epsilon_{\mathrm{final}})) \cdot \log^2(jn/\delta_{\mathrm{final}}))$. The factor $j^2 \log \kappa$ comes from the accuracy $\eta_i$ of step $i$, which shrinks like $\kappa^{-2(i-1)}$: an error made at step $i$ is later multiplied by up to $\sigma_{\max}^{2(i-1)}(A)$, while $\| b_{i-1} \|_2$ can be as large as $2\sigma_{\min}^{-2(i-1)}(A) \| b_0 \|_2$.

\subsection{General Case for Odd Power Algorithm}
\label{sub:tech:general_odd}

For general odd powers in the form of $\min_{x \in \R^d} \|A(A^\top A)^j x - b \|_2$, our argument is similar to preceding section, with a minor modification that we start by solving a linear regression on $A$ instead of a PSD regression on $A^\top A$. Similar to Section~\ref{sub:tech:particular_odd}, we will have to convert error on the cost to error on the solution, and the remaining are similar to that of Section~\ref{sub:tech:general_even}. We run the even power algorithm with accuracy $\epsilon_{\mathrm{final}}/\kappa$, so the running time has the same order as in the even case, and the guarantee has the same form as in Section~\ref{sub:tech:particular_odd}: the residual is at most $(1+\epsilon_{\mathrm{final}}) \min_{x \in \R^d} \|A(A^\top A)^j x - b\|_2 + \epsilon_{\mathrm{final}} \|b\|_2$.

\subsection{Attention Kernel Regression}
\label{sub:tech:kernel}

We start by recalling the definition of attention kernel regression: given an input matrix $A\in \R^{n\times d}$ and a response $b\in \R^n$, our task is in two-folds: 1).\ compute or approximate $\exp(AA^\top)$; 2).\ solve the regression on top of the (approximated) kernel matrix. A simple, exact algorithm would be computing the Gram matrix $AA^\top$, applying entrywise exponential activation and solving the regression. This would take $O(n^2d + n^3)$ time with naive matrix multiplication, inefficient if $d$ is large. 

The key is to identify the kernel structure of $\exp(AA^\top)$, where the $(i,j)$-th entry is $\exp(\langle a_i, a_j\rangle)$. We can apply sketches for polynomial kernels to efficiently approximate this kernel matrix, and then solve the regression with high precision. We follow the approach of~\cite{swyz21} for Gaussian kernel regression. For simplicity, we let $G:=\exp(AA^\top)$, which we assume to be positive definite, and we factor $G = Z^\top Z$. The goal is to find $x' \in \R^n$ such that 
\begin{align*}
    \|Gx' - b\|_2 \leq \epsilon \|b\|_2,
\end{align*}
where $\epsilon\in (0,0.1)$ is an accuracy parameter. First, we compute a sketch $W_g(A)\in \R^{m\times n}$ such that $(1-\epsilon') G \preceq W_g(A)^\top W_g(A) \preceq (1+\epsilon') G$. To do so, we use that the rows of $A$ have norm at most $1$, we truncate the Taylor series of the entrywise exponential after $q = \Theta(\log(n\kappa/\epsilon'))$ terms, where $\kappa \geq \kappa(Z)$ is a given upper bound, and we sketch each polynomial term with the sketch of~\cite{swyz21}. This gives $m = \wt{O}(\epsilon'^{-2} n)$. A smaller sketch dimension is not possible: since $G$ is positive definite, the lower bound $(1-\epsilon') G \preceq W_g(A)^\top W_g(A)$ forces $W_g(A)$ to have rank $n$, so $m \geq n$. Second, we solve the PSD regression $\min_{x \in \R^n} \|W_g(A)^\top W_g(A) x - b\|_2$ with Algorithm~\ref{alg:fast_psd_regression}. It computes a QR preconditioner $R \in \R^{n \times n}$ from a sketch of $W_g(A)$, runs preconditioned gradient descent, and returns $x'$ with $\|W_g(A)^\top W_g(A) x' - b\|_2 \leq (\epsilon/2) \|b\|_2$. Third, we move this guarantee from $W_g(A)^\top W_g(A)$ to $G$. We write $W_g(A)^\top W_g(A) = G^{1/2}(I+E)G^{1/2}$ with $\|E\| \leq \epsilon'$, and we note that
\begin{align*}
    Gx' - b = (W_g(A)^\top W_g(A) x' - b) - G^{1/2} E G^{1/2} x'.
\end{align*}
The norm of the last term is at most $2\epsilon'\kappa(Z)\|b\|_2$, because $\|G^{1/2}\| \cdot \|G^{-1/2}\| = \kappa(Z)$. To make it at most $(\epsilon/2)\|b\|_2$, we take $\epsilon' = \epsilon/(4\kappa)$. Then $m = \wt{O}(\epsilon^{-2}\kappa^2 n)$, and for failure probability $\delta$, the total running time is
\begin{align*}
    & ~ O((mn + n^3) \cdot \log(\kappa/\epsilon) \cdot \log^2(m/\delta)) \\
    + & ~ O(nd\log(n\kappa/\epsilon)\log(nd\kappa/(\epsilon\delta))).
\end{align*}
Since $\kappa^2 \geq \kappa(Z)^2 = \lambda_{\max}(G)/\lambda_{\min}(G) \geq n/e$ (the $l = 0$ term of the Taylor series gives $\lambda_{\max}(G) \geq n$, and $\lambda_{\min}(G) \leq G_{1,1} \leq e$), the sketch dimension is always $\wt{\Omega}(\epsilon^{-2} n^2)$.

Our second algorithm uses a sketch whose dimension, up to logarithmic factors, does not depend on $n$. For positive definite $G$, a sketch $W$ with $(1-\epsilon')G \preceq W^\top W$ needs $m\geq n$, but a spectral approximation of $G+\lambda I_n$ does not. We let $\lambda=\epsilon n/(4K)$ for a parameter $K\geq 1$, truncate the Taylor series after $q=\Theta(\log(K/\epsilon))$ terms, keep the degree-$0$ term ${\bf 1}_n{\bf 1}_n^\top$ exactly, and sketch each degree $1\leq l\leq q$ with the oblivious sketch of~\cite{akk+20} for the ridge $l!\,\lambda/(q+1)$ (as in the proof of Theorem~5 of~\cite{akk+20} for the Gaussian kernel, with this ridge split). The size of this sketch depends on the statistical dimension of the degree-$l$ term at that ridge, which is at most $\min\{\beta,4(q+1)K/(\epsilon\cdot l!)\}$, where $\beta$ bounds the ranks of the matrices $A^{\otimes l}\in\R^{n\times d^l}$ with rows $a_i^{\otimes l}$, $1\leq l\leq q$. Stacking the degrees gives a sketch $W\in\R^{m\times n}$ with $m=\wt O(\min\{\epsilon^{-2}K\beta,\epsilon^{-3}K^2\})$ rows such that, with probability at least $1-q/\poly(n)$, $(1-\epsilon_0)(G+\lambda I_n)\preceq W^\top W+\lambda I_n\preceq(1+\epsilon_0)(G+\lambda I_n)$ with $\epsilon_0=\epsilon/(8\sqrt{eK})$. So the sketch is small whenever $K$ is small, and also when $\beta$ is small. If the rows of $A$ lie in an $r$-dimensional subspace and $\beta<n$, then the explicit monomial features of degree at most $q$, with $\binom{r+q}{q}$ columns, also give the truncated kernel without sketching. The bound in terms of $\beta$ is the useful one when $\binom{r+q}{q}$ is much larger than $\beta$, for example when the rows lie on a few lines through the origin. When $m\geq n$, we solve $(W^\top W+\lambda I_n)x'=b$ with the same solver. When $m<n$, we solve the $m\times m$ system $(WW^\top+\lambda I_m)v=Wb$ with Algorithm~\ref{alg:fast_psd_regression} and return $x'=(b-W^\top v)/\lambda$. By the push-through identity $(W^\top W+\lambda I_n)W^\top=W^\top(WW^\top+\lambda I_m)$, this is $(W^\top W+\lambda I_n)^{-1}b$ when $v$ is exact. The vector $x'$ contains $b$ itself, but when $m<n$ this does not give $\epsilon\|b\|_2$ for every $b$ (Remark~\ref{rem:below_n_needed}). The guarantee is relative to an optimum instead. The ridge costs $\lambda\|x\|_2\leq(\epsilon/4)\|b\|_2$ on every solution $x$ of norm at most $(K/n)\|b\|_2$. This leads to the guarantee $\|Gx'-b\|_2\leq\epsilon\|b\|_2+(1+\sqrt{\epsilon})\OPT_K(b)$, where $\OPT_K(b)$ is the smallest residual $\|Gx-b\|_2$ over such $x$ (Theorem~\ref{thm:below_n}).

\section{Main Result}
\label{sec:main_result}

We now state our main algorithms and results. For the even/odd power regression, note that an exact method never forms the product $(A^\top A)^j$. It forms $A^\top A$ once in $O(nd^2)$ time ($O(nd^{\omega-1})$ time with fast matrix multiplication), factors it once in $O(d^3)$ time, and applies $(A^\top A)^{-1}$ $j$ times with pairs of triangular solves in $O(jd^2)$ time. This gives $x = (A^\top A)^{-j} b$ in the even case and $x = (A^\top A)^{-(j+1)} A^\top b$ in the odd case, so the exact cost is $O(nd^2 + d^3 + jd^2)$. While the dependence on $n$ is nearly optimal, the dependence on $d$ makes this impractical for large-scale problems. Let $J := j^2 \log \kappa + j \log(j/\epsilon_{\mathrm{final}})$, where $\kappa \geq \kappa(A)$ is the given upper bound on the condition number of $A$. Our new algorithms achieve a running time of $O((nd + d^3) \cdot J \cdot \log^2(jn/\delta_{\mathrm{final}}))$. Up to the $\log^2(jn/\delta_{\mathrm{final}})$ factor, this is faster when $J \ll d$ and $n \gg dJ$. Against fast matrix multiplication, the two conditions become $J \ll d^{\omega-2}$ and $n \gg d^{4-\omega} J$.

\begin{algorithm}[!ht]\caption{Algorithm for solving even power regression $\min_{x \in \R^d} \|(A^\top A)^j x - b \|_2$.}
\label{alg:even}
\begin{algorithmic}[1]
\Procedure{EvenPowers}{$A \in \R^{n \times d}, b \in \R^d,n \in \mathbb{Z}_+,d \in \mathbb{Z}_+,j \in \mathbb{Z}_+,\kappa \geq \kappa(A),\epsilon_{\mathrm{final}} \in (0, 0.1),\delta_{\mathrm{final}} \in (0, 0.1)$}
    \State $b_0 \gets b$
    \For{$k  =1 \to j$}
        \State $\delta_k \gets \delta_{\mathrm{final}}/j$
        \State $\eta_k \gets \epsilon_{\mathrm{final}} / (4 j \kappa^{2(k-1)})$
    \State $b_{k} \gets \textsc{FastPSDRegression}(A \in \R^{n \times d},b_{k-1} \in \R^d,n,d,\eta_k,\delta_k)$
    \EndFor
    \State $x' \gets b_j$
    \State \Return $x'$
\EndProcedure
\end{algorithmic}
\end{algorithm}

\begin{algorithm}[!ht]\caption{Algorithm for solving odd power regression $\min_{x \in \R^d} \|A(A^\top A)^j x - b \|_2$.}
\label{alg:odd}
\begin{algorithmic}[1]
\Procedure{OddPowers}{$A \in \R^{n \times d}, b \in \R^n,n \in \mathbb{Z}_+ ,d \in \mathbb{Z}_+,j \in \mathbb{Z}_+,\kappa \geq \kappa(A),\epsilon_{\mathrm{final}} \in (0, 0.1),\delta_{\mathrm{final}} \in (0, 0.1)$}
    \State $\epsilon_{1} \gets  0.1 \epsilon_{\mathrm{final}}$
    \State $\delta_{1} \gets \delta_{\mathrm{final}}/2$
    \State $b_1 \gets \textsc{FastLinearRegression}(A \in \R^{n \times d},b \in \R^n,n,d,\epsilon_{1},\delta_{1})$ \Comment{$b_1 \in \R^d$}
    \State $\epsilon_{\mathrm{even}} \gets \epsilon_{\mathrm{final}} / \kappa$
    \State $\delta_{\mathrm{even}} \gets \delta_{\mathrm{final}}/2$
    \State $x'\gets \textsc{EvenPowers}(A \in \R^{n \times d },b_1 \in \R^d,n,d,j,\kappa,\epsilon_{\mathrm{even}},\delta_{\mathrm{even}})$
    \State \Return $x'$
\EndProcedure
\end{algorithmic}
\end{algorithm}

\begin{algorithm}[!ht]\caption{Algorithm for solving attention kernel regression $\min_{x\in \R^n} \|\exp(AA^\top)x-b\|_2$ (following Algorithm 2 of~\cite{swyz21}).}
\label{alg:alg_2_in_swyz21}
\begin{algorithmic}[1]
\Procedure{PreconditionedGradientDescent}{$A\in \R^{n\times d}, b\in \R^n, \kappa \geq \kappa(Z), \epsilon_{\rm final}\in (0,0.1),\delta_{\rm final}\in (0,0.1)$}  \Comment{Theorem~\ref{thm:formal_exp}}
    \State $X \gets A^\top$ \Comment{$\exp(AA^\top) = Z^\top Z$}
    \State $\epsilon' \gets \epsilon_{\rm final}/(4\kappa)$
    \State Let $q$ be the smallest integer with $q+1 \geq \max\{2e, \log_2(4n\kappa^2/\epsilon')\}$
    \State Let $m$ be the sketch dimension of Theorem~\ref{thm:gaussian_kernel} with $r = 1$, accuracy $\epsilon'$ and failure probability $\delta_{\rm final}/2$
    \State Compute $W_g(X) \in \R^{m \times n}$ with accuracy $\epsilon'$ and failure probability $\delta_{\rm final}/2$ \Comment{Theorem~\ref{thm:gaussian_kernel}}
    \State $x' \gets \textsc{FastPSDRegression}(W_g(X) \in \R^{m \times n}, b \in \R^n, m, n, \epsilon_{\rm final}/2, \delta_{\rm final}/2)$ \Comment{Algorithm~\ref{alg:fast_psd_regression}}
    \State \Return $x'$
\EndProcedure
\end{algorithmic}
\end{algorithm}

\begin{algorithm}[!ht]
\caption{Attention kernel regression with a ridge-regularized sketch, whose sketch dimension can be smaller than $n$.}\label{alg:ridge_sketch}
\begin{algorithmic}[1]
\Procedure{RidgeSketchRegression}{$A\in\R^{n\times d}, b\in\R^n, K\geq 1, \beta\in[1,n], \epsilon\in(0,0.1), \delta\in(0,0.1)$} \Comment{Theorem~\ref{thm:below_n}}
\State $X\gets A^\top$, \quad $\lambda\gets \epsilon n/(4K)$, \quad $\epsilon_s\gets \epsilon/(16\sqrt{eK})$, \quad $\eta\gets \epsilon^3/(500K^2)$
\State $q\gets \lceil \log_2(212K^{3/2}/\epsilon^2)\rceil-1$ \Comment{$q=\Theta(\log(K/\epsilon))$}
\State $Z_0\gets {\bf 1}_n^\top\in\R^{1\times n}$, \quad $m_0\gets 1$
\For{$l=1\to q$}
\State $\mu_l\gets\max\{1,\min\{\beta,\lceil 4(q+1)K/(\epsilon\cdot l!)\rceil\}\}$
\State Draw $\Pi^l\in\R^{m_l\times d^l}$ from the distribution of Theorem~\ref{thm:akk20} with $p=l$, accuracy $\epsilon_s$, statistical dimension $\mu_l$ and $m_l=\wt\Theta(l^4\mu_l/\epsilon_s^2)$ rows, independently of $\Pi^1,\ldots,\Pi^{l-1}$
\State $Z_l\gets \Pi^l X^{\otimes l}\in\R^{m_l\times n}$ \Comment{as in Theorem~\ref{thm:akk20}, without forming $X^{\otimes l}$}
\EndFor
\State $W\gets \oplus_{l=0}^q \frac{Z_l}{\sqrt{l!}}\in\R^{m\times n}$, where $m=\sum_{l=0}^q m_l$
\If{$m<n$}
\State $u\gets Wb\in\R^m$, \quad $C\gets W^\top\oplus\sqrt{\lambda}\, I_m\in\R^{(n+m)\times m}$ \Comment{$C^\top C=WW^\top+\lambda I_m$}
\State $\wt v\gets\textsc{FastPSDRegression}(C,u,n+m,m,\eta,\delta)$ \Comment{Algorithm~\ref{alg:fast_psd_regression}}
\State $x'\gets (b-W^\top\wt v)/\lambda$ \Comment{$x'\approx(W^\top W+\lambda I_n)^{-1}b$}
\Else
\State $x'\gets\textsc{FastPSDRegression}(W\oplus\sqrt{\lambda}\, I_n,b,m+n,n,\eta,\delta)$ \Comment{$x'\approx(W^\top W+\lambda I_n)^{-1}b$}
\EndIf
\State \Return $x'$
\EndProcedure
\end{algorithmic}
\end{algorithm}

\begin{theorem}[Main Result for Even/Odd Power Regression]\label{thm:main_result}
Let $A \in \R^{n \times d}$ be a matrix with full column rank, and let $j$ be a positive integer. Let $\kappa \geq \kappa(A)$ be an upper bound on the condition number of $A$, given as input.
Let $\epsilon_{\mathrm{final}} \in (0,0.1)$ denote the accuracy parameter. Let $\delta_{\mathrm{final}} \in (0,0.1)$ denote the failure probability.
Then, there exist algorithms (Algorithm~\ref{alg:even} and \ref{alg:odd}) with the following guarantees. Each of them, with probability at least $1-\delta_{\mathrm{final}}$, stops within time
\begin{align*}
O ( (n d + d^3) \cdot ( j^2 \log \kappa + j \log(j / \epsilon_{\mathrm{final}}) ) \cdot \log^2(jn/\delta_{\mathrm{final}}) )
\end{align*}
and outputs a vector $x' \in \R^d$ as follows.
\begin{itemize}
    \item (Algorithm~\ref{alg:even}) Given $b \in \R^d$, the output satisfies $\| (A^\top A)^j x' - b \|_2 \leq \epsilon_{\mathrm{final}} \| b \|_2$.
    \item (Algorithm~\ref{alg:odd}) Given $b \in \R^n$, let $\OPT := \min_{x \in \R^d} \| A(A^\top A)^j x - b \|_2$. The output satisfies
    \begin{align*}
        \| A(A^\top A)^j x' - b \|_2 \leq (1+\epsilon_{\mathrm{final}}) \OPT + \epsilon_{\mathrm{final}} \| b \|_2.
    \end{align*}
\end{itemize}
\end{theorem}

We defer the details of the proofs to Appendix~\ref{sec:odd} and \ref{sec:even}.

\begin{theorem}[Main Result for Attention Kernel Regression]
    Let $A\in \R^{n\times d}$ and $b\in \R^n$, and let $\epsilon_{\rm final},\delta_{\rm final}\in (0,0.1)$ be the accuracy parameter and the failure probability, respectively. Suppose for all $i\in [n]$, $\|a_i\|_2\leq 1$, where $a_i$ is the $i$-th row of matrix $A$. Suppose that $G := \exp(AA^\top)$ is positive definite, which holds whenever $a_1, \ldots, a_n$ are distinct. Factor $G = Z^\top Z$. The condition number $\kappa(Z) = \sqrt{\lambda_{\max}(G)/\lambda_{\min}(G)}$ does not depend on the choice of the factor $Z$. Let $\kappa \geq \kappa(Z)$ be an upper bound given as input, and let
    \begin{align*}
        m = O(\epsilon_{\rm final}^{-2}\kappa^2 n \log^3(n\kappa/\epsilon_{\rm final}) \log^3(nd\kappa/(\epsilon_{\rm final}\delta_{\rm final})))
    \end{align*}
    be the sketch dimension. Then, there exists a randomized algorithm (Algorithm~\ref{alg:alg_2_in_swyz21}) that, with probability at least $1-\delta_{\rm final}$, computes a vector $x'\in \R^n$ such that
    \begin{align*}
        \|\exp(AA^\top)x'-b\|_2 \leq & ~ \epsilon_{\rm final}\cdot \|b\|_2
    \end{align*}
    in time
    \begin{align*}
        O((mn + n^3) \cdot \log(\kappa/\epsilon_{\rm final}) \cdot \log^2(m/\delta_{\rm final}) + nd\log(n\kappa/\epsilon_{\rm final})\log(nd\kappa/(\epsilon_{\rm final}\delta_{\rm final}))).
    \end{align*}
\end{theorem}
    
The algorithm and its analysis follow the sketching and preconditioning approach of~\cite{swyz21} for Gaussian kernel regression. For the complete algorithm and detailed analysis, we refer readers to Appendix~\ref{sec:attention_kernel}.

The sketch dimension in the theorem above is always at least of order $n^2/\epsilon_{\rm final}^2$, since $\kappa^2\geq n/e$. Our second algorithm for attention kernel regression (Algorithm~\ref{alg:ridge_sketch}) uses a sketch whose dimension, up to logarithmic factors, does not depend on $n$. It does not need $\exp(AA^\top)$ to be positive definite or a bound on a condition number. Instead, it takes a parameter $K\geq 1$, and its error is relative to the best residual among solutions of norm at most $(K/n)\|b\|_2$.

\begin{theorem}[Main Result for Attention Kernel Regression with Sketch Dimension Below $n$]\label{thm:main_below_n}
    Let $A\in \R^{n\times d}$ and $b\in \R^n$, and suppose for all $i\in [n]$, $\|a_i\|_2\leq 1$, where $a_i$ is the $i$-th row of matrix $A$. Let $\epsilon,\delta\in(0,0.1)$ be the accuracy parameter and the failure probability, and let $K\geq 1$ be given as input. Let $q=\Theta(\log(K/\epsilon))$ be as in Algorithm~\ref{alg:ridge_sketch}, and let $\beta\in[1,n]$ be an upper bound on $\rank(A^{\otimes l})$ for all $1\leq l\leq q$, where $A^{\otimes l}\in\R^{n\times d^l}$ is the matrix whose $i$-th row is $a_i^{\otimes l}$. The choice $\beta=n$ is always allowed. Let
    \begin{align*}
        \OPT_K(b):=\inf\Big\{\|\exp(AA^\top)x-b\|_2:\ x\in\R^n,\ \|x\|_2\leq \frac{K}{n}\|b\|_2\Big\}.
    \end{align*}
    Then Algorithm~\ref{alg:ridge_sketch} uses a sketch with
    \begin{align*}
        m=\wt O\big(\min\{\epsilon^{-2}K\beta,\ \epsilon^{-3}K^2\}\big)
    \end{align*}
    rows, where $\wt O$ hides $\poly\log(ndK/\epsilon)$ factors. With probability at least $1-\delta-O(\log(K/\epsilon))/\poly(n)$, it stops within time
    \begin{align*}
        O\big((nm+\min\{m,n\}^3)\cdot\log(K/\epsilon)\cdot\log^2((n+m)/\delta)\big)+\wt O\big(nm+\epsilon^{-2}K\cdot\nnz(A)\big)
    \end{align*}
    and outputs a vector $x'\in\R^n$ such that
    \begin{align*}
        \|\exp(AA^\top)x'-b\|_2\leq \epsilon\|b\|_2+(1+\sqrt{\epsilon})\cdot\OPT_K(b).
    \end{align*}
    In particular, $\|\exp(AA^\top)x'-b\|_2\leq\epsilon\|b\|_2$ whenever $b=\exp(AA^\top)x^*$ for some $x^*\in\R^n$ with $\|x^*\|_2\leq(K/n)\|b\|_2$.
\end{theorem}

The sketch has $m<n$ rows whenever $n$ exceeds $\min\{\epsilon^{-2}K\beta,\ \epsilon^{-3}K^2\}$ by a large enough $\poly\log(ndK/\epsilon)$ factor. We prove Theorem~\ref{thm:main_below_n} in Appendix~\ref{sec:kernel_below_n} (Theorem~\ref{thm:below_n}). There we also show that $\OPT_K(b)$ is at most the distance from $b$ to the span of the eigenvectors of $\exp(AA^\top)$ with eigenvalue at least $n/K$, and that this span has dimension at most $\min\{eK,1+q\beta\}$.

\section{Conclusion}\label{sec:conclusion}

Large language models have demonstrated remarkable performance in various tasks. One significant aspect, from a computational standpoint, is the computation of the attention matrix. Earlier research had thoroughly examined the feasibility and limitations of approximating the attention matrix. In this study, we introduce several new problems concerning computing or approximating proxies for the attention matrix, via powers of the matrix $A^\top A$ or entrywise exponential for the Gram matrix of data. We provide fast algorithms for the power regressions, based on sketching and preconditioning. For attention kernel regression, we obtain an algorithm by following the sketching and preconditioning approach of~\cite{swyz21} for Gaussian kernel regression. For rows of $A$ of norm at most $1$, we also give an algorithm whose sketch dimension, up to logarithmic factors, does not depend on $n$. For a given parameter $K\geq 1$, its error is at most $\epsilon\|b\|_2$ plus $(1+\sqrt{\epsilon})$ times the best residual among solutions of norm at most $(K/n)\|b\|_2$.

We note that while our algorithm for regression against a product of matrices runs in nearly linear time when $n \geq d^2$ and $j$ and $\kappa$ are small, its running time grows like $j^2 \log \kappa$ in the number of matrices $j$. In contrast, the exact method depends on $j$ only through the additive $jd^2$ term. In our analysis of the alternating scheme, an error made at step $k$ can be magnified by up to $\kappa^{2(k-1)}$, which forces step $k$ to be solved to accuracy about $\kappa^{-2(k-1)}$. We do not know whether this $j^2 \log \kappa$ dependence can be avoided. It will be interesting to devise an algorithm that both runs in nearly linear time and has better dependence on $j$. 

\subsubsection*{Acknowledgements}
Lichen Zhang was supported in part by NSF CCF-1955217 and NSF DMS-2022448. The corrections in this version were found and written with the help of Claude Opus 5.5. The authors have verified the corrections and take full responsibility for the content of this paper.
\ifdefined\isarxiv
\else
\bibliographystyle{plainnat}
\bibliography{ref}
\section*{Checklist}

 \begin{enumerate}

 \item For all models and algorithms presented, check if you include:
 \begin{enumerate}
   \item A clear description of the mathematical setting, assumptions, algorithm, and/or model. [Yes]
   \item An analysis of the properties and complexity (time, space, sample size) of any algorithm. [Yes]
   \item (Optional) Anonymized source code, with specification of all dependencies, including external libraries. [Not Applicable]
 \end{enumerate}

 \item For any theoretical claim, check if you include:
 \begin{enumerate}
   \item Statements of the full set of assumptions of all theoretical results. [Yes]
   \item Complete proofs of all theoretical results. [Yes]
   \item Clear explanations of any assumptions. [Yes]     
 \end{enumerate}

 \item For all figures and tables that present empirical results, check if you include:
 \begin{enumerate}
   \item The code, data, and instructions needed to reproduce the main experimental results (either in the supplemental material or as a URL). [Not Applicable]
   \item All the training details (e.g., data splits, hyperparameters, how they were chosen). [Not Applicable]
         \item A clear definition of the specific measure or statistics and error bars (e.g., with respect to the random seed after running experiments multiple times). [Not Applicable]
         \item A description of the computing infrastructure used. (e.g., type of GPUs, internal cluster, or cloud provider). [Not Applicable]
 \end{enumerate}

 \item If you are using existing assets (e.g., code, data, models) or curating/releasing new assets, check if you include:
 \begin{enumerate}
   \item Citations of the creator If your work uses existing assets. [Not Applicable]
   \item The license information of the assets, if applicable. [Not Applicable]
   \item New assets either in the supplemental material or as a URL, if applicable. [Not Applicable]
   \item Information about consent from data providers/curators. [Not Applicable]
   \item Discussion of sensible content if applicable, e.g., personally identifiable information or offensive content. [Not Applicable]
 \end{enumerate}

 \item If you used crowdsourcing or conducted research with human subjects, check if you include:
 \begin{enumerate}
   \item The full text of instructions given to participants and screenshots. [Not Applicable]
   \item Descriptions of potential participant risks, with links to Institutional Review Board (IRB) approvals if applicable. [Not Applicable]
   \item The estimated hourly wage paid to participants and the total amount spent on participant compensation. [Not Applicable]
 \end{enumerate}

 \end{enumerate}

\fi

\newpage
\onecolumn
\appendix

\section*{Appendix}

\paragraph{Roadmap.}

In Section~\ref{sec:preliminary}, we introduce the notations, basic definitions, and facts that we use. In Section~\ref{sec:exp_of_inner}, we present the background of the $\exp$ of the inner product kernel. In Section~\ref{sec:sketching}, we discuss the background of standard sketching. In Section~\ref{sec:high_sketching}, we introduce the background of the high precision sketching. 
In Section~\ref{sec:linear_regression}, we show the fast linear regression algorithm (see Algorithm~\ref{alg:iter_regression}), 
solving the regression problem containing one matrix, and analyze its properties, including its correctness and running time. 
In Section~\ref{sec:fast_attention}, we offer a new algorithm (see Algorithm~\ref{alg:AA^topAx_b_3}), 
which solves the regression problem containing three matrices, and analyzes its correctness and running time. In Section~\ref{sec:four_matrices}, we propose an algorithm (see Algorithm~\ref{alg:A^topAA^topAx_b_3}),  
which solves the regression problem containing four matrices, and analyzes its properties. In Section~\ref{sec:even}, we summarize and utilize the patterns of the previous sections to analyze the odd and even power regressions.
In Section~\ref{sec:attention_kernel}, we analyze the attention kernel.

\section{Preliminary}
\label{sec:preliminary}

In Section~\ref{sub:preliminary:notation}, we introduce the basic notations. In Section~\ref{sub:preliminary:definition}, we introduce some basic definitions and useful facts. In Section~\ref{sub:preliminary:attention}, we present some background about attention computation.

\subsection{Notations}
\label{sub:preliminary:notation}
We use $\R$ to denote the set of all real numbers. We use $\mathbb{Z}$ to denote the set of all integers and use $\mathbb{Z}_+$ to denote the set containing all positive integers. For any $n \in \mathbb{Z}_+$, we define $[n] := \{1, 2, 3, \dots, n \}$. 

For all $d \in \mathbb{Z}_+$, we use $\R^d$ to denote the set of all vectors with length $d$ and real entries and use $\mathbb{Z}_+^d$ to denote the set containing all vectors with length $d$ and entries of positive integers. For a vector $x \in \R^n$, we use $\|x\|_1$ to denote the $\ell_1$ norm, use $\|x\|_2$ to denote the $\ell_2$ norm, i.e., $\|x\|_2 = \left( \sum_{i = 1}^n x_i^2 \right)^\frac{1}{2}$, and use $\|x\|_\infty$ to denote the $\ell_\infty$ norm. 

For all $m,n \in \mathbb{Z}_+$, we use $\R^{m \times n}$ to denote the set containing all matrices with $m$ rows, $n$ columns, and real entries. For a matrix $A \in \R^{m \times n}$, we use $\|A\|$ to denote the spectral norm of $A$, i.e., $\|A\| = \max_{\|x\|_2 = 1} \|Ax\|_2$. We use $A^\top$ to denote the transpose of $A$. We use $\sigma_{\min}(A)$ to denote the minimum singular value of $A$, i.e., $\sigma_{\min}(A) = \min_{\|x\|_2 = 1} \|Ax\|_2$. Accordingly, we use $\sigma_{\max}(A)$ to denote the maximum singular value of $A$, so $\sigma_{\max}(A) = \|A\|$. Furthermore, we use $A^\dagger$ to denote the pseudo-inverse of $A$ and use $A^{-1}$ to denote the true inverse of $A$. The true inverse exists if $m = n$ and $\rank{(A)} = m$. For all $m_1, m_2, n \in \Z_+$, for all the matrices $A \in \R^{m_1 \times n}$ and $B \in \R^{m_2 \times n}$, we use $A \oplus B$ to denote the matrix $\begin{bmatrix} A \\ B \end{bmatrix}$. Correspondingly, for all $m_1, m_2, \ldots, m_p, p, n \in \Z_+$, for all the matrices $A_1 \in \R^{m_1 \times n}$, $A_2 \in \R^{m_2 \times n}$, $\cdots$, $A_p \in \R^{m_p \times n}$, we use $\oplus_{i = 1}^p A_i$ to denote $A_1 \oplus A_2 \oplus \dots \oplus A_p$.

We write $x = y\pm\epsilon$ if $x\in [y - \epsilon, y + \epsilon]$. ${\bf 1}_n$ represents an $n$-dimensional vector, whose entries are all $1$, and ${\bf 0}_n$ represents a $n$-dimensional vector, whose entries are all $0$. For a symmetric matrix $B\in\R^{n \times n}$, we say $B$ is positive semidefinite (denoted as $B \succeq 0$) if, for any vectors $x$ in $\R^n$, the inequality $x^\top B x \geq 0$ always holds. We also call $B$ a PSD matrix for simplicity.

\subsection{Definitions and Facts}
\label{sub:preliminary:definition}

In this section, we introduce the basic definitions and facts.

\begin{definition}\label{def:kappa}
We use $\kappa(A)$ to denote the condition number of $A$, i.e., 
\begin{align*} 
\kappa(A) := \sigma_{\max}(A)/\sigma_{\min}(A).
\end{align*}
\end{definition}

\begin{definition}[Hadamard matrix]\label{def:Hadamard_matrix}
    A Hadamard matrix is a square matrix of size $n$ with entries of either $1$ or $-1$, where each row of the matrix is orthogonal to every other row.
\end{definition}

\begin{fact}\label{fac:norm}
We have
\begin{itemize}
    \item For any matrix $A$ with full column rank, $\| (A^\top A)^\dagger \| = \sigma_{\min}(A)^{-2}$.
    \item  For any matrix $A$, $\| A^\top A \| = \sigma_{\max}(A)^2$.
    \item  For any matrix $A$ with full column rank and any positive integer $k$, we have $\| ( (A^\top A)^k)^\dagger \| \leq \sigma_{\min}(A)^{-2k}$.
    \item For any orthonormal column basis $U \in \R^{n \times d}$ ($n \geq d$), we have $\| U x \|_2 = \|x \|_2$.
    \item For any matrix $A$, we have $\| A x \| \leq \| A \| \cdot \|x \|_2 = \sigma_{\max}(A) \cdot \| x \|_2$.
    \item For any matrix $A$, we have $\| A x \|_2 \geq \sigma_{\min}(A) \cdot \| x \|_2$. 
    \item For any square invertible matrix $A$, $\kappa(A) = \kappa(A^\dagger)$.
    \item For any matrix $A$ with full column rank and any matrix $B$ of compatible size, $\kappa(B) \leq \kappa(AB) \cdot \kappa(A)$.
    \item For any matrices $A,B$, $\kappa(AB) \leq \kappa(A) \cdot \kappa(B)$.
\end{itemize}

\end{fact}
\begin{proof}
We prove the eighth item. Since $A$ has full column rank, $\sigma_{\min}(A) > 0$, and for every unit vector $x$ we have $\sigma_{\min}(A) \|Bx\|_2 \leq \|ABx\|_2 \leq \sigma_{\max}(A) \|Bx\|_2$. Taking the maximum and the minimum over unit vectors $x$ gives $\sigma_{\max}(B) \leq \sigma_{\max}(AB)/\sigma_{\min}(A)$ and $\sigma_{\min}(B) \geq \sigma_{\min}(AB)/\sigma_{\max}(A)$. Dividing the two bounds gives $\kappa(B) \leq \kappa(AB) \cdot \kappa(A)$. The other items are standard.
\end{proof}

\begin{definition}[Tensor Product]\label{def:vector_tensor_product}
Let $x \in \R^n$ and $y \in \R^m$. We use $x \otimes y$ to denote the tensor product of $x$ and $y$, and it is defined as
\begin{align*}
    x \otimes y := \mathrm{vec}(xy^\top).
\end{align*}
We use $x^{\otimes p}$ to represent $x$ tensoring with itself for $p$ times.
\end{definition}

\subsection{Attention Backgrounds}
\label{sub:preliminary:attention}

In this section, we introduce the important background of attention mechanism and matrix. Given matrices $Q, K , V \in \R^{d \times d}$ as weights, and $X \in \R^{n \times d}$ which be viewed as the embedding of a length-$n$ sentence where each length-$d$ vector is corresponding to a word. The attention matrix is defined as
\begin{align}\label{eq:attention}
 D(X)^{-1} \exp(X Q K^\top X^\top) X V
\end{align}
where $D(X) = \mathrm{diag} ( \exp( X Q K^\top X^\top ) {\bf 1}_n )$ and $\exp(\cdot)$ is applied entrywise.

\begin{figure}[!ht]
    \centering
    \includegraphics[width = \linewidth]{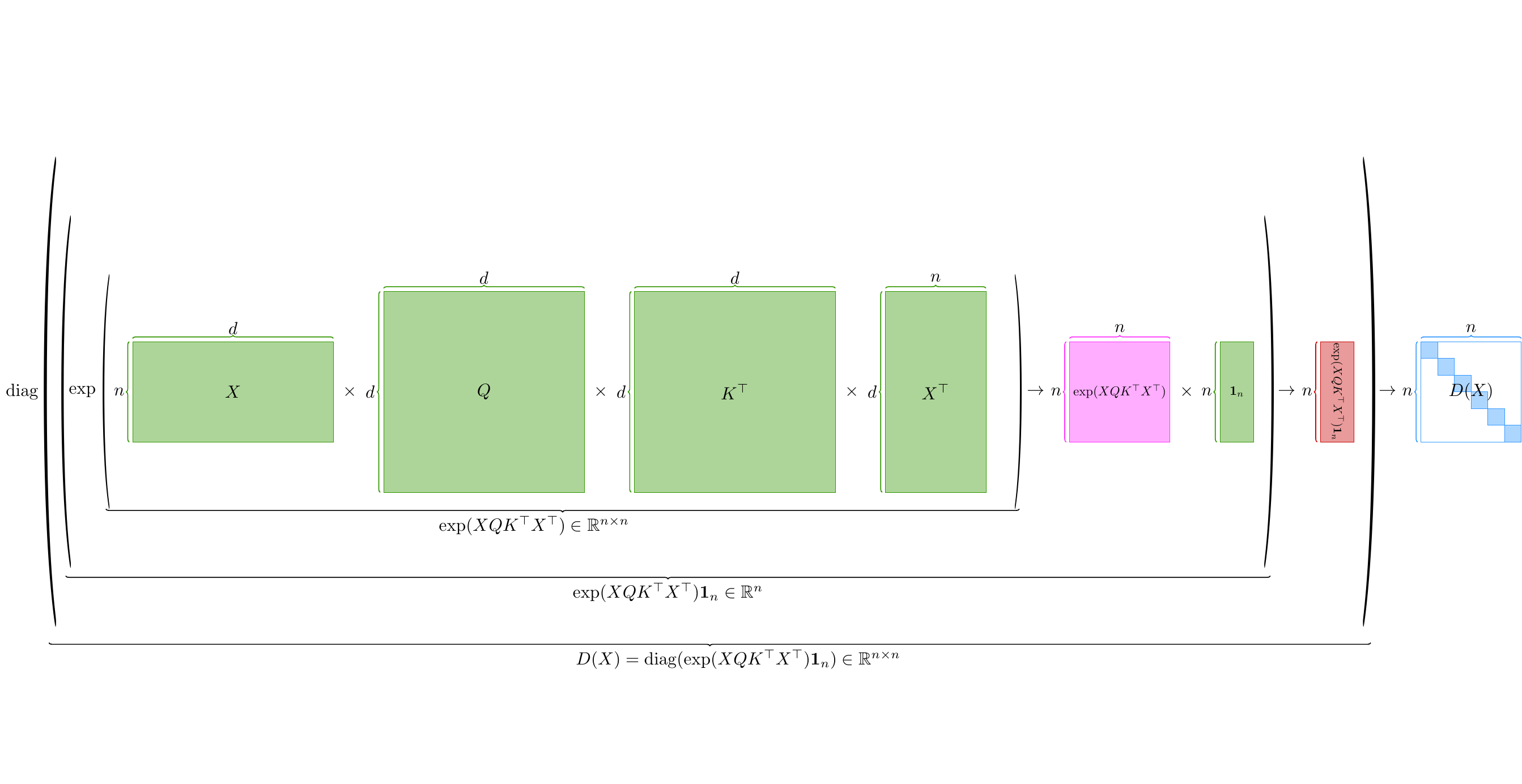}
    \caption{The visualization of the matrix $D(X) \in \R^{n \times n}$. Given $Q, K , V \in \R^{d \times d}$ and $X \in \R^{n \times d}$, we first compute $XQK^\top X^\top \in \R^{n \times n}$. Then, we find $\exp(XQK^\top X^\top) \in \R^{n \times n}$. After that, we multiply $\exp(XQK^\top X^\top) \in \R^{n \times n}$ with the vector ${\bf 1}_n \in \R^{n}$. Finally, we use $\diag(\cdot)$ to transform $\exp(XQK^\top X^\top) {\bf 1}_n \in \R^n$ into a diagonal matrix, which is $D(X) \in \R^{n \times n}$. In this figure, green matrices/vectors represent the terms that are given; the purple matrix represents the term after one operation; the red vector represents the term after two operations; the blue matrix represents the term after three operations.}
    \label{fig:DX}
\end{figure}

\begin{figure}[!ht]
    \centering
    \includegraphics[width = \linewidth]{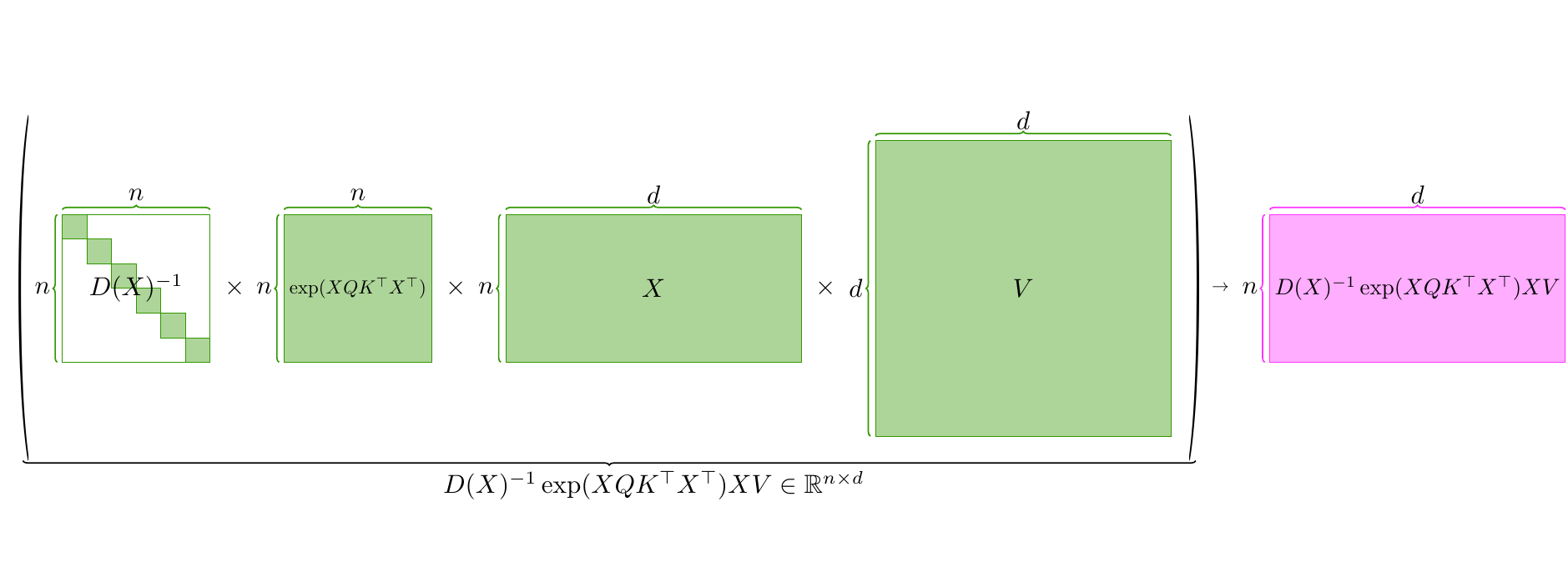}
    \caption{The visualization of the attention computation (see Eq.~\eqref{eq:attention}). Since we present the visualization of how we get $D(X) \in \R^{n \times n}$ and $\exp(XQK^\top X^\top) \in \R^{n \times n}$ in Figure~\ref{fig:DX}, we regard them as given. Moreover, we are also given $V \in \R^{d \times d}$ and $X \in \R^{n \times d}$. We compute their product, namely $D(X)^{-1} \exp(X Q K^\top X^\top) X V$. In this figure, green matrices represent the terms that are given, and the purple matrix represents the term after one operation.}
    \label{fig:attention_computation}
\end{figure}

In \cite{as23,bsz23}, they simplify the computation by treating $XQ$, $XK$, and $X V$ as $Q,K,V \in \R^{n \times d}$ so they obtain
\begin{align}\label{eq:attention_in_AS23_BSZ23}
    D^{-1} \exp (Q K^\top) V
\end{align}
where $D = \mathrm{diag}( \exp(Q K^\top) {\bf 1}_n )$.

\begin{figure}[!ht]
    \centering
    \includegraphics[width = \linewidth]{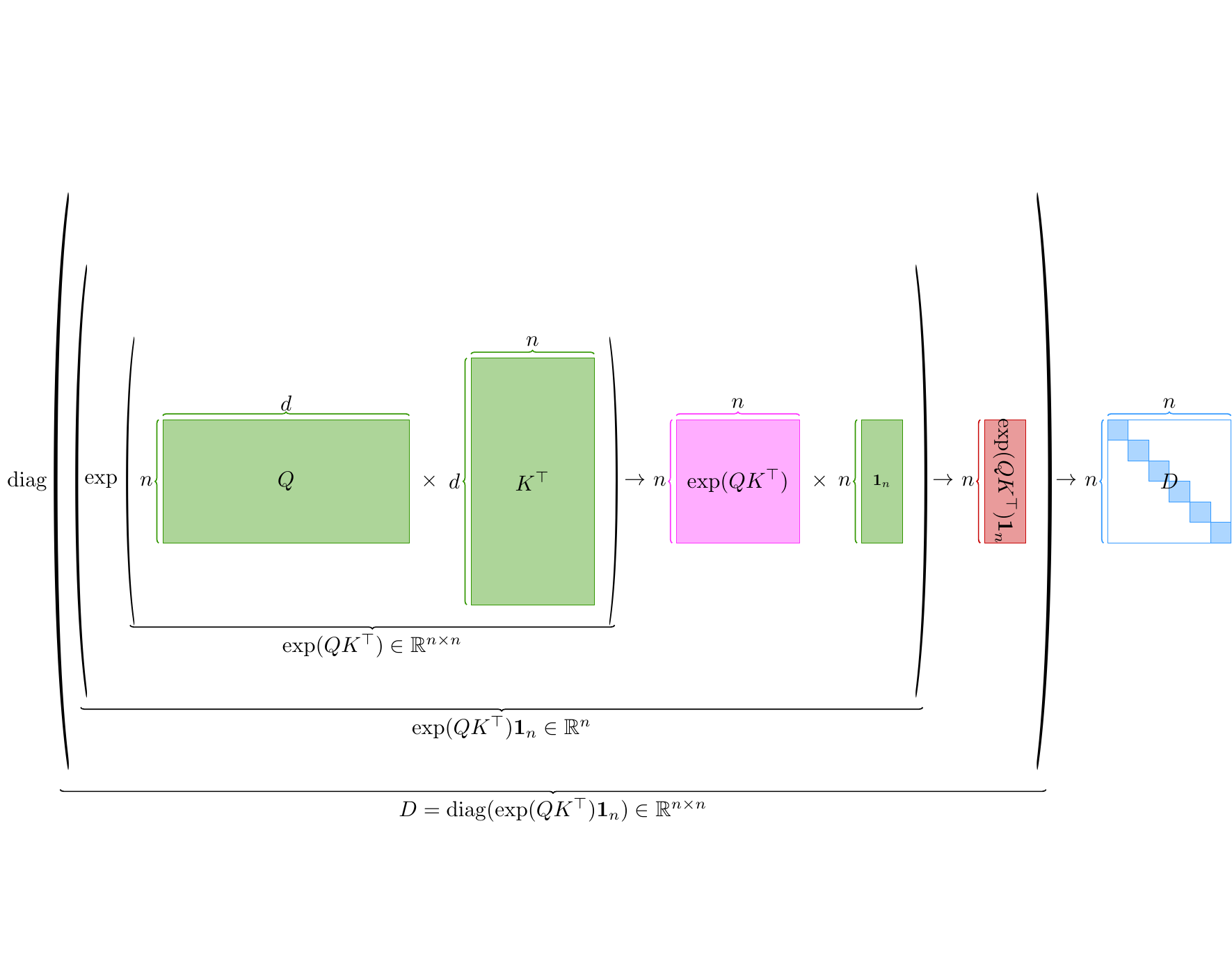}
    \caption{The visualization of the matrix $D \in \R^{n \times n}$. Given $Q, K, V \in \R^{n \times d}$, we first compute $QK^\top \in \R^{n \times n}$. Then, we find $\exp(QK^\top) \in \R^{n \times n}$. After that, we multiply $\exp(QK^\top) \in \R^{n \times n}$ with the vector ${\bf 1}_n \in \R^n$. Finally, we use $\diag(\cdot)$ to transform $\exp(QK^\top) {\bf 1}_n \in \R^n$ into a diagonal matrix, which is $D \in \R^{n \times n}$. In this figure, green matrices/vectors represent the terms that are given; the purple matrix represents the term after one operation; the red vector represents the term after two operations; the blue matrix represents the term after three operations.}
    \label{fig:D}
\end{figure}

\begin{figure}[!ht]
    \centering
    \includegraphics[width = \linewidth]{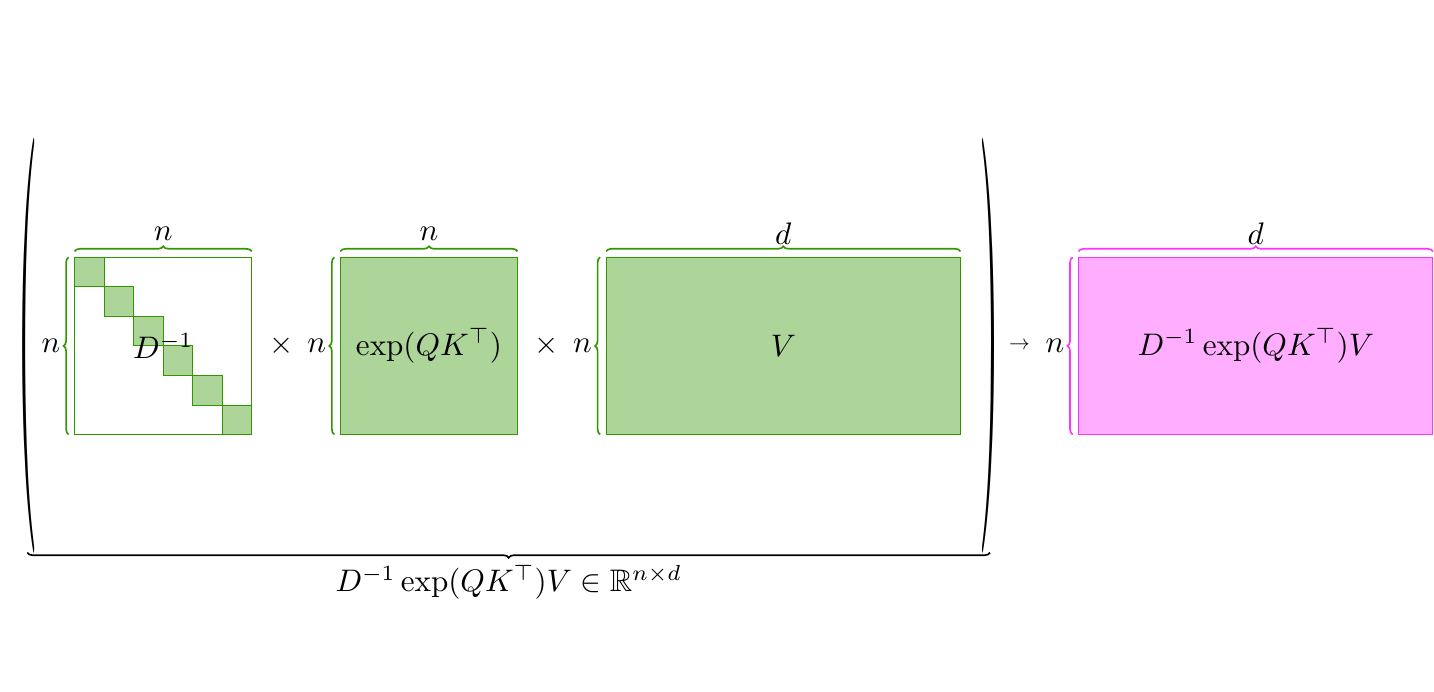}
    \caption{The visualization of the simplified version of attention computation in \cite{as23,bsz23} (see Eq.~\eqref{eq:attention_in_AS23_BSZ23}). Since we present the visualization of how we get $D \in \R^{n \times n}$ and $\exp(QK^\top) \in \R^{n \times n}$ in Figure~\ref{fig:D}, we regard them as given. Moreover, we are also given $V \in \R^{n \times d}$. We compute their product, namely $D^{-1} \exp(Q K^\top) V  \in \R^{n \times d}$. In this figure, green matrices represent the terms that are given, and the purple matrix represents the term after one operation.}
    \label{fig:attention_computation_AS23_BSZ23}
\end{figure}

In \cite{gsyz23_quantum}, they simplify the attention by treating $V$ as identity, namely
\begin{align}\label{eq:attention_in_gsyz23a}
    D^{-1} \exp (Q K^\top),
\end{align}
where $D, Q, K$ are defined same as above.

\begin{figure}[!ht]
    \centering
    \includegraphics[width = 0.7\linewidth]{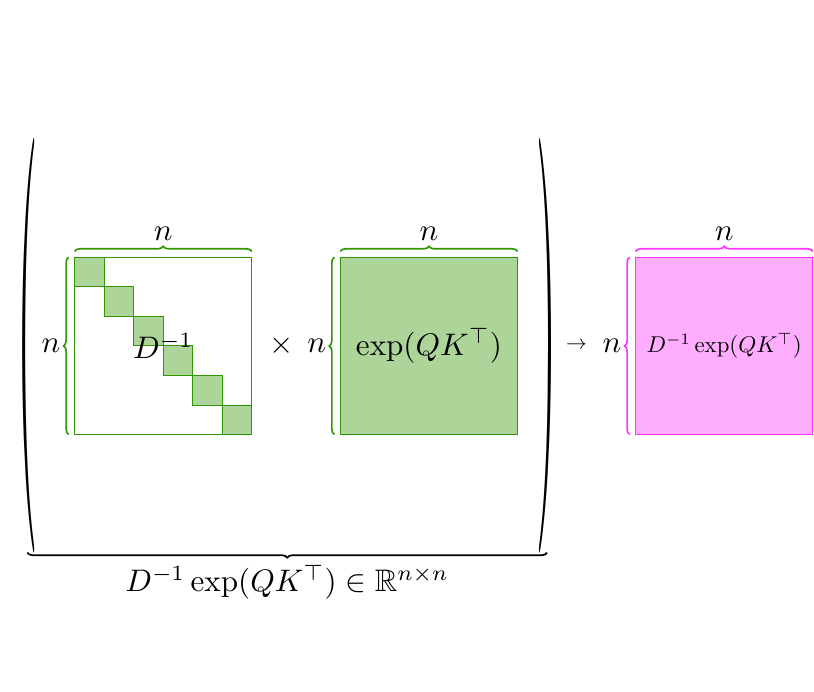}
    \caption{The visualization of the simplified version of attention computation in \cite{gsyz23_quantum} (see Eq.~\eqref{eq:attention_in_gsyz23a}). Since we present the visualization of how we get $D \in \R^{n \times n}$ and $\exp(QK^\top) \in \R^{n \times n}$ in Figure~\ref{fig:D}, we regard them as given. We compute their product, namely $D^{-1} \exp(Q K^\top) \in \R^{n \times n}$. In this figure, green matrices represent the terms that are given, and the purple matrix represents the term after one operation.}
    \label{fig:attention_computation_gsyz23}
\end{figure}

In addition, in \cite{dms23}, the attention is simplified into the form of 
\begin{align}\label{eq:attention_in_dms23}
    D^{-1} \exp (K K^\top),
\end{align}
where $D, Q, K$, for $(K = Q)$, are defined same as above.

\begin{figure}[!ht]
    \centering
    \includegraphics[width = \linewidth]{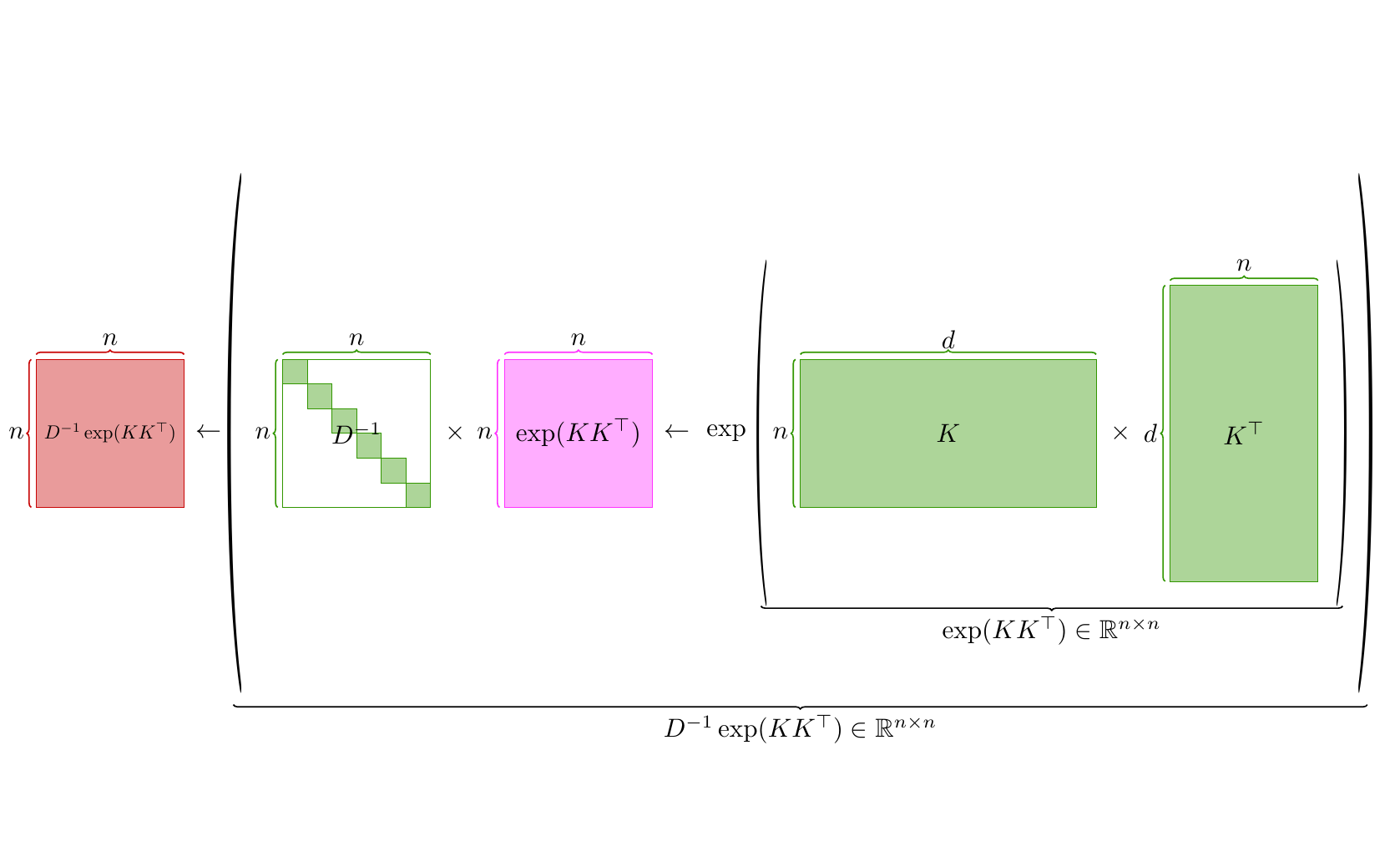}
    \caption{The visualization of the simplified version of attention computation in \cite{dms23} (see Eq.~\eqref{eq:attention_in_dms23}). Since we present the visualization of how we get $D \in \R^{n \times n}$ in Figure~\ref{fig:D}, we regard it as given. We are also given that $K \in \R^{n \times d}$, which implies that $K^\top \in \R^{d \times n}$. First, we compute their product, namely $KK^\top \in \R^{n \times n}$. Then, we find $\exp(KK^\top) \in \R^{n \times n}$. Finally, we multiply $D^{-1} \in \R^{n \times n}$ with $\exp(KK^\top) \in \R^{n \times n}$, which gives us $D^{-1} \exp(KK^\top) \in \R^{n \times n}$. In this figure, green matrices represent the terms that are given; the purple matrix represents the term after one operation; the red matrix represents the term after two operations.}
    \label{fig:attention_computation_DMS23}
\end{figure}

In this work, we provide a simplification of Eq.~\eqref{eq:attention} from a different perspective, by ignoring the factor of $D^{-1}$ and $\exp$, so that we can get
\begin{align*}
    X Q K^\top X^\top X V
\end{align*}
Further, we merge $QK^\top$ into one matrix $W$ and consider one column of $V$ a time,
\begin{align}\label{eq:our_attention}
    X W X^\top X v
\end{align}
where $v$ is a column of $V$.

\begin{figure}[!ht]
    \centering
    \includegraphics[width = \linewidth]{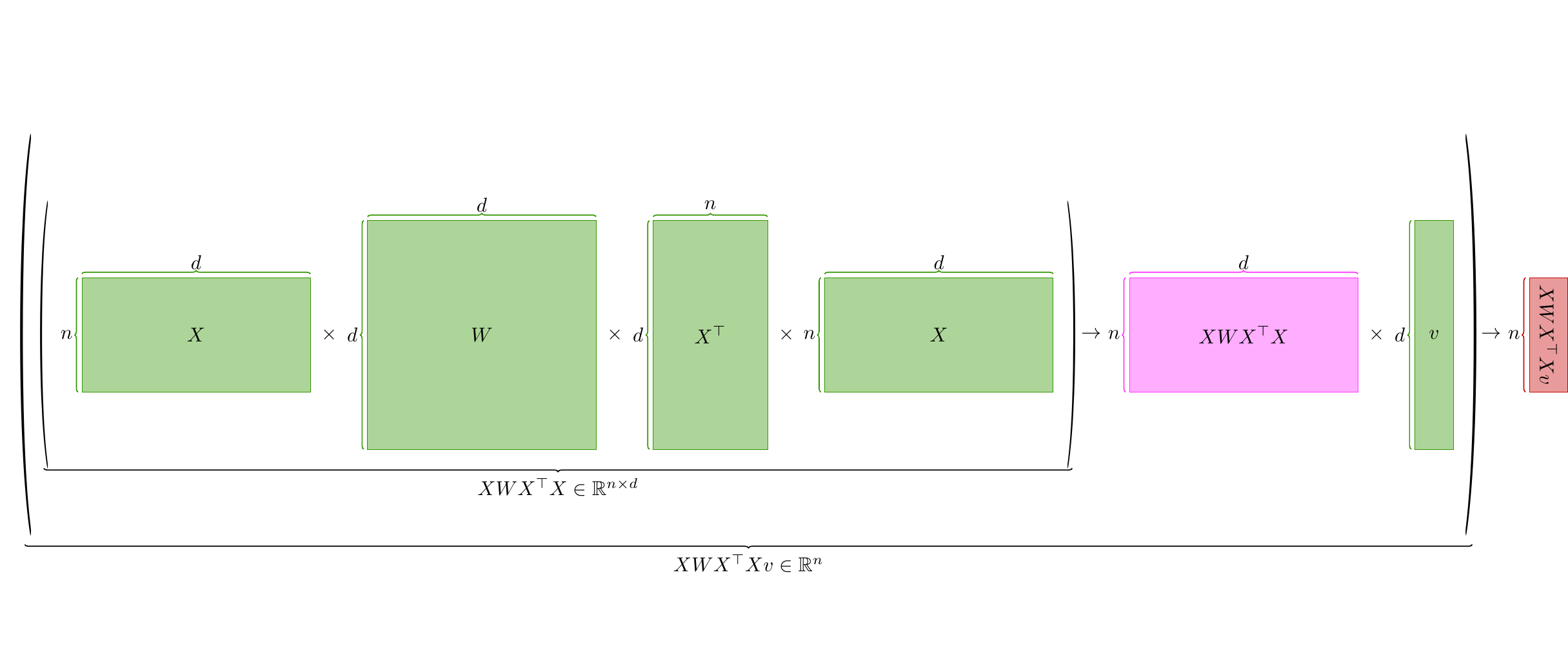}
    \caption{The visualization of the simplified version of attention computation that we analyze in this paper (see Eq.~\eqref{eq:our_attention}). We are given that $X \in \R^{n \times d}$, $W \in \R^{d \times d}$, and $v \in \R^d$. First, we compute the product of the matrices, namely $X W X^\top X \in \R^{n \times d}$. Then, we multiply $X W X^\top X \in \R^{n \times d}$ with the vector $v \in \R^d$, which gives us $X W X^\top Xv \in \R^{n}$ In this figure, green matrices represent the terms that are given; the purple matrix represents the term after one operation; the red matrix represents the term after two operations.}
    \label{fig:attention_computation_our}
\end{figure}

Thus, we can obtain the following definition of attention formulation.

\begin{definition}\label{def:X_W_X_X_v}
Given $X \in \R^{n \times d}$, $W \in \R^{d \times d}$ and $ y \in \R^n$, the goal is to solve the following regression problem
\begin{align*}
\min_{v \in \R^d} \| X W X^\top X \cdot v - y \|_2^2 
\end{align*}
\end{definition}

When $W$ is symmetric positive definite, we show that it is in fact equivalent to the odd power regression of degree 3 with the data matrix $XU$, where $W = UU^\top$, defined as follows.

\begin{definition}\label{def:X_X_X_v}
Given $X \in \R^{n \times d}$, $y \in \R^n$. The goal is to solve
\begin{align*}
\min_{v \in \R^d} \| X X^\top X v - y \|_2^2
\end{align*}
\end{definition}

The equivalence can be established as such.

\begin{lemma}
Let $X \in \R^{n \times d}$ and $y \in \R^n$. Let $W \in \R^{d \times d}$ be symmetric positive definite with Cholesky factorization $W = U U^\top$, where $U \in \R^{d \times d}$ is invertible. Let $\wt{X} := X U$, and for every $v \in \R^d$ let $\wt{v} := U^{-1} v$. Then for every $v \in \R^d$,
\begin{align*}
\| X W X^\top X v - y \|_2 = \| \wt{X} \wt{X}^\top \wt{X} \wt{v} - y \|_2.
\end{align*}
Hence problem~\ref{def:X_W_X_X_v} with $X$ and problem~\ref{def:X_X_X_v} with $\wt{X}$ have the same optimal value, every $v$ has the same cost in the first problem as $\wt{v} = U^{-1} v$ has in the second, and $v$ is optimal for the first if and only if $\wt{v}$ is optimal for the second.
\end{lemma}
\begin{proof}
Since $W = UU^\top$ and $UU^{-1} = I$, we have
\begin{align*}
X W X^\top X v = X U U^\top X^\top X U U^{-1} v = (XU)(XU)^\top (XU) (U^{-1} v) = \wt{X} \wt{X}^\top \wt{X} \wt{v},
\end{align*}
which gives the identity of the two residuals. Since $U$ is invertible, the map $v \mapsto U^{-1} v$ is a bijection of $\R^d$. So the two objective functions take the same values, the optimal values agree, and solutions map to solutions. In particular, if an algorithm for the odd power regression of degree 3 on $\wt{X}$ outputs $\wt{v}'$ with $\| \wt{X} \wt{X}^\top \wt{X} \wt{v}' - y \|_2 \leq (1+\epsilon) \OPT + \epsilon \|y\|_2$, where $\OPT$ is the common optimal value, then $v' := U \wt{v}'$ satisfies $\| X W X^\top X v' - y \|_2 \leq (1+\epsilon) \OPT + \epsilon \|y\|_2$.
\end{proof}

To keep the running time, the algorithm never forms $\wt{X} = XU$. It computes $U$ in $O(d^3)$ time, applies sketches as $(SX)U$, and computes matrix-vector products as $X(Uz)$ and $U^\top (X^\top w)$. To apply Algorithm~\ref{alg:AA^topAx_b_3} (or Theorem~\ref{thm:odd} with $j=1$) to $\wt{X}$, the matrix $X$ must have full column rank (then so does $\wt{X}$), and the input must satisfy $\kappa \geq \kappa(\wt{X})$. By the last item of Fact~\ref{fac:norm}, $\kappa(\wt{X}) \leq \kappa(X)\kappa(U) = \kappa(X)\sqrt{\lambda_{\max}(W)/\lambda_{\min}(W)}$. Alternatively, $\kappa(\wt{X})$ can be estimated from $(SX)U$ as in Section~\ref{sec:fast_attention}.

The lemma assumes that $W$ is symmetric positive definite, so it does not cover a general $W = QK^\top$, which need not be symmetric or invertible. Also, the matrix $W$ cannot simply be dropped: problem~\ref{def:X_W_X_X_v} with $X$ need not have the same optimal value as problem~\ref{def:X_X_X_v} with the same $X$. For example, let $n = d = 2$, $X = I_2$, $W = e_1 e_1^\top$ (which equals $QK^\top$ for $Q = K = e_1 e_1^\top$) and $y = e_2$. Then $X W X^\top X v = v_1 e_1$, so $\| X W X^\top X v - y \|_2^2 = v_1^2 + 1$, and the optimal value of problem~\ref{def:X_W_X_X_v} is $1$. On the other hand, $X X^\top X v = v$, so the optimal value of problem~\ref{def:X_X_X_v} is $0$, attained at $v = e_2$.

\section{Preliminary About Exponential of Inner Product Kernel}
\label{sec:exp_of_inner}

In Section~\ref{sub:exp_of_inner:attention}, we provide a formal definition of the attention kernel. In Section~\ref{sub:exp_of_inner:kernel}, we analyze the properties of the attention kernel.

\subsection{Definition of Attention Kernel}
\label{sub:exp_of_inner:attention}

Here, we start to present the definitions of the Gaussian Kernel and the Attention Kernel. We will focus on $\exp(\langle \cdot,\cdot \rangle)$ inner product. It is similar to the Gaussian kernel, let us first review the definition of the Gaussian Kernel

\begin{definition}[Gaussian Kernel
]\label{def:gaussian_kernel}
Let $X \in \R^{d \times n}$. Let $x_i$ be the $i$-th column of $X$. We say $G$ is the Gaussian kernel if for any two points $x, y\in \R^d$,
\begin{align*}
    G(x, y) = \exp(-\|x - y\|_2^2/2).
\end{align*}
We say $G$ is the Gaussian kernel on $X$ if
\begin{align*}
    G_{i,j} = \exp(-\|x_i - x_j\|_2^2/2).
\end{align*}
Here we abuse the notation and let $G$ be an $n\times n$ real matrix.
\end{definition}

We define the attention kernel as follows
\begin{definition}[Attention Kernel]\label{def:attention_kernel}
Let $X \in \R^{d \times n}$. Let $x_i$ be the $i$-th column of $X$. We say $G$ is the attention kernel if for any two points $x, y\in \R^d$,
\begin{align*}
    G(x, y) = \exp(\langle x,y\rangle).
\end{align*}
We say $G$ is the attention kernel on $X$ if
\begin{align*}
    G_{i,j} = \exp(\langle x_i, x_j\rangle).
\end{align*}
Here we abuse the notation and let $G$ be an $n\times n$ real matrix.
\end{definition}

\subsection{Properties of Attention Kernel}
\label{sub:exp_of_inner:kernel}

After we define the attention kernel, we start to analyze its properties.

\begin{fact}\label{fac:psd_diagonal_vs_off_diagonal}

Let $B$ be a PSD matrix in $\R^{n \times n}$.

Then we have 
\begin{align*}
B_{i,i} B_{j,j} \geq B_{i,j}^2, ~~~\forall i,j \in [n] \times [n]
\end{align*}

\end{fact}
\begin{proof}

Let $x = a \cdot e_i + b \cdot e_j$. 

For all arbitrary $a, b \in \R$, we can get
\begin{align}\label{eq:xT_B_x}
0 \leq & ~ x^\top B x \notag\\
= & ~ \begin{bmatrix} a & b \end{bmatrix} \begin{bmatrix} 
B_{i,i}  & B_{i,j}  \\
B_{j,i} & B_{j,j}
\end{bmatrix}
\begin{bmatrix} a \\ b \end{bmatrix},
\end{align}
where the first step follows from the fact that $B$ is a PSD matrix and the second step follows from expanding the equation.

Note that for all arbitrary $a, b \in \R$, Eq.~\eqref{eq:xT_B_x} holds.

Therefore,
\begin{align*}
\det \left(  \begin{bmatrix} 
B_{i,i}  & B_{i,j}  \\
B_{j,i} & B_{j,j}
\end{bmatrix} \right) \geq 0
\end{align*}
which is equivalent to say $B_{i,i} B_{j,j} \geq B_{i,j}^2$.
\end{proof}

\begin{lemma}
Let $X\in \R^{n\times d}$. We define attention kernel
\begin{align*}
    A:=\exp(XX^\top)\in \R^{n\times n}
\end{align*}
where $\exp(\cdot)$ is applied entrywise. Let $\epsilon \in (0,0.1)$, $r > 0$ and a matrix $B\in \R^{n\times n}$ satisfy the following conditions
\begin{itemize}
    \item Condition 1. $ \epsilon \leq  \frac{1}{4} \exp(-4r).$
    \item Condition 2. $A_{i,j}\in [\exp(-r),\exp(r)]$ for $i,j\in [n]\times [n]$.
    \item Condition 3. $(1-\epsilon)\cdot A\preceq B\preceq (1+\epsilon)\cdot A$.
\end{itemize} 

Then, we have
\begin{align*}
    B_{i,j} \in [(1-\sqrt{\epsilon})\exp(-r),(1+\epsilon)\exp(r)].
\end{align*} 
\end{lemma}

\begin{proof}
 
$B-(1-\epsilon)\cdot A$ is a PSD matrix.

Thus, we have
\begin{align}\label{eq:Bij_1_epsilon_Aij}
    |B_{i,j}-(1-\epsilon)\cdot A_{i,j}|\leq & ~ \sqrt{(B_{i,i}-(1-\epsilon)\cdot A_{i,i})(B_{j,j}-(1-\epsilon)\cdot A_{j,j})} \notag\\
    \leq & ~ 2\epsilon \sqrt{A_{i,i}A_{j,j}} \notag\\
    \leq & ~ 2\epsilon\exp(r)
\end{align}
where the 1st step is by Lemma~\ref{fac:psd_diagonal_vs_off_diagonal}, the 2nd step is due to $B_{i,i} \leq (1 + \epsilon) A_{i,i}$ (By condition 3 in lemma statement), and the 3rd step is because of the definition of $A_{i, j}$ from the lemma statement.

Based on the second condition from the lemma statement, we have
\begin{align}\label{eq:1_epsilon_Aij}
    (1-\epsilon)\cdot A_{i,j} \in & ~ [(1-\epsilon)\cdot \exp(-r), (1-\epsilon)\cdot \exp(r)],
\end{align}

Combining Eq.~\eqref{eq:Bij_1_epsilon_Aij} and Eq.~\eqref{eq:1_epsilon_Aij}, we have
\begin{align}\label{eq:range_1}
    B_{i,j} \in & ~ [(1-\epsilon)\exp(-r)-2\epsilon \exp(r), (1+\epsilon)\exp(r)]
\end{align}

By using $(1+\epsilon)\cdot A-B$ is a PSD matrix, we may do a symmetric argument as follows:
\begin{align}\label{eq:range_2}
    B_{i,j} \in & ~ [(1+\epsilon)\exp(-r)-2\epsilon \exp(r),(1+3\epsilon)\exp(r)].
\end{align}

The intersection of Eq.~\eqref{eq:range_1} and Eq.~\eqref{eq:range_2} gives us:
\begin{align*}
    B_{i,j} \in & ~ [(1+\epsilon)\exp(-r)-2\epsilon \exp(r),(1+\epsilon)\exp(r)]
\end{align*}

We know that
\begin{align*}
(1+\epsilon)\exp(-r)-2\epsilon \exp(r) 
\geq & ~ \exp(-r) - 2\epsilon \exp(r)  \\
= & ~ \exp(-r)-2 \sqrt{\epsilon} \cdot \sqrt{\epsilon} \cdot \exp(r)  \\
\geq & ~ \exp(-r) - 2 \cdot \sqrt{\epsilon} \cdot \frac{1}{2} \exp(-2r) \cdot \exp(r) \\
= & ~ \exp(-r) - \sqrt{\epsilon} \exp(-r) \\
= & ~ ( 1- \sqrt{\epsilon}) \cdot \exp(-r)
\end{align*} 
where the first step follows from $\epsilon \geq 0$, the second step follows from simple algebra, the third step follows from $\sqrt{\epsilon} \leq \frac{1}{2} \exp(-2r)$ (by condition 1 in lemma statement), the fourth step follows from simple algebra, and the last step follows from simple algebra.

Thus, we have
\begin{align*}
    B_{i,j} \in & ~ [(1- \sqrt{\epsilon} )\exp(-r),(1+\epsilon)\exp(r)].
\end{align*}
This completes the proof.
\end{proof}

\section{Preliminary About Standard Sketching}
\label{sec:sketching}

In Section~\ref{sub:sketching:matrices}, we provide two types of fast sketching matrices. In Section~\ref{sub:preliminary:subspace}, we introduce the formal definition of subspace embedding. In Section~\ref{sub:preliminary:subsample_embedding}, we interpret and analyze the property of subspace embedding by different matrices. In Section~\ref{sub:sketching:product}, we provide the definition of the Frobenius norm approximate matrix product.

\subsection{Sketching Matrices}
\label{sub:sketching:matrices}

We define subsampled randomized Hadamard transform ($\mathsf{SRHT}$) as follows.

\begin{definition}[Subsampled Randomized Hadamard Transform ($\mathsf{SRHT}$), see \cite{ldfu13,w14}]\label{def:SRHT}

    Let $H$ be a Hadamard matrix in $\R^{d \times d}$ (see Definition~\ref{def:Hadamard_matrix}). Let $D$ be a diagonal matrix in $\R^{d \times d}$, which satisfies that each diagonal entry is either $-1$ or $1$ with the same probability. Let $P$ be a matrix in $\{0,1\}^{m\times d}$, where each row of $P$ contains only one $1$ at a random entry. We define the matrix $S \in \R^{m\times d}$ as
    \begin{align*}
        S := \frac{1}{\sqrt{m}}PHD
    \end{align*}
    and call $S$ the $\mathsf{SRHT}$ matrix.
\end{definition}

We define $\mathsf{TensorSRHT}$ as follows:

\begin{definition}[Tensor Subsampled Randomized Hadamard Transform ($\mathsf{TensorSRHT}$), Definition 2.9 in \cite{swyz21}]\label{def:tensor_SRHT}

Let $P$ be a matrix in $\{0,1\}^{m\times d}$, where each row of $P$ contains only one $1$ at a random entry. $P$ can be regarded as the sampling matrix. Let $H$ be the Hadamard matrix in $\R^{d \times d}$ (see Definition~\ref{def:Hadamard_matrix}). Let $D_1$ and $D_2$ be diagonal matrices in $\R^{d \times d}$, which satisfy that each diagonal entry is either $-1$ or $1$ with the same probability. $D_1$ and $D_2$ are independent. Then, we define $\mathsf{TensorSRHT}$ as a function $S : \R^d \times \R^d \to \R^m$, which is defined as
\begin{align*}
    S := \frac{1}{\sqrt{m}}P \cdot (HD_1 \otimes HD_2).
\end{align*}
\end{definition}

\subsection{Subspace Embedding}
\label{sub:preliminary:subspace}

We define subspace embedding as follows.

\begin{definition}[Subspace Embedding $\mathsf{SE}(n,d,\epsilon,\delta)$,~\cite{s06}]\label{def:se}
Given a matrix $A \in \R^{n \times d}$, we say $S$ is an $(n,d,\epsilon,\delta)$ subspace embedding, if 
\begin{align*}
\Pr[ \forall x \in \R^d:~ (1-\epsilon) \cdot \| A x \|_2 \leq \|S A x \|_2 \leq (1+\epsilon) \cdot \| A x \|_2 ] \geq  & ~ 1-\delta.
\end{align*}
This holds in particular if
\begin{align*}
    \Pr[(1 - \epsilon) \cdot (A^\top A ) \preceq A^\top S^\top S A   \preceq (1 + \epsilon) \cdot (A^\top A )] \geq & ~ 1-\delta.
\end{align*}
\end{definition}

\subsection{Subspace Embedding by Different Sketches}
\label{sub:preliminary:subsample_embedding}

We specify several classes of subspace embedding that obtain a small sketching dimension and can be applied to $A$ fast.
\begin{lemma}\label{lem:subspace_embedding}
    Given a matrix $A \in \R^{n \times d}$, $S$ is an $(n,d,\epsilon,\delta)$ subspace embedding for
    \begin{itemize}
        \item Let $S\in \R^{m \times n}$ denote a $\mathsf{SRHT}$ matrix with $m=O(\epsilon^{-2} d \log(n/\delta))$ rows. In addition, $S A$ can be computed in $O(nd\log(n/\delta))$ time.
        \item Let $S \in \R^{m \times n}$ denote $\mathsf{OSNAP}$ matrix with $m = O(\epsilon^{-2} d\log(d/\delta))$ rows and column sparsity $s=O(\epsilon^{-1}\log(d/\delta))$. In addition, $SA$ can be computed in $O(s\cdot \nnz(A))$ time.
    \end{itemize}
\end{lemma}

\subsection{Approximate Matrix Product}
\label{sub:sketching:product}

Here, we define the Frobenius Norm Approximate Matrix Product as follows.

\begin{definition}[Frobenius Norm Approximate Matrix Product]
\label{def:famp}
    Let $A \in \R^{n \times d_1}, B \in \R^{n\times d_2}$ be two matrices. We say $S$ is Frobenius Norm Approximate Matrix Product $\mathsf{FAMP}(n,\epsilon,\delta)$ with respect to $A,B$ if
    \begin{align*}
        \Pr[ \| A^\top B - A^\top S^\top S B \|_F \leq \epsilon \| A \|_F \| B \|_F ] \geq 1-\delta.
    \end{align*}
\end{definition}

\section{Preliminary About High Precision Sketching}
\label{sec:high_sketching}

In Section~\ref{sub:preliminary:facts}, we explain more useful facts regarding matrix spectrum. 
In Section~\ref{sub:preliminary:well_condition}, we analyze the property of well-conditioned PSD regression. In Section~\ref{sub:preliminary:forward_error_simple}, we analyze the forward error for overconstrained systems. In Section~\ref{sub:preliminary:forward_error_PSD}, we study the forward error for PSD matrices.

\subsection{Norm Preservation and Spectrum}
\label{sub:preliminary:facts}

\begin{fact}\label{fac:subspace_equiv}
The following two conditions are equivalent:
\begin{itemize}
    \item For all unit vector $x$, $1-\epsilon \leq \| A x \|_2^2 \leq 1+\epsilon$;
    \item $\| A^\top A  - I \| \leq \epsilon$.
\end{itemize}
\end{fact}
\begin{proof}

We know that, 
\begin{align*}
\| A^\top A  - I \| \leq \epsilon
\end{align*}
is equivalent to
\begin{align*}
-\epsilon \leq x^\top ( A^\top A - I ) x \leq \epsilon, ~~~ \forall \| x \|_2=1
\end{align*}
move $-x^\top x$ to both sides we have
\begin{align*}
1 - \epsilon \leq x^\top A^\top A x \leq 1+\epsilon, ~~~\forall \| x \|_2=1
\end{align*}
which is equivalent to
\begin{align*}
1-\epsilon \leq \| A x \|_2^2 \leq 1+\epsilon, ~~~\forall \| x \|_2=1.
\end{align*}
\end{proof}

\subsection{Well-Conditioned PSD Regression}
\label{sub:preliminary:well_condition}

We state a lemma for solving well-conditioned PSD regression, due to~\cite{bpsw21}.
 
\begin{lemma}[Well-conditioned PSD regression, Lemma B.2 in \cite{bpsw21}]\label{lem:well_condition}
     Consider the regression problem
     \begin{align*}
         \min_{x \in \R^d} \|Bx - y\|_2^2.
     \end{align*}

     Suppose $B \in \R^{d \times d}$ is a PSD matrix with
     \begin{align*}
         \frac{3}{4} \leq \|Bx\|_2 \leq \frac{5}{4}, ~~~~\forall x \text{~such~that~}\| x \|_2 = 1.
     \end{align*}

     Using gradient descent update
     \begin{align*}
        x_{t+1} = x_t - B^\top ( B x_t - y ).
     \end{align*}
     Then, after $t$ iterations, we obtain
     \begin{align*}
         \|B(x_t - x^*)\|_2 \leq c^t \|B (x_0 - x^*)\|_2
     \end{align*}
     for some constant $c \in (0, 9/16]$.
\end{lemma}
\begin{proof}
     The gradient at time $t$ is $B^\top (B x_t - y)$ and 
     \begin{align}\label{eq:x_t+1}
         x_{t + 1} = x_t - B^\top (Bx_t - y),
     \end{align}
     so we have
     \begin{align*}
         \|Bx_{t + 1} - Bx^*\|_2
         = & ~ \|B(x_t - B^\top (Bx_t - y)) - Bx^*\|_2\\
         = & ~ \|B(x_t - x^*) - BB^\top B x_t + BB^\top B x^*\|_2\\
         = & ~ \|B(x_t - x^*) - BB^\top B (x_t - x^*)\|_2\\
         = & ~ \|(I - BB^\top)B(x_t - x^*)\|_2\\
         \leq & ~ \|(I - BB^\top)\| \cdot \|B(x_t - x^*)\|_2\\
         \leq & ~ \frac{9}{16} \|B(x_t - x^*)\|_2,
     \end{align*}
     where the first step follows from Eq.~\eqref{eq:x_t+1}, the second step follows from $B^\top B x^* = B^\top y$, the third step follows from simple algebra, the fourth step follows from simple algebra, the fifth step follows from $\|Ax\|_2 \leq \|A\| \cdot \|x\|_2$, and the last step follows from the fact that the eigenvalue of $BB^\top$ belongs to $[\frac{9}{16}, \frac{25}{16}]$ by our assumption. 
     
     Thus we complete the proof.
\end{proof}

\subsection{Forward Error for Overconstrained System}
\label{sub:preliminary:forward_error_simple}

Given a fast regression solver with error guarantees on the cost, it can be conveniently converted to guarantees on the solution. We include the proof of a lemma in~\cite{gsyz23_completion} for completeness.

\begin{lemma}[Lemma 5.5 in \cite{gsyz23_completion}]\label{lem:forward_erorr}
Given a matrix $A \in \R^{n \times d}$ with full column rank, a vector $b \in \R^n$. Suppose there is a vector $x' \in \R^d$ such that
\begin{align*}
\| A x' - b \|_2 \leq (1+\epsilon_1) \min_{x \in \R^d} \| A x - b \|_2
\end{align*}
Let $x_*$ denote the exact solution to the regression problem, then it holds that
\begin{align*}
\| x' - x_* \|_2 \leq O( \sqrt{\epsilon_1} ) \cdot \frac{1}{ \sigma_{\min}(A) } \cdot \| A x_* - b \|_2.
\end{align*}

\end{lemma}
\begin{proof}

Note that 
\begin{align*}
    \|Ax'-Ax_*\|_2=\|Ax'-b-(Ax_*-b)\|_2,
\end{align*}
so we can perform the following decomposition:
\begin{align}\label{eq:A_x'_xopt}
    \|A(x'-x_*)\|_2^2 
    = & ~ \|Ax'-b-(Ax_*-b)\|_2^2 \notag\\
    = & ~ \|Ax'-b\|_2^2-\|Ax_*-b\|_2^2 \notag\\
    \leq & ~ (1+\epsilon_1)^2 \|Ax_*-b\|_2^2 - \|Ax_*-b\|_2^2 \notag\\
    \leq & ~ 4 \epsilon_1 \cdot \|Ax_*-b\|_2^2,
\end{align}
where the first step follows from simple algebra, the second step follows from the Pythagorean theorem, the third step follows from the assumption in Lemma statement, and the fourth step follows from simple algebra. 

Since $A$ has full column rank, $A^\dagger A=I$. Therefore, we have
\begin{align*}
    \|x'-x_*\|_2 
    = & ~ \| A^\dagger A ( x' - x_* ) \|_2 \\
    \leq & ~ \|  A ( x' - x_* ) \|_2 \cdot \| A^\dagger \| \\
    \leq & ~ 2\sqrt{\epsilon_1} \cdot \|Ax_*-b\|_2 \cdot \|A^\dagger\| \\
    = & ~ \frac{2 \sqrt{\epsilon_1} }{\sigma_{\min}(A)}\cdot \|Ax_*-b\|_2,
\end{align*}
where the first step follows from $A^\dagger A=I$, the second step follows from $\|ABx\|_2 \leq \|Ax\|_2 \|B\|$, the third step follows from Eq.~\eqref{eq:A_x'_xopt}, and the last step follows from $\|A^\dagger\| = 1/\sigma_{\min}(A)$, which holds because $A$ has full column rank.
\end{proof}

\subsection{Forward Error for PSD Matrices}
\label{sub:preliminary:forward_error_PSD}

When the matrix in question is PSD, we can obtain a similar error guarantees, in terms $\|b\|_2$ instead of the cost $\|Ax_*-b\|_2$.

\begin{lemma}[PSD version of Lemma~\ref{lem:forward_erorr}]\label{lem:forward_erorr_psd}
Given a matrix $A \in \R^{n \times d}$ with full column rank, a vector $b \in \R^d$. Suppose there is a vector $x' \in \R^d$ such that
\begin{align*}
\| A^\top A x' - b \|_2 \leq \epsilon_2 \| b \|_2 
\end{align*}
Let $x_*$ denote the exact solution to the regression problem, then it holds that
\begin{align*}
\| x' - x_* \|_2 \leq \epsilon_2 \cdot \frac{1}{ \sigma_{\min}(A)^2 } \cdot \| b \|_2.
\end{align*}

\end{lemma}
\begin{proof}

Note that 
\begin{align*}
    \|A^\top Ax' - A^\top Ax_*\|_2=\| ( A^\top Ax' - b ) - ( A^\top Ax_*-b)\|_2,
\end{align*}
so we can perform the following decomposition:
\begin{align}\label{eq:A^topA_x'_x*}
    \| A^\top A (x'-x_*)\|_2^2 
    = & ~ \| (A^\top Ax'-b )-( A^\top A x_*-b)\|_2^2 \notag\\
    = & ~ \| A^\top A x' - b \|_2^2-\| A^\top A x_* - b \|_2^2 \notag\\
    \leq & ~ \epsilon_2^2 \| b \|_2^2
\end{align}
where the first step follows from simple algebra, the second step follows from the Pythagorean theorem, the third step follows from the assumption in lemma statement. 

Since $A$ has full column rank, $(A^\top A)^\dagger (A^\top A ) = I$. Therefore, we have
\begin{align*}
    \|x'-x_*\|_2 
    = & ~ \| (A^\top A)^\dagger A^\top A ( x' - x_* ) \|_2 \\
    \leq & ~ \|  A^\top A ( x' - x_* ) \|_2 \cdot \| (A^\top A)^\dagger \| \\
    \leq & ~ \epsilon_2 \cdot \| b \|_2 \cdot \| (A^\top A)^\dagger\| \\
    = & ~ \epsilon_2 \cdot \sigma_{\min}(A)^{-2} \| b \|_2
\end{align*}
where the first step follows from $(A^\top A)^\dagger (A^\top A)=I$, the second step follows from $\|ABx\|_2 \leq \|Ax\|_2 \|B\|$, the third step follows from Eq.~\eqref{eq:A^topA_x'_x*}, and the last step follows from Fact~\ref{fac:norm}. 
\end{proof}


\section{Fast Regression via Sketching and Preconditioning}
\label{sec:linear_regression}

\subsection{Linear Regression}

In this section, we present the fast linear regression algorithm and analyze its error guarantees and efficiency. 

\begin{algorithm}[!ht]\caption{Algorithm for solving $\min_x \| A x - b_1 \|_2$.}
\label{alg:iter_regression}
\begin{algorithmic}[1]
\Procedure{FastLinearRegression}{$A \in \R^{n \times d}, b_1 \in \R^n, n \in \mathbb{Z}_+, d \in \mathbb{Z}_+, \epsilon_1 \in (0,1),\delta_1 \in (0,1)$}
\State Compute a subspace embedding $S_1$ and apply it to $[A, b_1]$ \Comment{Let $S_1$ denote a $\mathsf{SE}(n,d+1,\epsilon_{\ose}=0.1,\delta_{\ose} = \delta_1/2)$ for $[A, b_1]$}
\State Compute $R$ such that $S_1 AR$ are orthonormal columns via QR decomposition $R \in \R^{d \times d}$
\State $T_1\gets  \Theta( \log(1/\epsilon_1) )$
\State $z_0 \gets \arg\min_{x} \| S_1 A R x - S_1 b_1\|_2$
\For{$t = 0 \to T_1 - 1$}
    \State $z_{t+1} \gets z_t - (R^\top A^\top) ( A R z_t - b_1 )$
\EndFor
\State \Return $R z_{T_1}$
\EndProcedure
\end{algorithmic}
\end{algorithm}

\begin{lemma}[Dense and High Accuracy Regression, Lemma 5.4 in \cite{gsyz23_completion}]\label{lem:alg_takes_time}
Given a matrix $A\in \R^{n\times d}$ with full column rank and a vector $b_1 \in \R^n$, let $\epsilon_1 \in (0, 0.1)$ and $\delta_1 \in (0,0.1)$, there exists an algorithm that takes time  
\begin{align*}
    O((nd+d^3) \cdot \log(1/\epsilon_1) \cdot \log^2(n/\delta_1))
\end{align*}
and outputs $x' \in \R^d$ such that
\begin{align*}
\| A x' - b_1 \|_2 \leq (1+\epsilon_1) \min_{x \in \R^d} \| A x - b_1 \|_2
\end{align*}
holds with probability $1-\delta_1$.
\end{lemma}

\begin{proof}

Let us analyze Algorithm~\ref{alg:iter_regression}, first its convergence then its runtime. Note that the $S_1$ we choose is an $\mathsf{SE}(n,d+1,\epsilon_{\ose},\delta_{\ose})$ for $[A, b_1]$ (Definition~\ref{def:se}). So with probability $1-\delta_{\ose} = 1 - \delta_1/2 \geq 1-\delta_1$, for every $w \in \R^{d+1}$ we have
\begin{align*}
(1-\epsilon_{\ose}) \|[A, b_1] w\|_2 \leq \|S_1 [A, b_1] w\|_2 \leq (1+\epsilon_{\ose}) \|[A, b_1] w\|_2,
\end{align*}
and we assume this from now on. In particular, this holds for every vector $Ax$ and for every vector $Ax - b_1$ with $x \in \R^d$, since both lie in the span of the columns of $[A, b_1]$. Since $A$ has full column rank, so does $S_1 A$, and $S_1A=QR^{-1}$ where $Q$ has orthonormal columns. Hence $\|S_1 A R x\|_2 = \|x\|_2$ for every $x \in \R^d$, and the singular values of $AR$ lie in $[1/(1+\epsilon_{\ose}), 1/(1-\epsilon_{\ose})]$. This interval is contained in $[1-\epsilon_{\ose}, 1+2\epsilon_{\ose}]$, because $(1-\epsilon_{\ose})(1+\epsilon_{\ose}) \leq 1$ and $(1+2\epsilon_{\ose})(1-\epsilon_{\ose}) = 1 + \epsilon_{\ose} - 2\epsilon_{\ose}^2 \geq 1$. Let $AR=U\Sigma V^\top$ be the SVD of $AR$ and $z^*$ denote the optimal solution to the regression $\min_{x\in \R^d} \|ARx-b_1\|_2$. Since $z^*$ minimizes this least squares cost, it satisfies the normal equations
\begin{align}\label{eq:normal_equations_AR}
R^\top A^\top b_1 = R^\top A^\top A R z^*.
\end{align}
Let us consider
\begin{align}\label{eq:AR_xt+1_x*}
    AR(z_{t+1}-z^*) \notag 
    = & ~ AR (z_t+R^\top A^\top (b_1-ARz_t)-z^*) \notag \\
    = & ~ AR(z_t-z^*)+ARR^\top A^\top b_1-ARR^\top A^\top ARz_t \notag \\
    = & ~  AR(z_t-z^*)+ARR^\top A^\top ARz^*-ARR^\top A^\top ARz_t \notag \\
    = & ~ (AR-ARR^\top A^\top AR)(z_t-z^*) \notag \\
    = & ~ (U\Sigma V^\top - U\Sigma^3 V^\top) (z_t-z^*),
\end{align}
where the first step follows from the definition of $z_{t + 1}$ from Algorithm~\ref{alg:iter_regression}, the second step follows from simple algebra, the third step follows from the normal equations $R^\top A^\top b_1 = R^\top A^\top A R z^*$ (Eq.~\eqref{eq:normal_equations_AR}), the fourth step follows from simple algebra, the last step follows from the SVD, $AR=U\Sigma V^\top$.

Therefore,
\begin{align*}
    \|AR(z_{t+1}-z^*)\|_2 
    = & ~ \|(U\Sigma V^\top - U\Sigma^3 V^\top) (z_t-z^*) \|_2 \\
    = & ~ \|(\Sigma-\Sigma^3) V^\top (z_t-z^*)\|_2 \\
    \leq & ~ O(\epsilon_{\ose})\cdot \|V^\top (z_t-z^*)\|_2 \\
    \leq & ~ \frac{O(\epsilon_{\ose})}{1-\epsilon_{\ose}} \|\Sigma V^\top (z_t-z^*)\|_2 \\
    = & ~ O(\epsilon_{\ose})\cdot \|\Sigma V^\top (z_t-z^*)\|_2 \\
    = & ~ O(\epsilon_{\ose})\cdot \|U\Sigma V^\top (z_t-z^*)\|_2 \\
    = & ~ O(\epsilon_{\ose})\cdot \|AR(z_t-z^*)\|_2,
\end{align*}
where the first step follows from Eq.~\eqref{eq:AR_xt+1_x*}, the second step follows from $U^\top U = I$, the third step follows from $\|AB\| \leq \|A\| \cdot \|B\|$, the fourth step follows from $\sigma_{\min}(\Sigma) \geq 1-\epsilon_{\ose}$, the fifth step follows from $\epsilon_{\ose} \in (0,0.1)$, the sixth step follows from $U^\top U = I$, and the last step follows from the SVD, $AR=U\Sigma V^\top$.

This means the error shrinks by a factor of $O(\epsilon_{\ose})$ per iteration. Concretely, for $\epsilon_{\ose} = 0.1$, every singular value $\sigma$ of $AR$ lies in $[1/1.1, 1/0.9]$, so $|1-\sigma^2| \leq \max\{1 - 1/1.21, 1/0.81 - 1\} = 1/0.81 - 1$ and $|\sigma - \sigma^3| = \sigma |1-\sigma^2| \leq (1/0.9) \cdot (1/0.81 - 1) < 0.27$. Hence the factor in the third and fourth steps above is at most $0.27/0.9 = 0.3$. The loop of Algorithm~\ref{alg:iter_regression} runs exactly $T_1$ iterations. Choosing $T_1 = \lceil \log(1/\epsilon_1) / \log(1/0.3) \rceil = \Theta(\log(1/\epsilon_1))$, we have
\begin{align}\label{eq:init_T_gap}
    \|AR(z_{T_1}-z^*)\|_2 \leq & ~ 0.3^{T_1} \cdot \|AR(z_0-z^*)\|_2 \leq \epsilon_1 \cdot \|AR(z_0-z^*)\|_2.
\end{align}
For the initial solution $z_0$, we have
\begin{align*}
    (1-\epsilon_{\ose}) \cdot \|ARz_0-b_1\|_2 \leq & ~ \|S_1 A R z_0 - S_1 b_1\|_2 \\
    \leq & ~ \|S_1 A R z^* - S_1 b_1\|_2 \\
    \leq & ~ (1+\epsilon_{\ose})\cdot \|ARz^*-b_1\|_2,
\end{align*}
where the first and the last steps follow from the subspace embedding property for the vectors $ARz_0 - b_1$ and $ARz^* - b_1$, which lie in the span of the columns of $[A, b_1]$, and the second step follows from the definition of $z_0$. So
\begin{align*}
    \|ARz_0-b_1\|_2 \leq & ~ \frac{1+\epsilon_{\ose}}{1-\epsilon_{\ose}} \cdot \|ARz^*-b_1\|_2.
\end{align*}
The above equation implies that
\begin{align}\label{eq:crude_sketch}
\|ARz_0-b_1\|_2^2  - \|ARz^*-b_1\|_2^2 \leq & ~ \Big( \frac{(1+\epsilon_{\ose})^2}{(1-\epsilon_{\ose})^2} - 1 \Big) \|ARz^*-b_1\|_2^2 \notag \\
= & ~ \frac{4 \epsilon_{\ose}}{(1-\epsilon_{\ose})^2} \|ARz^*-b_1\|_2^2 \notag \\
\leq & ~ 5 \epsilon_{\ose} \|ARz^*-b_1\|_2^2,
\end{align}
where the second step follows from $(1+\epsilon_{\ose})^2 - (1-\epsilon_{\ose})^2 = 4\epsilon_{\ose}$, and the last step follows from $4/(1-\epsilon_{\ose})^2 = 4/0.81 < 5$.

We can wrap up the proof as follows:
\begin{align*}
    \|ARz_{T_1}-b_1\|_2^2
    = & ~ \|AR(z_{T_1}-z^*)\|_2^2+\|ARz^*-b_1\|_2^2 \\
    \leq & ~ \epsilon_1^2 \cdot \|AR(z_0-z^*)\|_2^2+\|ARz^*-b_1\|_2^2 \\
    = & ~ \epsilon_1^2 \cdot (\|ARz_0-b_1\|_2^2-\|ARz^*-b_1\|_2^2) + \|ARz^*-b_1\|_2^2 \\
    \leq & ~ \epsilon_1^2 \cdot ( 5 \epsilon_{\ose} \|ARz^*-b_1\|_2^2 ) + \|ARz^*-b_1\|_2^2 \\
    \leq & ~ (1+\epsilon_1^2)\cdot \|ARz^*-b_1\|_2^2,
\end{align*}
where the first step follows from the Pythagorean theorem and Eq.~\eqref{eq:normal_equations_AR}, the second step follows from Eq.~\eqref{eq:init_T_gap}, the third step follows from the Pythagorean theorem again, the fourth step follows from Eq.~\eqref{eq:crude_sketch}, and the fifth step follows from $5\epsilon_{\ose} = 0.5 \leq 1$. Since $R$ is invertible, $\min_{z \in \R^d} \|ARz - b_1\|_2 = \min_{x \in \R^d} \|Ax - b_1\|_2$, and $1 + \epsilon_1^2 \leq (1+\epsilon_1)^2$. So the output $x' = R z_{T_1}$ satisfies $\|Ax' - b_1\|_2 \leq (1+\epsilon_1) \min_{x \in \R^d} \|Ax - b_1\|_2$.

It remains to show the runtime. Applying $S_1$ to $[A, b_1]$ takes $O(nd\log n)$ time, the QR decomposition takes
$O(m_{\mathrm{sk}} d^2)=O(d^3\log^2(n/\delta_{\ose}))$ time.

Inverting the $d\times d$ triangular factor $R^{-1}$ of the QR decomposition takes $O(d^3)$ time. To solve for $z_0$, we need to multiply $S_1A$ with $R$ in $O(m_{\mathrm{sk}} d^2)$ time and the solve takes $O(m_{\mathrm{sk}} d^2)$ time as well. To implement each of the $T_1$ iterations, we multiply from right to left which takes $O(nd)$ time. Putting things together gives the desired runtime.
\end{proof}

\subsection{Fast PSD Regression Solver}
\label{sec:fast_PSD}

We can extend the algorithm in the preceding section for solving PSD regression.
\begin{algorithm}[!ht]
\caption{Algorithm for solving $\min_{x \in \R^d} \| A^\top A x - b_2 \|_2$.
} \label{alg:fast_psd_regression} 
\begin{algorithmic}[1]
    \Procedure{FastPSDRegression}{$A \in \R^{n \times d}, b_2 \in \R^d, n \in \mathbb{Z}_+, d \in \mathbb{Z}_+, \epsilon_2 \in (0,0.1), \delta_2 \in (0,0.1)$}  \Comment{Lemma~\ref{lem:regression_solver}}\label{line:fast_regression_begin}
    \State Apply $S_2$ to matrix $A$ \Comment{Let $S_2$ denote a $\mathsf{SE}(n,d,\epsilon_{\ose}=0.1, \delta_{\ose}=\delta_2/2)$} \label{line:subspace_embedding} 
    \State Compute $R$ such that $S_2 AR$ orthonormal columns via QR decomposition \Comment{$R \in \R^{d \times d}$}
    \label{line:compute_R}
    \State $z_0 \gets {\bf 0}_d$ 
    \State $t \gets 0$
    \label{line:z0_gets_0}
    \While{$\|A^\top ARz_t - b_2 \|_2 > \epsilon_2 \|b_2\|_2$}
        \State $z_{t + 1} \gets z_t - (R^\top A^\top A R)^\top (R^\top A^\top ARz_t - R^\top b_2)$
        \State $t \gets t+ 1$
    \EndWhile
    \State \Return $Rz_t$
\EndProcedure \label{line:completion_procedure_end}
\end{algorithmic}
\end{algorithm}
\begin{lemma}[Lemma B.1 in \cite{bpsw21}]\label{lem:regression_solver}
    Given a matrix $A \in \R^{n \times d}$ with full column rank and a vector $b_2 \in \R^d$, let $\kappa$ denote the condition number of $A$, i.e. $\kappa = \sigma_{\max}(A)/\sigma_{\min}(A)$, consider the following regression problem
    \begin{align}\label{eq:min_A^top_Ax_y}
        \min_{x \in \R^d} \|A^\top Ax - b_2\|_2.
    \end{align}
    Let $\epsilon_2 \in (0,0.1)$ and $\delta_2 \in (0,0.1)$. There is an algorithm (Algorithm~\ref{alg:fast_psd_regression}) that, with probability at least $1-\delta_2$, stops within time
     \begin{align*}
   O( (n d + d^3) \cdot \log(\kappa/ \epsilon_2) \cdot \log^2(n/\delta_2)).
    \end{align*}
    Whenever it stops, its output $x'\in \R^d$ satisfies
    \begin{align*}
        \|A^\top A x' - b_2\|_2 \leq \epsilon_2 \|b_2\|_2.
    \end{align*}
    The algorithm does not take $\kappa$ as input.
\end{lemma}
\begin{proof}
    Let $S_2 \in \R^{s_2 \times n}$ be the $\mathsf{SE}(n,d, \epsilon_{\ose} = 0.1, \delta_2/2)$ (Definition~\ref{def:se}) for $A$ in Line~\ref{line:subspace_embedding} of Algorithm~\ref{alg:fast_psd_regression}. With probability at least $1 - \delta_2/2 \geq 1-\delta_2$, the following holds for every $x \in \R^d$
    \begin{align}\label{eq:SAx_spectral_norm}
        \|S_2Ax\|_2 = (1 \pm \epsilon_{\ose}) \|Ax\|_2.
    \end{align}
    We call this the success event. The paragraphs on the preconditioner, the convergence and the number of iterations assume that it holds. The paragraph on the output guarantee needs no assumption.

    {\bf The preconditioner.} Since $A$ has full column rank, $\|S_2 A x\|_2 \geq (1-\epsilon_{\ose}) \|Ax\|_2 > 0$ for every $x \neq 0$, so $S_2 A$ has full column rank. Hence the QR decomposition $S_2 A = Q R_0$ has an invertible upper triangular factor $R_0 \in \R^{d \times d}$, and $R := R_0^{-1}$ is invertible and makes $S_2 A R = Q$ have orthonormal columns. We use $R$ as a preconditioner for matrix $A$. Formally, for any $x \in \R^d$ satisfying $\|x\|_2 = 1$, we have $\|S_2 A R x\|_2 = \|x\|_2 = 1$. Applying Eq.~\eqref{eq:SAx_spectral_norm} to the vector $Rx$ gives $(1-\epsilon_{\ose}) \|ARx\|_2 \leq 1 \leq (1+\epsilon_{\ose}) \|ARx\|_2$, that is,
    \begin{align}\label{eq:ARx}
        \frac{1}{1+\epsilon_{\ose}} \leq \|ARx\|_2 \leq \frac{1}{1-\epsilon_{\ose}}, ~~~ \forall x \text{~such~that~} \|x\|_2 = 1.
    \end{align}
    The matrix $R^\top A^\top A R \in \R^{d \times d}$ is symmetric and PSD, and $x^\top R^\top A^\top A R x = \|ARx\|_2^2$ for every $x \in \R^d$. By Eq.~\eqref{eq:ARx}, $(1+\epsilon_{\ose})^{-2} \leq x^\top R^\top A^\top A R x \leq (1-\epsilon_{\ose})^{-2}$ for every unit vector $x$, so every eigenvalue of $R^\top A^\top A R$ lies in $[(1+\epsilon_{\ose})^{-2}, (1-\epsilon_{\ose})^{-2}]$. Since $R^\top A^\top A R$ is symmetric, $\|R^\top A^\top A R x\|_2$ lies between its smallest and its largest eigenvalue for every unit vector $x$. With $\epsilon_{\ose} = 0.1$, we have $(1+\epsilon_{\ose})^{-2} = 1/1.21 > 0.826$ and $(1-\epsilon_{\ose})^{-2} = 1/0.81 < 1.235$, so
    \begin{align}\label{eq:B_well_conditioned}
        \frac{3}{4} < 0.826 < \|R^\top A^\top A R x\|_2 < 1.235 < \frac{5}{4}, ~~~ \forall x \text{~such~that~} \|x\|_2 = 1.
    \end{align}

    {\bf Convergence.} Consider the regression problem
    \begin{align}\label{eq:min_R^top_A^top_ARz_R^top_y}
        \min_{z \in \R^d} \| R^\top A^\top ARz - R^\top b_2\|_2.
    \end{align}
    By Eq.~\eqref{eq:B_well_conditioned}, $R^\top A^\top A R$ is invertible, so the optimal solution to Eq.~\eqref{eq:min_R^top_A^top_ARz_R^top_y} is
    \begin{align}\label{eq:z^*}
        z^* = (R^\top A^\top AR)^{-1} R^\top b_2,
    \end{align}
    and $R^\top A^\top A R z^* = R^\top b_2$. The update of Algorithm~\ref{alg:fast_psd_regression} is the gradient descent update of Lemma~\ref{lem:well_condition} with the matrix $R^\top A^\top A R$ and the vector $R^\top b_2$, and Eq.~\eqref{eq:B_well_conditioned} is the hypothesis of that lemma. By Lemma~\ref{lem:well_condition}, after $t$ iterations we have
    \begin{align}\label{eq:R^top_A^top_AR_zt_z^*}
        \| R^\top A^\top AR (z_t - z^*) \|_2 \leq c^t \| R^\top A^\top AR (z_0 - z^*) \|_2,
    \end{align}
    where $c \leq 9/16$. Let
    \begin{align}\label{eq:x_t}
        x_t = Rz_t.
    \end{align}
    Loading Eq.~\eqref{eq:z^*}, the definition $z_0 = 0$ and Eq.~\eqref{eq:x_t} into Eq.~\eqref{eq:R^top_A^top_AR_zt_z^*}, we have
\begin{align}\label{eq:R^top_A^top_A_xt_R^top_y_1}
\| R^\top A^\top A x_t - R^\top b_2 \|_2 = & ~ \| R^\top A^\top A R (z_t - z^*) \|_2 \notag\\
\leq & ~ c^t \cdot \| R^\top b_2 \|_2 \notag\\
\leq & ~ c^t \cdot \sigma_{\max}(R) \cdot \| b_2 \|_2,
\end{align}
where the first step follows from $x_t = R z_t$ and $R^\top A^\top A R z^* = R^\top b_2$, the second step follows from Eq.~\eqref{eq:R^top_A^top_AR_zt_z^*} and $R^\top A^\top A R (z_0 - z^*) = -R^\top b_2$, and the third step follows from $\sigma_{\max}(R^\top) = \sigma_{\max}(R)$, which holds because $R$ is a square matrix. On the other hand, we have
    \begin{align}\label{eq:R^top_A^top_A_xt_R^top_y_2}
        \| R^\top A^\top A x_t - R^\top b_2 \|_2
        = & ~ \| R^\top( A^\top A x_t - b_2) \|_2 \notag\\
        \geq & ~ \sigma_{\min}(R) \|A^\top A x_t - b_2\|_2,
    \end{align}
    where the first step follows from simple algebra and the second step follows from the definition of $\sigma_{\min}(R^\top)$ and $\sigma_{\min}(R^\top) = \sigma_{\min}(R)$, which holds because $R$ is a square matrix. Note that $\sigma_{\min}(R) > 0$ because $R$ is invertible.

    Next, we bound $\kappa(R)$. For every unit vector $x \in \R^d$, we have
    \begin{align*}
        \sigma_{\min}(A) \|Rx\|_2 \leq \|ARx\|_2 \leq \sigma_{\max}(A) \|Rx\|_2.
    \end{align*}
    Since $A$ has full column rank, $\sigma_{\min}(A) > 0$. Taking the maximum over unit vectors $x$ in the first inequality gives $\sigma_{\max}(R) \leq \sigma_{\max}(AR) / \sigma_{\min}(A)$, and taking the minimum over unit vectors $x$ in the second inequality gives $\sigma_{\min}(R) \geq \sigma_{\min}(AR)/\sigma_{\max}(A)$. Dividing the two bounds, we obtain
    \begin{align}\label{eq:kappa_R_bound}
        \kappa(R) = \frac{\sigma_{\max}(R)}{\sigma_{\min}(R)} \leq \frac{\sigma_{\max}(AR)}{\sigma_{\min}(AR)} \cdot \frac{\sigma_{\max}(A)}{\sigma_{\min}(A)} = \kappa(AR) \kappa(A).
    \end{align}
    By Eq.~\eqref{eq:ARx}, $\sigma_{\max}(AR) \leq 1/(1-\epsilon_{\ose})$ and $\sigma_{\min}(AR) \geq 1/(1+\epsilon_{\ose})$, so $\kappa(AR) \leq (1+\epsilon_{\ose})/(1-\epsilon_{\ose}) = 1.1/0.9 = 11/9$.

    Putting it all together, after $t$ iterations we have
    \begin{align}\label{eq:psd_residual_t}
        \|A^\top A x_t - b_2\|_2
        \leq & ~ c^t \kappa(R) \|b_2\|_2 \notag\\
        \leq & ~ c^t \kappa(AR) \kappa (A) \|b_2\|_2 \notag\\
        \leq & ~ 2 c^t \kappa(A) \|b_2\|_2,
    \end{align}
    where the first step follows from Eq.~\eqref{eq:R^top_A^top_A_xt_R^top_y_1} and Eq.~\eqref{eq:R^top_A^top_A_xt_R^top_y_2}, the second step follows from Eq.~\eqref{eq:kappa_R_bound}, and the last step follows from $\kappa(AR) \leq 11/9 \leq 2$.

    {\bf Number of iterations.} The algorithm evaluates its test before every update, at $t = 0, 1, 2, \ldots$. Let
    \begin{align*}
        T_2 := \left\lceil \frac{\log(2\kappa(A)/\epsilon_2)}{\log(16/9)} \right\rceil = O(\log(\kappa/\epsilon_2)).
    \end{align*}
    Since $c \leq 9/16$, we have $2 c^{T_2} \kappa(A) \leq 2 (9/16)^{T_2} \kappa(A) \leq \epsilon_2$. So Eq.~\eqref{eq:psd_residual_t} gives $\|A^\top A R z_{T_2} - b_2\|_2 \leq \epsilon_2 \|b_2\|_2$, and the test fails at the latest when $t = T_2$. Hence the loop performs at most $T_2$ updates. In particular, if $b_2 = 0$, then $\|A^\top A R z_0 - b_2\|_2 = 0$, and the algorithm returns $0$ at once.

    {\bf Output guarantee.} When the loop stops at some $t$, its test has failed, which means $\|A^\top A R z_t - b_2\|_2 \leq \epsilon_2 \|b_2\|_2$. The output is $x' = R z_t$, so
    \begin{align*}
        \|A^\top A x' - b_2\|_2 \leq \epsilon_2 \|b_2\|_2.
    \end{align*}
    This holds whenever the algorithm stops, whether or not the success event holds.

    {\bf Running time.} By Lemma~\ref{lem:subspace_embedding} with $\epsilon = 0.1$, an $\mathsf{SRHT}$ matrix $S_2$ with $s_2 = O(d \log(n/\delta_2))$ rows works, and $S_2 A$ takes $O(nd\log(n/\delta_2))$ time. The QR decomposition of $S_2 A \in \R^{s_2 \times d}$ takes $O(s_2 d^2) = O(d^3 \log(n/\delta_2))$ time, and inverting the triangular factor $R_0$ takes $O(d^3)$ time. Each evaluation of the test and each update multiplies vectors by $R$, $A$, $A^\top$ and $R^\top$ from right to left, which takes $O(nd + d^2)$ time. On the success event, there are at most $T_2$ updates and $T_2 + 1$ evaluations of the test, so the total running time is
    \begin{align*}
        O((nd + d^3) \cdot \log(n/\delta_2) + (nd + d^2) \cdot (T_2 + 1)) = O( (n d + d^3) \cdot \log(\kappa/ \epsilon_2) \cdot \log^2(n/\delta_2)).
    \end{align*}

    {\bf Failure probability.} The success event holds with probability at least $1-\delta_2$. On this event, the algorithm stops within the running time above, and its output satisfies $\|A^\top A x' - b_2\|_2 \leq \epsilon_2 \|b_2\|_2$ by the output guarantee.
\end{proof}

\section{Even and Odd Power Regression: Base Case}
\label{sec:fast_attention}

We showcase the algorithms for solving even and odd power regression when there are only 4 (for even) and 3 (for odd) matrices are involved. This simple case serves as a basis, both algorithmically and analytically, for the more complicated algorithm.

The algorithms in this section, and Algorithms~\ref{alg:even} and \ref{alg:odd}, set their accuracy parameters from the condition number of $A$. They take as input a number $\kappa \geq \kappa(A)$. Such a $\kappa$ with $\kappa \leq 1.5 \kappa(A)$ can be computed from a sketch, without computing $\kappa(A)$ exactly. Let $S$ be an $\mathsf{SE}(n,d,0.1,\delta)$ for $A$ (Definition~\ref{def:se}). Then with probability $1-\delta$, we have $0.9 \|Ax\|_2 \leq \|SAx\|_2 \leq 1.1 \|Ax\|_2$ for every $x \in \R^d$. Taking the maximum and the minimum over unit vectors $x$ gives $0.9 \sigma_{\max}(A) \leq \sigma_{\max}(SA) \leq 1.1 \sigma_{\max}(A)$ and $0.9 \sigma_{\min}(A) \leq \sigma_{\min}(SA) \leq 1.1 \sigma_{\min}(A)$, so $(9/11) \kappa(A) \leq \kappa(SA) \leq (11/9) \kappa(A)$. Hence $\wh{\kappa} := (11/9) \kappa(SA)$ satisfies $\kappa(A) \leq \wh{\kappa} \leq (11/9)^2 \kappa(A) \leq 1.5 \kappa(A)$. With an $\mathsf{SRHT}$ matrix $S$ (Lemma~\ref{lem:subspace_embedding}), computing $SA$ and the singular values of $SA$ takes $O(nd\log(n/\delta) + d^3 \log(n/\delta))$ extra time, and the failure probability $\delta$ joins the union bound. The bounds below then hold with $\wh{\kappa}$ as $\kappa$. Since $\wh{\kappa} \leq 1.5 \kappa(A)$, we have $\log \wh{\kappa} \leq \log \kappa(A) + \log 1.5$. So each factor $\log(\wh{\kappa}/\epsilon)$ in the running times of the algorithms in this section exceeds $\log(\kappa(A)/\epsilon)$ by at most $\log 1.5$, and the term $j^2 \log \wh{\kappa}$ in the running times of Algorithms~\ref{alg:even} and \ref{alg:odd} exceeds $j^2 \log \kappa(A)$ by at most $j^2 \log 1.5$.

\subsection{Three Matrices}

Our algorithm will be first solving an overconstrained regression, then followed with a PSD regression.

\begin{algorithm}[!ht]\caption{Algorithm for solving the regression $\min_x \| A A^\top A x - b \|_2$.}
\label{alg:AA^topAx_b_3}
\begin{algorithmic}[1]
\Procedure{FastAttentionRegression}{$A \in \R^{n \times d}, b \in \R^n,n \in \mathbb{Z}_+, d \in \mathbb{Z}_+, \kappa \geq \kappa(A), \epsilon_3 \in (0, 0.1),\delta_3 \in (0, 0.1)$}
    \State $\epsilon_1 \gets  0.1 \epsilon_3$
    \State $\delta_1 \gets \delta_3/2$
    \State $b_2 \gets \textsc{FastLinearRegression}(A \in \R^{n \times d},b \in \R^n,n,d,\epsilon_1,\delta_1)$ \Comment{$b_2 \in \R^d$}
    \State $\epsilon_2 \gets \epsilon_3 / \kappa$
    \State $\delta_2 \gets \delta_3/2$
    \State $x'\gets \textsc{FastPSDRegression}(A \in \R^{n \times d },b_2 \in \R^d,n,d,\epsilon_2,\delta_2)$ 
    \State \Return $x'$
\EndProcedure
\end{algorithmic}
\end{algorithm}

\begin{lemma}
Let $A \in \R^{n \times d}$ be a matrix with full column rank and $b \in \R^n$ be a vector. Let $\kappa \geq \kappa(A)$ be an upper bound on the condition number of $A$ (see Definition~\ref{def:kappa}), given as input. Let $\epsilon_3, \delta_3 \in (0, 0.1)$. Consider the regression problem (defined in Definition~\ref{def:3_A}):
\begin{align*}
    \min_{x \in \R^d} \| AA^\top A x - b \|_2.
\end{align*}
There exists an algorithm (Algorithm~\ref{alg:AA^topAx_b_3}) that, with probability at least $1-\delta_3$, stops within time
\begin{align*}
O ( (n d + d^3) \cdot \log(\kappa/ \epsilon_3) \cdot \log^2(n/\delta_3) )
\end{align*}
and outputs a vector $x' \in \R^d$ such that
\begin{align*}
    \| AA^\top A x' - b \|_2 \leq (1+\epsilon_3) \min_{x \in \R^d} \| A A^\top A x - b\|_2 + \epsilon_3 \| b \|_2.
\end{align*}
\end{lemma}

\begin{proof}
We define $\OPT$ as
\begin{align*}
\OPT:=  \min_{x \in \R^d} \| A A^\top A x - b\|_2 .
\end{align*}
First, we use Algorithm~\ref{alg:iter_regression} to solve 
    \begin{align}\label{eq:Ab_2_b_3}
        \min_{y \in \R^d} \| A y - b \|_2.
    \end{align}

    Let $y_*$ denote the exact solution to this regression problem. By Lemma~\ref{lem:alg_takes_time}, we can get $y' \in \R^d$ such that the following holds with probability $1-\delta_1$,
    \begin{align}\label{eq:Ay_b_3}
        \| A y' - b \|_2 \leq & ~ (1+\epsilon_1) \cdot \min_{y \in \R^d} \| A y - b \|_2 \notag \\
        \leq & ~ (1+\epsilon_1) \cdot \OPT,
    \end{align}
    where the last step follows from the $A^\top A x$ might not be able to do a better job than minimizer $y$ (in terms of minimizing cost).
    
    This step takes time 
    \begin{align*}
        O((nd +d^3 ) \cdot \log(1/\epsilon_1) \cdot  \log^2(n/\delta_1) ) .
    \end{align*}

By triangle inequality, we can show
\begin{align}\label{eq:||y'||_2}
\| y'\|_2 
= & ~ \|y' - y_* + y_* \|_2 \notag \\
\leq & ~ \|y' - y_* \|_2 + \| y_* \|_2.
\end{align}

To bound the first term of Eq.~\eqref{eq:||y'||_2}, we have
\begin{align}\label{eq:bound_y'_y_*}
    \|y' - y_* \|_2
    \leq & ~ O(\sqrt{\epsilon_1}) \cdot \sigma_{\min}(A)^{-1} \| A y_* - b \|_2 \notag\\
    \leq & ~ O(\sqrt{\epsilon_1}) \cdot \sigma_{\min}(A)^{-1} \OPT,
\end{align}
where the first step follows from Lemma~\ref{lem:forward_erorr} and the second step follows from Eq.~\eqref{eq:Ay_b_3}.

To bound the second term of Eq.~\eqref{eq:||y'||_2}, we have
\begin{align}\label{eq:bound_y_*}
    \| y_* \|_2
    = & ~ \| A^\dagger b \|_2 \notag\\
    \leq & ~ \| A^\dagger\| \cdot \|b \|_2 \notag \\
    \leq & ~ \sigma_{\min}(A)^{-1} \cdot \| b \|_2,
\end{align}
where the first step follows from $ y_* = A^\dagger b $, the second step follows from $\|Ax\|_2 \leq \|A\| \|x\|_2$, and the third step follows from $\| A^\dagger\| = \sigma_{\min}(A)^{-1}$.

By plugging Eq.~\eqref{eq:bound_y'_y_*} and Eq.~\eqref{eq:bound_y_*} into Eq.~\eqref{eq:||y'||_2}, we have
\begin{align}\label{eq:finish_||y'||_2}
    \| y'\|_2 \leq O(\sqrt{\epsilon_1}) \cdot \sigma_{\min}(A)^{-1} \OPT + \sigma_{\min}(A)^{-1} \| b \|_2.
\end{align}

Let $b_2 = y' \in \R^d$ and $x' \in \R^d$. Then, using Algorithm~\ref{alg:fast_psd_regression}, we solve 
    \begin{align*}
        \min_{x' \in \R^d} \| A^\top A x' - b_2 \|_2.
    \end{align*}
    
    By Lemma~\ref{lem:regression_solver}, we find an $x'\in \R^d$ such that
    \begin{align}\label{eq:A^topAx'-b_2_leq_epsilon_2_b_2_2}
        \|A^\top A x' - b_2\|_2 \leq \epsilon_2 \|b_2\|_2
    \end{align}
    holds with probability $1-\delta_2$.

    This step takes time
    \begin{align*}
        O((nd + d^3 ) \cdot \log(\kappa(A)/ \epsilon_2) \cdot \log^2(n/\delta_2) ).
    \end{align*}

{\bf Correctness.}

To bound $\| A A^\top A x' - b \|_2$, we have
\begin{align}\label{eq:AA^topAx'_b_3}
 \| A A^\top A x' - b \|_2
 = & ~ \| A A^\top A x' - A y' + Ay' - b \|_2 \notag \\
 \leq & ~ \| A A^\top A x' - A y' \|_2 + \| Ay'-b \|_2 \notag \\
 \leq & ~ \| A A^\top A x' - A y' \|_2  + (1+\epsilon_1) \cdot \OPT,
 \end{align}
 where the first step follows from adding and subtracting the same thing, the second step follows from triangle inequality, and the third step follows from Eq.~\eqref{eq:Ay_b_3}. Let's consider the first term of Eq.~\eqref{eq:AA^topAx'_b_3}. We have
 \begin{align}\label{eq:bound_AA^topAx'_Ay'}
     \| A A^\top A x' - A y' \|_2
     = & ~ \| A (A^\top A x' - y') \|_2 \notag \\
     \leq & ~ \|A \| \cdot \| A^\top A x' - y' \|_2 \notag \\
     \leq & ~ \| A \| \cdot \epsilon_2 \| y' \|_2 \notag \\
     \leq & ~ \| A \| \cdot \epsilon_2 \left(O(\sqrt{\epsilon_1})  \cdot \sigma_{\min}(A)^{-1} \OPT + \sigma_{\min}(A)^{-1} \| b \|_2 \right) \notag \\
     = & ~ \| A \| \cdot \epsilon_2 O(\sqrt{\epsilon_1}) \cdot \sigma_{\min}(A)^{-1} \OPT + \| A \| \cdot \epsilon_2 \sigma_{\min}(A)^{-1} \| b \|_2,
 \end{align}
 where the first step follows from simple algebra, the second step follows from $\|Ax\|_2 \leq \|A\| \|x\|_2$, the third step follows from Eq.~\eqref{eq:A^topAx'-b_2_leq_epsilon_2_b_2_2}, the fourth step follows from Eq.~\eqref{eq:finish_||y'||_2}, and the last step follows from simple algebra.

 Then, to bound the first term of Eq.~\eqref{eq:bound_AA^topAx'_Ay'}, we have
 \begin{align}\label{eq:||A||_epsilon_2_O}
     \| A \| \cdot \epsilon_2 O(\sqrt{\epsilon_1}) \cdot \sigma_{\min}(A)^{-1} \OPT
     \leq & ~ \sigma_{\max}(A)\sigma_{\min}(A)^{-1} \cdot \epsilon_2 O(\sqrt{\epsilon_1}) \cdot \OPT \notag \\
     = & ~ \sigma_{\max}(A)\sigma_{\min}(A)^{-1} \cdot O(\sqrt{\epsilon_1}\epsilon_2) \cdot \OPT \notag \\
     = & ~ O(\sqrt{\epsilon_1} \epsilon_2) \kappa(A) \OPT \notag \\
     \leq & ~ O(\sqrt{\epsilon_1} \epsilon_3) \OPT,
 \end{align}
 where the first step follows from $\| A \| \leq \sigma_{\max}(A)$, the second step follows from the property of $O(\cdot)$, the third step follows from the definition of $\kappa(A)$ (see Definition~\ref{def:kappa}), and the last step follows from $\epsilon_2 \kappa(A) = \epsilon_3 \kappa(A) / \kappa$ and $\kappa(A)/\kappa \leq 1$. Similarly, to bound the second term of Eq.~\eqref{eq:bound_AA^topAx'_Ay'}, we get
 \begin{align}\label{eq:||A||_epsilon_2_sigma}
     \| A \| \cdot \epsilon_2 \sigma_{\min}(A)^{-1} \| b \|_2
     \leq & ~ \sigma_{\max}(A) \sigma_{\min}(A)^{-1} \cdot \epsilon_2 \| b \|_2 \notag \\
    = & ~ \kappa(A) \cdot \epsilon_2 \| b \|_2 \notag \\
    \leq & ~ \epsilon_3 \| b \|_2,
 \end{align}
 where the first step follows from $\| A \| \leq \sigma_{\max}(A)$, the second step follows from the definition of $\kappa(A)$ (see Definition~\ref{def:kappa}), and the last step follows from $\epsilon_2 \kappa(A) = \epsilon_3 \kappa(A) / \kappa$ and $\kappa(A)/\kappa \leq 1$.

 Plugging Eq.~\eqref{eq:||A||_epsilon_2_O} and Eq.~\eqref{eq:||A||_epsilon_2_sigma} into Eq.~\eqref{eq:bound_AA^topAx'_Ay'}, we get 
 \begin{align}\label{eq:finish_bound_AA^topAx'_Ay'}
     \| A A^\top A x' - A y' \|_2 \leq O(\sqrt{\epsilon_1} \epsilon_3) \OPT + \epsilon_3 \| b \|_2.
 \end{align}

 Therefore, by plugging Eq.~\eqref{eq:finish_bound_AA^topAx'_Ay'} into \eqref{eq:AA^topAx'_b_3}, we have
 \begin{align*}
     \| A A^\top A x' - b \|_2 
     \leq & ~ O(\sqrt{\epsilon_1} \epsilon_3) \OPT + \epsilon_3 \| b \|_2 + (1+\epsilon_1) \cdot \OPT \\
     \leq & ~ (1+\epsilon_3) \cdot \OPT + \epsilon_3 \| b \|_2,
 \end{align*}
where the last step follows from $\epsilon_1 = 0.1 \epsilon_3$ and from the constant $2$ in the $O(\sqrt{\epsilon_1})$ of Lemma~\ref{lem:forward_erorr} (its proof gives this constant), which give $\epsilon_1 + 2\sqrt{\epsilon_1} \epsilon_3 \leq 0.1 \epsilon_3 + 2\sqrt{0.1} \cdot \epsilon_3 < \epsilon_3$, since $\epsilon_1 < 0.1$ and $0.1 + 2\sqrt{0.1} < 0.74$. Therefore, we complete bounding $\| A A^\top A x' - b \|_2$.

{\bf Running time.}

We have $\epsilon_1 = 0.1 \epsilon_3$, $\delta_1 = \delta_2 = \delta_3/2$ and $\kappa(A)/\epsilon_2 = \kappa(A) \kappa / \epsilon_3 \leq (\kappa/\epsilon_3)^2$. So the overall running time is
\begin{align*}
        O((nd+d^3)\log(\kappa /\epsilon_3) \cdot \log^2(n/\delta_3))
\end{align*}

{\bf Failure probability.} 

By taking a union over two events, the failure probability is at most $\delta_1 + \delta_2 =\delta_3 $.
\end{proof}

\subsection{Four Matrices}
\label{sec:four_matrices}

For four matrices, the algorithm will be alternating two PSD regressions.
\begin{algorithm}[!ht]\caption{Algorithm for solving $\min_{x \in \R^d} \| A^\top AA^\top A x - b_4 \|_2$.}
\label{alg:A^topAA^topAx_b_3}
\begin{algorithmic}[1]
\Procedure{FourMatrices}{$A \in \R^{n \times d}, b_4 \in \R^d ,n \in \mathbb{Z}_+, d \in \mathbb{Z}_+,\kappa \geq \kappa(A),\epsilon_4 \in (0, 0.1),\delta_4 \in (0, 0.1)$}
    \State $\epsilon_2 \gets 0.1 \epsilon_4 /  \kappa^2 $
    \State $\delta_2 \gets \delta_4/2$
    \State $b_2 \gets \textsc{FastPSDRegression}(A \in \R^{n \times d},b_4 \in \R^d,n,d,\epsilon_2,\delta_2)$

    \State $x'\gets \textsc{FastPSDRegression}(A \in \R^{n \times d },b_2 \in \R^d,n,d,\epsilon_2,\delta_2)$  
    \State \Return $x'$
\EndProcedure
\end{algorithmic}
\end{algorithm}

\begin{lemma}
Let $A \in \R^{n \times d}$ be a matrix with full column rank and $b_4 \in \R^d$ be a vector. Let $\kappa \geq \kappa(A)$ be an upper bound on the condition number of $A$ (see Definition~\ref{def:kappa}), given as input. Let $\epsilon_4, \delta_4 \in (0, 0.1)$. Consider the regression problem
\begin{align*}
    \min_{x \in \R^d} \| A^\top AA^\top A x - b_4 \|_2.
\end{align*}
There exists an algorithm (Algorithm~\ref{alg:A^topAA^topAx_b_3}) that, with probability at least $1-\delta_4$, stops within time
\begin{align*}
O ( (n d + d^3) \cdot \log(\kappa/ \epsilon_4) \cdot \log^2(n/\delta_4) )
\end{align*}
and outputs a vector $x' \in \R^d$ such that
\begin{align*}
    \| A^\top AA^\top A x' - b_4 \|_2 \leq  \epsilon_4 \| b_4 \|_2.
\end{align*}
\end{lemma}
\begin{proof}
First, we use Algorithm~\ref{alg:fast_psd_regression} to solve
\begin{align*}
    \min_{y \in \R^d} \| A^\top A y - b_4 \|_2.
\end{align*}
Let $y_*$ denote the exact solution to this regression problem. By Lemma~\ref{lem:regression_solver}, we get 
\begin{align}\label{eq:A^topAy'_b_4}
    \|A^\top A y' - b_4\|_2 
    \leq & ~ \epsilon_2 \|b_4\|_2 .
\end{align}
This step takes time
\begin{align*}
    O( (n d + d^3) \cdot \log(\kappa(A)/ \epsilon_2) \cdot \log^2(n/\delta_2)).
\end{align*}
By triangle inequality, we can show that 
\begin{align}\label{eq:four_matrix_y'}
    \|y'\|_2 
    \leq & ~ \|y' - y_* + y_*\|_2 \notag\\
    \leq & ~ \|y' - y_*\|_2 + \|y_*\|_2.
\end{align}
To bound the first term of Eq.~\eqref{eq:four_matrix_y'}, we have
\begin{align}\label{eq:four_matrix_y'_y_*}
    \|y' - y_*\|_2 \leq \epsilon_2 \cdot \sigma_{\min}(A)^{-2} \cdot \| b_4 \|_2.
\end{align}
where the last step follows from Lemma~\ref{lem:forward_erorr_psd}. To bound the second term of Eq.~\eqref{eq:four_matrix_y'}, we have
\begin{align}\label{eq:four_matrix_y_*}
    \|y_*\|_2 
    = & ~ \| (A^\top A)^\dagger b_4 \|_2 \notag\\
   \leq & ~ \| (A^\top A)^\dagger \| \cdot \| b_4 \|_2 \notag \\
    \leq & ~ \sigma_{\min} (A)^{-2} \cdot \|b_4\|_2,
\end{align}
where the first step follows from the definition of $y_*$, the second step follows from $\|Ax\|_2 \leq \|A\| \|x\|_2$, the last step follows from Fact~\ref{fac:norm}. Then, plugging Eq.~\eqref{eq:four_matrix_y'_y_*} and Eq.~\eqref{eq:four_matrix_y_*} into Eq.~\eqref{eq:four_matrix_y'}, we have
\begin{align}\label{eq:four_matrix_final_y'}
    \|y'\|_2 
    \leq & ~ (\epsilon_2 + 1) \sigma_{\min} (A)^{-2}   \|b_4\|_2 \notag\\
    \leq & ~ 2 \sigma_{\min} (A)^{-2}   \|b_4\|_2,
\end{align}
where the first step follows from simple algebra and the second step follows from $\epsilon_2 < 1$. Let $b_2 = y' \in \R^d$ and $x' \in \R^d$. Then, using Algorithm~\ref{alg:fast_psd_regression} again, we solve
\begin{align*}
    \min_{y \in \R^d} \| A^\top A x' - b_2 \|_2.
\end{align*}

By Lemma~\ref{lem:regression_solver}, we can find $x' \in \R^d$ such that 
\begin{align}\label{eq:A^topAx'_b_2}
    \|A^\top A x' - b_2\|_2 \leq \epsilon_2 \|b_2\|_2
\end{align}
holds with probability $1 - \delta_2$. This step takes time
\begin{align*}
    O( (n d + d^3) \cdot \log(\kappa(A)/ \epsilon_2) \cdot \log^2(n/\delta_2)).
\end{align*}

{\bf Correctness.}

To bound $\|A^\top A A^\top A x' - b_4 \|_2$, we have
\begin{align}\label{eq:A^topAA^topAx'_b_3}
 \|A^\top A A^\top A x' - b_4 \|_2
 = & ~ \|A^\top A A^\top A x' - A^\top A y' + A^\top Ay' - b_4 \|_2 \notag \\
 \leq & ~ \| A^\top A A^\top A x' - A^\top A y' \|_2 + \| A^\top A y' - b_4 \|_2 \notag \\
 \leq & ~ \| A^\top A A^\top A x' - A^\top A y' \|_2  + \epsilon_2 \| b_4 \|_2
 \end{align}
 where the first step follows from adding and subtracting the same thing, the second step follows from triangle inequality, and the third step follows from Eq.~\eqref{eq:A^topAy'_b_4}. Let's consider the first term of Eq.~\eqref{eq:A^topAA^topAx'_b_3}. We have
 \begin{align}\label{eq:bound_A^topAA^topAx'_Ay'}
     \| A^\top A A^\top A x' - A^\top A y' \|_2
     = & ~ \| A^\top A (A^\top A x' - y') \|_2 \notag \\
     \leq & ~ \|A^\top A \| \cdot \| A^\top A x' - y' \|_2 \notag \\
     \leq & ~ \| A^\top A \| \cdot \epsilon_2 \| y' \|_2 \notag \\
     \leq & ~ \| A^\top A \| \cdot \epsilon_2 \cdot (2 \sigma_{\min} (A)^{-2} \|b_4\|_2) \notag \\
     \leq & ~ \sigma_{\max}(A)^2 \cdot \epsilon_2  \cdot ( 2 \sigma_{\min} (A)^{-2}   \|b_4\|_2 ) \notag \\
     = & ~ 2 \kappa(A)^2 \epsilon_2 \| b_4 \|_2 \notag \\
     \leq & ~ 0.2 \epsilon_4 \| b_4 \|_2,
 \end{align}
 where the first step follows from simple algebra, the second step follows from $\|Ax\|_2 \leq \|A\| \|x\|_2$, the third step follows from Eq.~\eqref{eq:A^topAx'_b_2}, the fourth step follows from Eq.~\eqref{eq:four_matrix_final_y'}, the fifth step follows from Fact~\ref{fac:norm}, the sixth step follows from the definition of $\kappa(A)$ (see Definition~\ref{def:kappa}), and the last step follows from $\epsilon_2 = 0.1 \epsilon_4/\kappa^2$ and $\kappa(A) \leq \kappa$, which give $2\kappa(A)^2 \epsilon_2 = 0.2 \epsilon_4 (\kappa(A)/\kappa)^2 \leq 0.2 \epsilon_4$.

Plugging Eq.~\eqref{eq:bound_A^topAA^topAx'_Ay'} into Eq.~\eqref{eq:A^topAA^topAx'_b_3}, we have
\begin{align*}
    \|A^\top A A^\top A x' - b_4 \|_2 \leq (\epsilon_2 + 0.2 \epsilon_4) \| b_4 \|_2 \leq \epsilon_4 \| b_4 \|_2,
\end{align*}
where the last step follows from $\epsilon_2 = 0.1 \epsilon_4 / \kappa^2 \leq 0.1 \epsilon_4$, since $\kappa \geq \kappa(A) \geq 1$. Therefore, we complete bounding $ \|A^\top A A^\top A x' - b_4 \|_2 $.

{\bf Running time.}

We have $\delta_2 = \delta_4/2$ and $\kappa(A)/\epsilon_2 = 10 \kappa(A) \kappa^2/\epsilon_4 \leq 10 (\kappa/\epsilon_4)^3$. So the total running time of the two steps is
\begin{align*}
    O ( (n d + d^3) \cdot \log(\kappa/ \epsilon_4) \cdot \log^2(n/\delta_4) ).
\end{align*}

{\bf Failure probability.}

By taking a union over two events, the failure probability is at most $\delta_2 + \delta_2 =\delta_4 $.
\end{proof}

\section{Even and Odd Power Regression: Complete Algorithm}
\label{sec:even}

We provide the complete algorithm and analysis for our even and odd power regressions via induction.

\subsection{Induction Hypothesis and Step}
\label{sub:even:induction}

In this section, we present our induction hypothesis and prove the induction step.

\begin{lemma}[Induction Hypothesis]\label{lem:induction_hypothesis}
Let $A \in \R^{n \times d}$ be a matrix with full column rank and $b \in \R^d$ be a vector. Let $\kappa \geq \kappa(A)$ be an upper bound on the condition number of $A$ (see Definition~\ref{def:kappa}), given as input. Let $j$ be a positive integer, and let $\epsilon_{\mathrm{final}}, \delta_{\mathrm{final}} \in (0,0.1)$. Let $b_0 := b$, and for $k \in [j]$, let $b_k$ be the vector computed in the $k$-th iteration of Algorithm~\ref{alg:even}, that is, the output of Algorithm~\ref{alg:fast_psd_regression} on $A$ and $b_{k-1}$ with accuracy $\eta_k = \epsilon_{\mathrm{final}}/(4 j \kappa^{2(k-1)})$ and failure probability $\delta_k = \delta_{\mathrm{final}}/j$. Define
\begin{align*}
    \epsilon_i := \frac{i \cdot \epsilon_{\mathrm{final}}}{2j}, ~~~ \forall i \in \{0, 1, \ldots, j\},
\end{align*}
so that $\epsilon_0 = 0$ and $\epsilon_{i-1} \leq \epsilon_i$ for all $i \in [j]$. Let $C > 0$ be a constant such that the running time bound of Lemma~\ref{lem:regression_solver} reads $C \cdot (nd + d^3) \cdot \log(\kappa(A)/\epsilon_2) \cdot \log^2(n/\delta_2)$ for all $\epsilon_2, \delta_2 \in (0, 0.1)$.

For $k \in \{0, 1, \ldots, j\}$, let $\mathcal{E}_k$ denote the event that the following three statements hold:
\begin{enumerate}
    \item $\| (A^\top A)^i b_i - b_0 \|_2 \leq \epsilon_i \| b_0 \|_2$ for all $i \in \{0, 1, \ldots, k\}$.
    \item $\| b_i \|_2 \leq 2\sigma_{\min}(A)^{-2i} \| b_0 \|_2$ for all $i \in \{0, 1, \ldots, k\}$.
    \item The first $k$ calls to Algorithm~\ref{alg:fast_psd_regression} take total time at most
    \begin{align*}
        C \cdot (nd + d^3) \cdot \sum_{i=1}^{k} \Big( (2i-1) \log \kappa + \log(4j/\epsilon_{\mathrm{final}}) \Big) \cdot \log^2(jn/\delta_{\mathrm{final}}).
    \end{align*}
\end{enumerate}
Suppose that $\Pr[\mathcal{E}_k] \geq 1 - k\delta_{\mathrm{final}}/j$ for some $k \in \{0, 1, \ldots, j-1\}$. Then $\Pr[\mathcal{E}_{k+1}] \geq 1 - (k+1)\delta_{\mathrm{final}}/j$.
\end{lemma}

\begin{proof}

By hypothesis, $\mathcal{E}_k$ fails with probability at most $k \delta_{\mathrm{final}}/j$. Let $\mathcal{F}_{k+1}$ denote the event that the $(k+1)$-th call to Algorithm~\ref{alg:fast_psd_regression} stops within time $C \cdot (nd + d^3) \cdot \log(\kappa(A)/\eta_{k+1}) \cdot \log^2(n/\delta_{k+1})$. This call draws fresh random sketches, so Lemma~\ref{lem:regression_solver} applies to it for every value of its input $b_k$. Since $\eta_{k+1} \leq \epsilon_{\mathrm{final}}/(4j) < 0.1$ and $\delta_{k+1} = \delta_{\mathrm{final}}/j < 0.1$, the lemma shows that $\mathcal{F}_{k+1}$ fails with probability at most $\delta_{k+1}$, and that whenever the call stops, its output satisfies
\begin{align}\label{eq:part_1_one_step}
    \| A^\top A b_{k+1} - b_k \|_2 \leq \eta_{k+1} \| b_k \|_2 = \frac{\epsilon_{\mathrm{final}}}{4 j \kappa^{2k}} \cdot \| b_k \|_2.
\end{align}
In the proofs of Parts 1, 2 and 3, we assume that $\mathcal{E}_k$ and $\mathcal{F}_{k+1}$ both hold. Statements 1 and 2 for $i \leq k$ are part of $\mathcal{E}_k$, so it remains to prove them for $i = k+1$.

{\bf Proof of Part 1.}

We have
\begin{align}\label{eq:even_correctness}
        \| (A^\top A)^{k + 1} b_{k+1} - b_0 \|_2
        = & ~ \| (A^\top A)^{k} (A^\top A b_{k+1} - b_k) + ( (A^\top A)^{k} b_k - b_0 ) \|_2 \notag\\
        \leq & ~ \| (A^\top A)^{k}\| \cdot \| A^\top A b_{k+1} - b_k\|_2 + \|(A^\top A)^{k} b_k - b_0 \|_2 \notag\\
        \leq & ~ \sigma_{\max}(A)^{2k} \cdot \frac{\epsilon_{\mathrm{final}}}{4 j \kappa^{2k}} \cdot \| b_k\|_2 + \epsilon_k \| b_0 \|_2 \notag\\
        \leq & ~ \sigma_{\max}(A)^{2k} \cdot \frac{\epsilon_{\mathrm{final}}}{4 j \kappa^{2k}} \cdot 2 \sigma_{\min}(A)^{-2k} \| b_0 \|_2 + \epsilon_k \| b_0 \|_2 \notag\\
        = & ~ \Big( \frac{\kappa(A)}{\kappa} \Big)^{2k} \cdot \frac{\epsilon_{\mathrm{final}}}{2j} \cdot \| b_0 \|_2 + \epsilon_k \| b_0 \|_2 \notag\\
        \leq & ~ \Big( \frac{\epsilon_{\mathrm{final}}}{2j} + \epsilon_k \Big) \| b_0 \|_2 \notag\\
        = & ~ \epsilon_{k+1} \| b_0 \|_2,
    \end{align}
    where the first step follows from adding and subtracting $(A^\top A)^{k} b_k$, the second step follows from the triangle inequality and $\|Mx\|_2 \leq \|M\| \|x\|_2$, the third step follows from $\|(A^\top A)^{k}\| = \|A^\top A\|^k = \sigma_{\max}(A)^{2k}$ (because $A^\top A$ is symmetric and PSD, and by Fact~\ref{fac:norm}), Eq.~\eqref{eq:part_1_one_step} and statement 1 for $i = k$, the fourth step follows from statement 2 for $i = k$, the fifth step follows from $\kappa(A) = \sigma_{\max}(A)/\sigma_{\min}(A)$, the sixth step follows from $\kappa(A) \leq \kappa$, and the last step follows from $\epsilon_k + \epsilon_{\mathrm{final}}/(2j) = (k+1) \epsilon_{\mathrm{final}}/(2j) = \epsilon_{k+1}$.

{\bf Proof of Part 2.}

We have
\begin{align}\label{eq:leq_2b_0_2}
\|(A^\top A)^{k+1}  b_{k+1} \|_2
\leq & ~ \|(A^\top A)^{k+1}  b_{k+1} - b_0 \|_2 + \| b_0 \|_2 \notag\\
\leq & ~ (1+\epsilon_{k+1}) \|  b_0 \|_2 \notag\\
\leq & ~ 2 \| b_0 \|_2,
\end{align}
where the first step follows triangle inequality, the second step follows from Part 1, and the third step follows from $\epsilon_{k+1} \leq \epsilon_j = \epsilon_{\mathrm{final}}/2 < 1$.

Thus,
\begin{align*}
\| b_{k+1} \|_2
\leq & ~ \| ((A^\top A)^{k+1})^{-1} \cdot (A^\top A)^{k+1}  b_{k+1} \|_2 \\
\leq & ~ \| ((A^\top A)^{k+1})^{-1} \| \cdot \| (A^\top A)^{k+1}  b_{k+1} \|_2 \\
\leq & ~ \| ((A^\top A)^{k+1})^{-1} \| \cdot 2 \| b_0 \|_2 \\
\leq & ~ 2 \sigma_{\min}(A)^{-2(k+1)} \| b_0 \|_2,
\end{align*}
where the first step follows from $((A^\top A)^{k+1})^{-1} \cdot (A^\top A)^{k+1} = I$, which holds because $A$ has full column rank, the second step follows from $\|Ax\|_2 \leq \|A\| \|x\|_2$, the third step follows from Eq.~\eqref{eq:leq_2b_0_2}, and the last step follows from Fact~\ref{fac:norm}.

{\bf Proof of Part 3.}

On $\mathcal{F}_{k+1}$, the $(k+1)$-th call takes time at most $C \cdot (nd + d^3) \cdot \log(\kappa(A)/\eta_{k+1}) \cdot \log^2(n/\delta_{k+1})$. Since $\kappa(A) \leq \kappa$ and $\eta_{k+1} = \epsilon_{\mathrm{final}}/(4 j \kappa^{2k})$, we have
\begin{align*}
    \log(\kappa(A)/\eta_{k+1}) \leq \log(\kappa/\eta_{k+1}) = \log \Big( \frac{4 j \kappa^{2k+1}}{\epsilon_{\mathrm{final}}} \Big) = (2(k+1)-1) \log \kappa + \log(4j/\epsilon_{\mathrm{final}}).
\end{align*}
Since $\delta_{k+1} = \delta_{\mathrm{final}}/j$, we have $\log^2(n/\delta_{k+1}) = \log^2(jn/\delta_{\mathrm{final}})$. So the $(k+1)$-th call takes time at most
\begin{align*}
    C \cdot (nd + d^3) \cdot \Big( (2(k+1)-1) \log \kappa + \log(4j/\epsilon_{\mathrm{final}}) \Big) \cdot \log^2(jn/\delta_{\mathrm{final}}),
\end{align*}
which is the term $i = k+1$ of the sum in statement 3. Adding it to the time of the first $k$ calls, which statement 3 for $k$ bounds on $\mathcal{E}_k$, gives statement 3 for $k+1$.

{\bf Probability bound.}

By Parts 1, 2 and 3, $\mathcal{E}_{k+1}$ holds whenever $\mathcal{E}_k$ and $\mathcal{F}_{k+1}$ both hold, so by a union bound
\begin{align*}
    \Pr[\neg\mathcal{E}_{k+1}] \leq \frac{k \delta_{\mathrm{final}}}{j} + \delta_{k+1} = \frac{k \delta_{\mathrm{final}}}{j} + \frac{\delta_{\mathrm{final}}}{j} = \frac{(k+1) \delta_{\mathrm{final}}}{j}.
\end{align*}
\end{proof}

\subsection{Even Number of Matrices Regression}
\label{sub:even:main}

In this section, we present and prove our main result for even power regression.

\begin{theorem}\label{thm:even}
Let $A \in \R^{n \times d}$ be a matrix with full column rank and $b \in \R^d$ be a vector. Let $\kappa \geq \kappa(A)$ be an upper bound on the condition number of $A$ (see Definition~\ref{def:kappa}), given as input. Consider the regression problem
\begin{align*}
    \min_{x \in \R^d} \| (A^\top A)^j x - b \|_2.
\end{align*}
Let $\epsilon_{\mathrm{final}} \in (0,0.1)$ denote the accuracy parameter. Let $\delta_{\mathrm{final}} \in (0,0.1)$ denote the failure probability. There exists an algorithm (Algorithm~\ref{alg:even}) that, with probability at least $1-\delta_{\mathrm{final}}$, stops within time
\begin{align*}
O ( (n d + d^3) \cdot ( j^2 \log \kappa + j \log(j / \epsilon_{\mathrm{final}}) ) \cdot \log^2(jn/\delta_{\mathrm{final}}) )
\end{align*}
and outputs a vector $x' \in \R^d$ such that
\begin{align*}
     \| (A^\top A)^j x' - b \|_2 \leq  \epsilon_{\mathrm{final}} \| b \|_2.
\end{align*}
\end{theorem}

\begin{proof}
    We analyze Algorithm~\ref{alg:even}, with $b_0 = b$, the vectors $b_1, \ldots, b_j$ that it computes, and $\epsilon_i = i \epsilon_{\mathrm{final}}/(2j)$ for $i \in \{0, 1, \ldots, j\}$, as in Lemma~\ref{lem:induction_hypothesis}. The proof is a standard induction on $k$, which shows that $\Pr[\mathcal{E}_k] \geq 1 - k\delta_{\mathrm{final}}/j$ for every $k \in \{0, 1, \ldots, j\}$, where $\mathcal{E}_k$ is the event defined in Lemma~\ref{lem:induction_hypothesis}.

    {\bf Base case:}

    When $k = 0$, statements 1, 2 and 3 hold surely. Statement 1 reads $\|(A^\top A)^0 b_0 - b_0\|_2 = 0 = \epsilon_0 \|b_0\|_2$, and statement 2 reads $\|b_0\|_2 \leq 2 \|b_0\|_2$. No call to Algorithm~\ref{alg:fast_psd_regression} has been made, so the time in statement 3 is $0$. Hence $\Pr[\mathcal{E}_0] = 1$.

    {\bf Inductive case:}

    If $\Pr[\mathcal{E}_k] \geq 1 - k\delta_{\mathrm{final}}/j$ for some $k \in \{0, 1, \ldots, j-1\}$, then $\Pr[\mathcal{E}_{k+1}] \geq 1 - (k+1)\delta_{\mathrm{final}}/j$ by Lemma~\ref{lem:induction_hypothesis}.

    {\bf Correctness and failure probability.}

    Hence $\mathcal{E}_j$ holds with probability at least $1 - j \cdot \delta_{\mathrm{final}}/j = 1 - \delta_{\mathrm{final}}$. On this event, the output $x' = b_j$ satisfies
    \begin{align*}
        \| (A^\top A)^j x' - b \|_2 = \| (A^\top A)^j b_j - b_0 \|_2 \leq \epsilon_j \| b_0 \|_2 = \frac{\epsilon_{\mathrm{final}}}{2} \| b \|_2 \leq \epsilon_{\mathrm{final}} \| b \|_2.
    \end{align*}

    {\bf Running time.}

    On the same event, statement 3 for $k = j$ bounds the total time of the $j$ calls to Algorithm~\ref{alg:fast_psd_regression} by
    \begin{align*}
        & ~ C \cdot (nd + d^3) \cdot \sum_{i=1}^{j} \Big( (2i-1) \log \kappa + \log(4j/\epsilon_{\mathrm{final}}) \Big) \cdot \log^2(jn/\delta_{\mathrm{final}}) \\
        = & ~ C \cdot (nd + d^3) \cdot \Big( j^2 \log \kappa + j \log(4j/\epsilon_{\mathrm{final}}) \Big) \cdot \log^2(jn/\delta_{\mathrm{final}}) \\
        \leq & ~ 2C \cdot (nd + d^3) \cdot \Big( j^2 \log \kappa + j \log(j/\epsilon_{\mathrm{final}}) \Big) \cdot \log^2(jn/\delta_{\mathrm{final}}),
    \end{align*}
    where the first step follows from $\sum_{i=1}^{j} (2i-1) = j^2$, and the second step follows from $\log(4j/\epsilon_{\mathrm{final}}) = \log 4 + \log(j/\epsilon_{\mathrm{final}}) \leq 2 \log(j/\epsilon_{\mathrm{final}})$, which holds because $j/\epsilon_{\mathrm{final}} > 10 > 4$. The other lines of Algorithm~\ref{alg:even} take $O(j + d)$ time in total. This gives the stated running time.
\end{proof}

\subsection{Odd Number of Matrices Regression}
\label{sec:odd}

In this section, we analyze the odd power algorithm with its correctness and running time.

\begin{theorem}\label{thm:odd}
Let $A \in \R^{n \times d}$ be a matrix with full column rank and $b \in \R^n$ be a vector. Let $\kappa \geq \kappa(A)$ be an upper bound on the condition number of $A$ (see Definition~\ref{def:kappa}), given as input. Consider the regression problem
\begin{align*}
    \min_{x \in \R^d} \| A(A^\top A)^j x - b \|_2,
\end{align*}
and let $\OPT := \min_{x \in \R^d} \| A(A^\top A)^j x - b \|_2$ denote its optimal value.
Let $\epsilon_{\mathrm{final}} \in (0,0.1)$ denote the accuracy parameter. Let $\delta_{\mathrm{final}} \in (0,0.1)$ denote the failure probability. There exists an algorithm (Algorithm~\ref{alg:odd}) that, with probability at least $1-\delta_{\mathrm{final}}$, stops within time
\begin{align*}
O ( (n d + d^3) \cdot ( j^2 \log \kappa + j \log(j / \epsilon_{\mathrm{final}}) ) \cdot \log^2(jn/\delta_{\mathrm{final}}) )
\end{align*}
and outputs a vector $x' \in \R^d$ such that
\begin{align*}
     \| A(A^\top A)^j x' - b \|_2 \leq (1+\epsilon_{\mathrm{final}}) \OPT + \epsilon_{\mathrm{final}} \| b \|_2.
\end{align*}
\end{theorem}

The term $\OPT$ cannot be avoided. Since $A$ has full column rank, the matrix $(A^\top A)^j$ is invertible, so the vectors $A(A^\top A)^j x$ with $x \in \R^d$ form exactly the column space of $A$. Hence $\OPT = \|(I - AA^\dagger) b\|_2$ is the distance from $b$ to that space, and no output has a smaller residual. When $b$ lies in the column space of $A$, we have $\OPT = 0$, and the bound reads $\| A(A^\top A)^j x' - b \|_2 \leq \epsilon_{\mathrm{final}} \| b \|_2$.

\begin{proof}

For convenience, we use $b_{\odd}$ to denote $b$. We define $\OPT$ as
\begin{align*}
\OPT:=  \min_{x \in \R^d} \| A (A^\top A)^j x - b_{\odd}\|_2 .
\end{align*}

    First, we use Algorithm~\ref{alg:iter_regression} to solve 
    \begin{align}\label{eq:Ab_2_b_3:odd}
        \min_{y \in \R^d} \| A y - b_{\odd} \|_2.
    \end{align}

    Let $y_*$ denote the exact solution to this regression problem. By Lemma~\ref{lem:alg_takes_time}, we can get $y' \in \R^d$ such that the following holds with probability $1-\delta_1$,
    \begin{align}\label{eq:Ay_b_3:odd}
        \| A y' - b_{\odd} \|_2 \leq & ~ (1+\epsilon_1) \cdot \min_{y \in \R^d} \| A y - b_{\odd} \|_2 \notag \\
        \leq & ~ (1+\epsilon_1) \cdot \OPT,
    \end{align}
    where the last step follows from $A(A^\top A)^j x = Ay$ for $y = (A^\top A)^j x$, so $A(A^\top A)^j x$ cannot do a better job than the minimizer $y$ (in terms of minimizing cost). Since $\epsilon_1 = 0.1 \epsilon_{\final} \in (0, 0.01)$ and $\delta_1 = \delta_{\final}/2 \in (0, 0.05)$, this step takes time
    \begin{align*}
        O((nd +d^3 ) \cdot \log(1/\epsilon_1) \cdot  \log^2(n/\delta_1) ) = O((nd +d^3 ) \cdot \log(1/\epsilon_{\final}) \cdot  \log^2(n/\delta_{\final}) ).
    \end{align*}

By triangle inequality, we can show
\begin{align}\label{eq:||y'||_2:odd}
\| y'\|_2 
= & ~ \|y' - y_* + y_* \|_2 \notag \\
\leq & ~ \|y' - y_* \|_2 + \| y_* \|_2.
\end{align}

To bound the first term of Eq.~\eqref{eq:||y'||_2:odd}, we have
\begin{align}\label{eq:bound_y'_y_*:odd}
    \|y' - y_* \|_2
    \leq & ~ 2\sqrt{\epsilon_1} \cdot \sigma_{\min}(A)^{-1} \| A y_* - b_{\odd} \|_2 \notag\\
    \leq & ~ 2\sqrt{\epsilon_1} \cdot \sigma_{\min}(A)^{-1} \OPT,
\end{align}
where the first step follows from Lemma~\ref{lem:forward_erorr} (its proof gives the constant $2$), and the second step follows from $\| A y_* - b_{\odd} \|_2 = \min_{y \in \R^d} \| A y - b_{\odd} \|_2 \leq \OPT$, as in Eq.~\eqref{eq:Ay_b_3:odd}.

To bound the second term of Eq.~\eqref{eq:||y'||_2:odd}, we have
\begin{align}\label{eq:bound_y_*:odd}
    \| y_* \|_2
    = & ~ \| A^\dagger b_{\odd} \|_2 \notag\\
    \leq & ~ \| A^\dagger\| \cdot \| b_{\odd} \|_2 \notag \\
    \leq & ~ \sigma_{\min}(A)^{-1} \cdot \| b_{\odd} \|_2,
\end{align}
where the first step follows from $ y_* = A^\dagger b_{\odd} $, the second step follows from $\|Ax\|_2 \leq \|A\| \|x\|_2$, and the third step follows from $\| A^\dagger\| = \sigma_{\min}(A)^{-1}$.

By plugging Eq.~\eqref{eq:bound_y'_y_*:odd} and Eq.~\eqref{eq:bound_y_*:odd} into Eq.~\eqref{eq:||y'||_2:odd}, we have
\begin{align}\label{eq:finish_||y'||_2:odd}
    \| y'\|_2 \leq 2\sqrt{\epsilon_1} \cdot \sigma_{\min}(A)^{-1} \OPT + \sigma_{\min}(A)^{-1} \| b_{\odd} \|_2.
\end{align}

Let $b_{\even} = y' \in \R^d$ and $x' \in \R^d$. Then, using Algorithm~\ref{alg:even} with $\epsilon_{\even} = \epsilon_{\final}/\kappa$ and $\delta_{\even} = \delta_{\final}/2$, we solve
    \begin{align*}
        \min_{x' \in \R^d} \| (A^\top A)^j x' - b_{\even} \|_2.
    \end{align*}

    Since $\epsilon_{\even} \leq \epsilon_{\final} < 0.1$ and $\delta_{\even} < 0.1$, Theorem~\ref{thm:even} applies, and we find an $x'\in \R^d$ such that
    \begin{align}\label{eq:A^topAx'-b_2_leq_epsilon_2_b_2_2:odd}
        \| (A^\top A)^j x' - b_{\even} \|_2 \leq \epsilon_{\even}  \|b_{\even}\|_2
    \end{align}
    holds with probability $1-\delta_{\even}$.

    This step takes time
    \begin{align*}
        & ~ O((nd + d^3 ) \cdot ( j^2 \log \kappa + j \log(j/ \epsilon_{\even}) ) \cdot \log^2(jn/\delta_{\even}) ) \\
        = & ~ O((nd + d^3 ) \cdot ( j^2 \log \kappa + j \log(j \kappa/ \epsilon_{\final}) ) \cdot \log^2(jn/\delta_{\final}) ) \\
        = & ~ O((nd + d^3 ) \cdot ( j^2 \log \kappa + j \log(j/ \epsilon_{\final}) ) \cdot \log^2(jn/\delta_{\final}) ),
    \end{align*}
    where the first step follows from $\epsilon_{\even} = \epsilon_{\final}/\kappa$ and $\log(2jn/\delta_{\final}) \leq 2 \log(jn/\delta_{\final})$, which holds because $jn/\delta_{\final} > 10 > 2$, and the second step follows from $j \log(j \kappa/\epsilon_{\final}) = j \log \kappa + j \log(j/\epsilon_{\final})$ and $j \log \kappa \leq j^2 \log \kappa$.

{\bf Correctness.}

To bound $\| A (A^\top A)^j x' - b_{\odd} \|_2$, we have 
\begin{align}\label{eq:AA^topAx'_b_3:odd}
 \| A (A^\top A)^j x' - b_{\odd} \|_2
 = & ~ \| A (A^\top A)^j x' - A y' + Ay' - b_{\odd} \|_2 \notag \\
 \leq & ~ \| A (A^\top A)^j x' - A y' \|_2 + \| Ay'-b_{\odd} \|_2 \notag \\
 \leq & ~ \| A (A^\top A)^j x' - A y' \|_2  + (1+\epsilon_1) \cdot \OPT,
 \end{align}
 where the first step follows from adding and subtracting the same thing, the second step follows from the triangle inequality, and the third step follows from Eq.~\eqref{eq:Ay_b_3:odd}. Let's consider the first term of Eq.~\eqref{eq:AA^topAx'_b_3:odd}. We have
 \begin{align}\label{eq:bound_AA^topAx'_Ay':odd}
     \| A (A^\top A)^j x' - A y' \|_2
     = & ~ \| A ((A^\top A)^j x' - y') \|_2 \notag \\
     \leq & ~ \|A \| \cdot \| (A^\top A)^j x' - y' \|_2 \notag \\
     \leq & ~ \| A \| \cdot \epsilon_{\even}  \| y' \|_2 \notag \\
     \leq & ~ \| A \| \cdot \epsilon_{\even}  \left(2\sqrt{\epsilon_1}  \cdot \sigma_{\min}(A)^{-1} \OPT + \sigma_{\min}(A)^{-1} \| b_{\odd} \|_2 \right) \notag \\
     = & ~ \| A \| \cdot \epsilon_{\even} \cdot 2\sqrt{\epsilon_1} \cdot \sigma_{\min}(A)^{-1} \OPT + \| A \| \cdot \epsilon_{\even}  \sigma_{\min}(A)^{-1} \| b_{\odd} \|_2,
 \end{align}
 where the first step follows from simple algebra, the second step follows from $\|Ax\|_2 \leq \|A\| \|x\|_2$, the third step follows from Eq.~\eqref{eq:A^topAx'-b_2_leq_epsilon_2_b_2_2:odd}, the fourth step follows from Eq.~\eqref{eq:finish_||y'||_2:odd}, and the last step follows from simple algebra.

 Then, to bound the first term of Eq.~\eqref{eq:bound_AA^topAx'_Ay':odd}, we have
 \begin{align}\label{eq:||A||_epsilon_2_O:odd}
     \| A \| \cdot \epsilon_{\even} \cdot 2\sqrt{\epsilon_1} \cdot \sigma_{\min}(A)^{-1} \OPT
     \leq & ~ \sigma_{\max}(A)\sigma_{\min}(A)^{-1} \cdot \epsilon_{\even} \cdot 2\sqrt{\epsilon_1} \cdot \OPT \notag \\
     = & ~ 2\sqrt{\epsilon_1} \cdot \kappa(A) \epsilon_{\even} \cdot \OPT \notag \\
     \leq & ~ 2\sqrt{\epsilon_1} \epsilon_{\final} \cdot \OPT,
 \end{align}
 where the first step follows from $\| A \| \leq \sigma_{\max}(A)$, the second step follows from the definition of $\kappa(A)$ (see Definition~\ref{def:kappa}), and the last step follows from $\kappa(A) \epsilon_{\even} = \kappa(A) \epsilon_{\final} / \kappa \leq \epsilon_{\final}$, since $\kappa(A) \leq \kappa$. Similarly, to bound the second term of Eq.~\eqref{eq:bound_AA^topAx'_Ay':odd}, we get
 \begin{align}\label{eq:||A||_epsilon_2_sigma:odd}
     \| A \| \cdot \epsilon_{\even}  \sigma_{\min}(A)^{-1} \| b_{\odd} \|_2
     \leq & ~ \sigma_{\max}(A) \sigma_{\min}(A)^{-1} \cdot \epsilon_{\even}  \| b_{\odd} \|_2 \notag \\
    = & ~ \kappa(A) \cdot \epsilon_{\even}  \| b_{\odd} \|_2 \notag \\
    \leq & ~ \epsilon_{\final}  \| b_{\odd} \|_2,
 \end{align}
 where the first step follows from $\| A \| \leq \sigma_{\max}(A)$, the second step follows from the definition of $\kappa(A)$ (see Definition~\ref{def:kappa}), and the last step follows from $\kappa(A) \epsilon_{\even} = \kappa(A) \epsilon_{\final} / \kappa \leq \epsilon_{\final}$.

 Plugging Eq.~\eqref{eq:||A||_epsilon_2_O:odd} and Eq.~\eqref{eq:||A||_epsilon_2_sigma:odd} into Eq.~\eqref{eq:bound_AA^topAx'_Ay':odd}, we get
 \begin{align}\label{eq:finish_bound_AA^topAx'_Ay':odd}
     \| A (A^\top A)^j x' - A y' \|_2 \leq 2\sqrt{\epsilon_1} \epsilon_{\final} \cdot \OPT + \epsilon_{\final}  \| b_{\odd} \|_2.
 \end{align}

 Therefore, by plugging Eq.~\eqref{eq:finish_bound_AA^topAx'_Ay':odd} into \eqref{eq:AA^topAx'_b_3:odd}, we have
 \begin{align*}
     \| A (A^\top A)^j x' - b_{\odd} \|_2
     \leq & ~ 2\sqrt{\epsilon_1} \epsilon_{\final} \cdot \OPT + \epsilon_{\final}  \| b_{\odd} \|_2 + (1+\epsilon_1) \cdot \OPT \\
     = & ~ (1 + \epsilon_1 + 2\sqrt{\epsilon_1} \epsilon_{\final}) \cdot \OPT + \epsilon_{\final}  \| b_{\odd} \|_2 \\
     \leq & ~ (1 + 0.1 \epsilon_{\final} + 0.2 \epsilon_{\final}) \cdot \OPT + \epsilon_{\final}  \| b_{\odd} \|_2 \\
     \leq & ~ (1+\epsilon_{\final} ) \cdot \OPT + \epsilon_{\final}  \| b_{\odd} \|_2,
 \end{align*}
where the second step follows from simple algebra, the third step follows from $\epsilon_1 = 0.1 \epsilon_{\final}$ and $\sqrt{\epsilon_1} \leq 0.1$, which holds because $\epsilon_1 = 0.1 \epsilon_{\final} \leq 0.01$, and the last step follows from $0.3 \epsilon_{\final} \leq \epsilon_{\final}$. Therefore, we complete bounding $\| A (A^\top A)^j x' - b_{\odd} \|_2$.

{\bf Running time.}

Adding the times of the two steps, and using $\log(1/\epsilon_{\final}) \leq \log(j/\epsilon_{\final})$ and $\log(n/\delta_{\final}) \leq \log(jn/\delta_{\final})$, the overall running time is
\begin{align*}
        O((nd+d^3) \cdot ( j^2 \log \kappa + j \log(j /\epsilon_{\final} ) ) \cdot \log^2(jn/\delta_{\final})).
\end{align*}
This has the same order as the running time in Theorem~\ref{thm:even}.

{\bf Failure probability.} 

By taking a union over two events, the failure probability is at most $\delta_1 + \delta_{\even} =\delta_{\final} $.
\end{proof}


\section{Attention Kernel Regression}
\label{sec:attention_kernel}

In this section, we demonstrate algorithms for solving attention kernel regressions, inspired by algorithms for Gaussian kernel regression. We follow the approach of~\cite{swyz21} for Gaussian kernel regression, which sketches the kernel matrix with a sketch dimension $m \geq n$ and then solves the regression against the sketched kernel with a preconditioned solver. In Section~\ref{sec:kernel_below_n}, we give a second algorithm whose sketch dimension can be smaller than $n$. It sketches the kernel plus a ridge term, and its error is relative to the best residual among solutions of norm at most $(K/n)\|b\|_2$, for a parameter $K \geq 1$ given as input.

\subsection{Fast Subspace Embedding for Tensors}
We need the following tools from prior results. We review an algorithm due to~\cite{swyz21}.

\begin{algorithm}[!ht]
\caption{Algorithm of~\cite{swyz21}.}\label{alg:1_in_swyz21}
\begin{algorithmic}[1]
\Procedure{TensorSketchViaLimRand}{$x \in \R^d$, $p \in (1, \infty)$, $S \in \R^{m \times m^2}$, $T \in \R^{m \times d}$} \Comment{Theorem~\ref{thm:4.8_in_swyz21} and~\ref{thm:5.1_in_swyz21}}
\State $q \gets 2^{\lfloor\log_2 p\rfloor}$ \label{line:def_q} 
\State $w_0 \gets Tx$ \label{line:apply_T}
\Comment{$T$ can be $\mathsf{SRHT}$ (Definition~\ref{def:SRHT})}
\For{$l = 1$ to $\log_2 q$}
\State $w_l \gets S(w_{l-1}^{\otimes 2})$ 
\Comment{$S$ can be $\mathsf{TensorSRHT}$ (Definition~\ref{def:tensor_SRHT})}
\EndFor
\State \Comment{Let $b$ be the binary representation of $p$, and let $E = \{i : b_i = 1, i \in \{0, \dots, \log_2 p\}\}$}
\State \Comment{Let $z = w_j$, where $j$ is the lowest bit of $b$ where $b_j = 1$}
\For{$i$ in $E \setminus \{j\}$}
\State $z = S(z \otimes w_i)$
\EndFor
\State \Return $z$ \Comment{$z \in \R^m$}
\EndProcedure
\end{algorithmic}
\end{algorithm}

Before stating the guarantees of this algorithm, we need some auxiliary definitions.
\begin{definition}\label{def:4.2_in_swyz21}
    Let $S \in \R^{m^2} \to \R^m$ and $T : \R^d \to \R^m$ be base sketches. Let $X \in \R^{d \times n}$ be an input matrix. We define $\mathcal{Z}(S, T, X)$ to be the matrix for which we apply Algorithm~\ref{alg:1_in_swyz21} on each column of $X$, with base sketches $S$ and $T$.
\end{definition}

\begin{theorem}[Theorem 4.8 in \cite{swyz21}]\label{thm:4.8_in_swyz21}
    Let 
    \begin{align*}
        S: \R^{m^2} \to \R^m
    \end{align*}
    be an $(n, d, \epsilon, \delta)$-$\mathsf{SE}$ for degree-2 tensors and 
    \begin{align*}
        T: \R^d \to \R^m
    \end{align*}
    be an $(n, d, \epsilon, \delta)$-$\mathsf{SE}$. Let $p$ be a positive integer, $Z = \mathcal{Z}(S, T, X)$ be the matrix as defined in Definition~\ref{def:4.2_in_swyz21}. Then for any $y \in \R^n$, we have
    \begin{align*}
        (1-\epsilon)^{3p} \cdot \|X^{\otimes p} y\|_2 \leq \|Zy\|_2 \leq (1+\epsilon)^{3p} \cdot \|X^{\otimes p} y\|_2.
    \end{align*}
\end{theorem}

The next theorem states Algorithm~\ref{alg:1_in_swyz21} can be quickly applied to a structured matrix.

\begin{theorem}[Theorem 5.1 in \cite{swyz21}, with the logarithmic factors written out]\label{thm:5.1_in_swyz21}
    Let $p \in \mathbb{Z}_+$ and let $\epsilon, \delta \in (0,1)$. For every $X \in \R^{d \times n}$, there exists a distribution over oblivious linear sketches $\Pi : \R^{d^p} \to \R^m$ such that if
    \begin{align*}
        m = \Theta(\epsilon^{-2} n p^2 \log^2(nd/(\epsilon\delta)) \log(n/\delta)),
    \end{align*}
    then with probability at least $1-\delta$, we have
    \begin{align*}
        (1-\epsilon) \cdot (X ^{\otimes^p})^\top X ^{\otimes^p} \preceq (\Pi X^{\otimes^p})^\top \Pi X^{\otimes^p} \preceq (1+\epsilon) \cdot (X ^{\otimes^p})^\top X ^{\otimes^p}.
    \end{align*}

    Moreover, using Algorithm~\ref{alg:1_in_swyz21},
    \begin{align*}
        \Pi X^{\otimes^p} = \mathcal{Z}(S,T,X)
    \end{align*}
    can be computed in time
    \begin{align*}
        O(nd \log(nd/(\epsilon\delta)) + \epsilon^{-2} n^2 p^2 \log^2(nd/(\epsilon\delta)) \log(n/\delta)).
    \end{align*}

\end{theorem}

Following the proof of Theorem A.1 in~\cite{swyz21}, we prove a result regarding the exponential kernel.

\begin{theorem}\label{thm:gaussian_kernel}

    For every $r > 0$, every positive integers $n, d$, and every $X \in \R^{d \times n}$ such that $\|x_i\|_2 \leq r$ for all $i \in [n]$, where $x_i$ is the $i$-th column of $X$, suppose $K \in \R^{n \times n}$ is the attention kernel matrix i.e.,
    \begin{align*}
        K_{j,k} = e^{\langle x_j, x_k\rangle}
    \end{align*}
    for all $j, k \in [n]$. Suppose that $K$ is positive definite, factor $K = Z^\top Z$, and let $\kappa \geq \kappa(Z)$. Let $\epsilon, \delta \in (0, 0.1)$. Let $q$ be the smallest integer such that
    \begin{align*}
        q + 1 \geq \max\{ 2e r^2, \log_2(4n\kappa^2/\epsilon) \},
    \end{align*}
    so that $q = \Theta(r^2 + \log(n\kappa/\epsilon))$, and let
    \begin{align*}
        m = \Theta(\epsilon^{-2} n q^3 \log^2(ndq/(\epsilon\delta)) \log(nq/\delta)) = \wt{\Theta}(q^3 n/\epsilon^2).
    \end{align*}
    There exists an algorithm which computes $W_g(X) \in \R^{m \times n}$ in time
    \begin{align*}
        O(mn + q \cdot nd\log(ndq/(\epsilon\delta)))
    \end{align*}
    such that
    \begin{align*}
        \Pr_{W_g}[(1 - \epsilon)K \preceq (W_g(X))^\top W_g(X) \preceq (1 + \epsilon)K] \geq 1 - \delta.
    \end{align*}
\end{theorem}

\begin{proof}
    Note that the Taylor series expansion for kernel $K$ gives
    \begin{align*}
        K = \sum\limits_{l=0}^{\infty} \frac{(X^{\otimes l})^\top X^{\otimes l}}{l!},
    \end{align*}
    because the $(j,k)$-th entry of $(X^{\otimes l})^\top X^{\otimes l}$ is $\langle x_j, x_k\rangle^l$. Each term $(X^{\otimes l})^\top X^{\otimes l}/l!$ of this series is PSD.

    Let $q$ be as in the statement. Since $\log_2(4n\kappa^2/\epsilon) > \log_2 40 > 5$, we have $q \geq 5$. Let
    \begin{align*}
        Q = \sum_{l=0}^{q}\frac{(X^{\otimes l})^{\top}X^{\otimes l}}{l!}
    \end{align*}
    be the sum of the first $q+1$ terms of $K$.

    {\bf Truncation error.} For every $l \geq 0$, every entry of $(X^{\otimes l})^\top X^{\otimes l}$ has absolute value $|\langle x_j, x_k\rangle|^l \leq r^{2l}$, so $\|(X^{\otimes l})^\top X^{\otimes l}\|_F \leq n r^{2l}$. For every $l \geq q+1$, we have
    \begin{align*}
        \frac{r^{2l}}{l!} = \frac{r^{2(q+1)}}{(q+1)!} \cdot \prod_{i=q+2}^{l} \frac{r^2}{i} \leq \frac{r^{2(q+1)}}{(q+1)!} \cdot \Big( \frac{r^2}{q+2} \Big)^{l-q-1},
    \end{align*}
    and $r^2/(q+2) < r^2/(q+1) \leq 1/(2e) < 1/2$ because $q+1 \geq 2er^2$. Moreover, $(q+1)! \geq ((q+1)/e)^{q+1}$. By triangle inequality, we have:
    \begin{align*}
        \|K - Q\|
        \leq & ~ \sum_{l>q} \| \frac{(X^{\otimes l})^\top X^{\otimes l}}{l!}\|  \\
        \leq & ~ \sum_{l>q} \| \frac{(X^{\otimes l})^\top X^{\otimes l}}{l!}\|_F  \\
        \leq & ~ \sum_{l>q} \frac{n \cdot r^{2l}}{l!} \\
        \leq & ~ \frac{n \cdot r^{2(q+1)}}{(q+1)!} \cdot \sum_{k=0}^{\infty} 2^{-k} \\
        \leq & ~ 2n \cdot \Big( \frac{e r^2}{q+1} \Big)^{q+1} \\
        \leq & ~ 2n \cdot 2^{-(q+1)} \\
        \leq & ~ \frac{\epsilon}{2\kappa^2},
    \end{align*}
    where the first step follows from the definition of $Q$, the second step follows from $\|A\| \leq \|A\|_F$ for all matrix $A$, the third step follows from the bound on the Frobenius norm above, the fourth step follows from the bound on $r^{2l}/l!$ above and $r^2/(q+2) < 1/2$, the fifth step follows from $\sum_{k=0}^{\infty} 2^{-k} = 2$ and $(q+1)! \geq ((q+1)/e)^{q+1}$, the sixth step follows from $e r^2/(q+1) \leq 1/2$, and the last step follows from $q+1 \geq \log_2(4n\kappa^2/\epsilon)$, that is, $2^{-(q+1)} \leq \epsilon/(4n\kappa^2)$.

    {\bf From an additive to a relative bound.} Since $K = Z^\top Z$, we have $\sigma_{\max}(Z)^2 = \lambda_{\max}(K) = \|K\|$ and $\sigma_{\min}(Z)^2 = \lambda_{\min}(K)$, so $\kappa(Z)^2 = \|K\|/\lambda_{\min}(K)$. Moreover, $\|K\| \geq e_1^\top K e_1 = K_{1,1} = \exp(\|x_1\|_2^2) \geq 1$. Hence
    \begin{align*}
        \lambda_{\min}(K) = \frac{\|K\|}{\kappa(Z)^2} \geq \frac{1}{\kappa^2}.
    \end{align*}
    The matrix $K - Q = \sum_{l > q} (X^{\otimes l})^\top X^{\otimes l}/l!$ is PSD, as a sum of PSD terms. So
    \begin{align*}
        0 \preceq K - Q \preceq \|K - Q\| \cdot I \preceq \frac{\epsilon}{2\kappa^2} \cdot I \preceq \frac{\epsilon}{2} \cdot \lambda_{\min}(K) \cdot I \preceq \frac{\epsilon}{2} \cdot K,
    \end{align*}
    which gives
    \begin{align}\label{eq:Q_vs_K}
        (1 - \epsilon/2) \cdot K \preceq Q \preceq K.
    \end{align}

    {\bf Sketching each term.} For $l = 0$, the matrix $X^{\otimes 0} = {\bf 1}_n^\top \in \R^{1 \times n}$ is the all-ones row, and we set $Z_0 := X^{\otimes 0}$ and $m_0 := 1$. For each term $(X^{\otimes l})^\top X^{\otimes l}$ with $1 \leq l \leq q$, we run Algorithm~\ref{alg:1_in_swyz21} to approximate $X^{\otimes l}$.

Let $Z_l = \Pi^l X^{\otimes l} \in \R^{m_l \times n}$ be the resulting matrix $\mathcal{Z}(S, T, X)$, where
\begin{align*}
    m_l = \Theta(\epsilon^{-2} n l^2 \log^2(nd(q+1)/(\epsilon\delta)) \log(n(q+1)/\delta)).
\end{align*}
Then by Theorem~\ref{thm:5.1_in_swyz21} with accuracy $\epsilon/2$ and failure probability $\delta/(q+1)$, we get
\begin{align}\label{eq:4_in_swyz21}
    (1 - \epsilon/2)(X^{\otimes l})^\top X^{\otimes l} \preceq (\Pi^l X^{\otimes l}  )^\top \Pi^l X^{\otimes l}   \preceq (1 + \epsilon/2)(X^{\otimes l} )^\top X^{\otimes l}
\end{align}
with probability at least $1 - \frac{\delta}{q + 1}$. For $l = 0$, Eq.~\eqref{eq:4_in_swyz21} holds with $Z_0$ in place of $\Pi^0 X^{\otimes 0}$ and with probability $1$.

Moreover, apart from applying $T$ to $X$, which we discuss below, $Z_l$ can be computed in time
\begin{align*}
    O(\epsilon^{-2}n^2 l^{2}\cdot\log^{2}(nd(q+1)/(\epsilon\delta))\cdot\log(n(q+1)/\delta)) = O(n m_l).
\end{align*}
Our algorithm will simply compute $Z_l$ from $l = 0$ to $q$, normalize each $Z_l$ by $\frac{1}{\sqrt{l!}}$.

More precisely, the approximation $W_g(X)$ will be
\begin{align*}
    W_g(X) = (\oplus_{l = 0}^q \frac{Z_l}{\sqrt{l!}}).
\end{align*}

Notice $W_g(X) \in \R^{m \times n}$ with $m = \sum_{l=0}^{q} m_l$.

The following holds for $W_g(X)^\top W_g(X)$:
\begin{align*}
    W_g(X)^\top W_g(X)
    = & ~ \sum_{l=0}^{q} \frac{Z_l^\top Z_l}{l!}.
\end{align*}

By multiplying Eq.~\eqref{eq:4_in_swyz21} by $1/l!$, summing over $0 \leq l \leq q$, and using a union bound over the $q$ random terms $1 \leq l \leq q$, we obtain that with probability at least $1 - q\delta/(q+1) \geq 1 - \delta$, we have the following:
\begin{align*}
    (1 - \epsilon/2) \cdot Q\preceq W_g(X)^{\top} W_g(X) \preceq (1 + \epsilon/2) \cdot Q.
\end{align*}

Combining this with Eq.~\eqref{eq:Q_vs_K}, we get
\begin{align*}
    W_g(X)^{\top} W_g(X) \succeq (1-\epsilon/2) \cdot Q \succeq (1-\epsilon/2)^2 \cdot K \succeq (1-\epsilon) \cdot K,
\end{align*}
where the last step follows from $(1-\epsilon/2)^2 = 1 - \epsilon + \epsilon^2/4 \geq 1-\epsilon$, and
\begin{align*}
    W_g(X)^{\top} W_g(X) \preceq (1+\epsilon/2) \cdot Q \preceq (1+\epsilon/2) \cdot K \preceq (1+\epsilon) \cdot K.
\end{align*}
Thus, we conclude that
\begin{align*}
    (1 - \epsilon) \cdot K \preceq W_g(X)^{\top} W_g(X) \preceq (1 + \epsilon) \cdot K.
\end{align*}

Note the target dimension of $W_g$ is
\begin{align*}
    m
    = & ~ \sum_{l=0}^{q} m_{l} \\
    = & ~ 1 + \sum_{l=1}^{q} \Theta(\epsilon^{-2} n l^2 \log^2(nd(q+1)/(\epsilon\delta)) \log(n(q+1)/\delta)) \\
    = & ~ \Theta(\epsilon^{-2}nq^{3} \cdot \log^{2}(ndq/(\epsilon\delta)) \cdot \log(nq/\delta)),
\end{align*}
where the first step follows from the construction of $W_g(X)$, the second step follows from the choice of $m_l$, and the last step follows from $\sum_{l=1}^{q} l^2 = q(q+1)(2q+1)/6 = \Theta(q^3)$ and $q < q+1 \leq 2q$. This is the sketch dimension in the statement.

Also, apart from applying $T$ to $X$, the time to compute $Z_0, \ldots, Z_q$ is
\begin{align*}
    t
    = & ~ \sum_{l=0}^{q} O(n m_l) \\
    = & ~ O(nm).
\end{align*}

Applying $\mathsf{SRHT}$ to $X$ (line~\ref{line:apply_T} of Algorithm~\ref{alg:1_in_swyz21}) costs $O(nd\log(ndq/(\epsilon\delta)))$ for each of the $q$ degrees, so the final runtime is
\begin{align*}
O(mn + q \cdot nd\log(ndq/(\epsilon\delta))).
\end{align*}
\end{proof}

\begin{remark}\label{rem:kernel_rank_n}
No sketch with $m < n$ rows can satisfy the guarantee of Theorem~\ref{thm:gaussian_kernel}. Since $K$ is positive definite, $(1-\epsilon) K \preceq W_g(X)^\top W_g(X)$ gives $\|W_g(X) x\|_2^2 \geq (1-\epsilon) \cdot x^\top K x > 0$ for every $x \neq 0$, so $W_g(X)$ has rank $n$, which forces $m \geq n$. With a ridge term, this obstruction disappears (Theorem~\ref{thm:below_n}).
\end{remark}

\subsection{Main Result}
\begin{theorem}[Formal version of Theorem~\ref{thm:informal_exp}]\label{thm:formal_exp}
    Let $A \in \R^{n \times d}$ and $X := A^\top \in \R^{d \times n}$, and assume that for all $i \in [n]$, the $i$-th column $x_i$ of $X$ satisfies $\|x_i\|_2 \leq 1$. Let $G \in \R^{n \times n}$ be the Attention kernel matrix (Definition~\ref{def:attention_kernel}) for $X$, that is, $G = \exp(AA^\top)$. Assume that $G$ is positive definite, which holds whenever $x_1, \ldots, x_n$ are distinct. Factor $G=Z^\top Z$. The condition number $\kappa(Z) = \sqrt{\lambda_{\max}(G)/\lambda_{\min}(G)}$ does not depend on the choice of the factor $Z$. Let $\kappa \geq \kappa(Z)$ be an upper bound given as input, let $\epsilon, \delta \in (0,0.1)$, and let $y \in \R^n$. Let $m$ be the sketch dimension of Theorem~\ref{thm:gaussian_kernel} with $r = 1$, accuracy $\epsilon/(4\kappa)$ and failure probability $\delta/2$, which satisfies
    \begin{align*}
        m = O(\epsilon^{-2} \kappa^2 n \log^3(n\kappa/\epsilon) \log^3(nd\kappa/(\epsilon\delta))).
    \end{align*}
    Then Algorithm~\ref{alg:alg_2_in_swyz21}, run on $A$, $b := y$, $\kappa$, $\epsilon_{\rm final} := \epsilon$ and $\delta_{\rm final} := \delta$, with probability at least $1 - \delta$, computes an $\wh{x} \in \R^n$ satisfying
    \begin{align*}
        \|G\wh{x} - y\|_2 \leq \epsilon \|y\|_2
    \end{align*}
    in time
    \begin{align*}
       O( (mn + n^3) \cdot \log(\kappa/\epsilon) \cdot \log^2(m/\delta) + nd \log(n\kappa/\epsilon) \log(nd\kappa/(\epsilon\delta)) ).
    \end{align*}
\end{theorem}
\begin{proof}
    Let $\epsilon' := \epsilon/(4\kappa)$, as in Algorithm~\ref{alg:alg_2_in_swyz21}, and write $W_g := W_g(X)$. Since $\kappa \geq \kappa(Z) \geq 1$, we have $\epsilon' \leq \epsilon/4 < 0.025$.

    {\bf Positive definiteness.} We first explain why $G$ is positive definite when $x_1, \ldots, x_n$ are distinct. For all $i, j \in [n]$, we have
    \begin{align*}
        \langle x_i, x_j\rangle = \frac{1}{2} \|x_i\|_2^2 + \frac{1}{2} \|x_j\|_2^2 - \frac{1}{2} \|x_i - x_j\|_2^2.
    \end{align*}
    Hence $G = D G' D$, where $D$ is the diagonal matrix with $D_{i,i} = \exp(\|x_i\|_2^2/2)$ for all $i \in [n]$, and $G'$ is the Gaussian kernel matrix on $X$ (Definition~\ref{def:gaussian_kernel}), so $G$ is positive definite if and only if $G'$ is. For $c \in \R^n$, the Taylor series in the proof of Theorem~\ref{thm:gaussian_kernel} gives $c^\top G c = \sum_{l \geq 0} \|\sum_{i=1}^n c_i x_i^{\otimes l}\|_2^2 / l!$. If this is $0$, then $\sum_i c_i x_i^{\otimes l} = 0$ for all $l$. Every monomial of degree $l$ is an entry of $x^{\otimes l}$, so $\sum_i c_i p(x_i) = 0$ for every polynomial $p$ on $\R^d$. Fix $k$, pick $u$ with $\langle u, x_k - x_i\rangle \neq 0$ for all $i \neq k$ (possible since the points are distinct), and take $p(x) = \prod_{i \neq k}\langle u, x - x_i\rangle$. Then $c_k p(x_k) = 0$ with $p(x_k) \neq 0$, so $c_k = 0$. Hence $G$ is positive definite. Conversely, if $x_i = x_j$ for some $i \neq j$, two rows of $G$ coincide and $G$ is singular.

    {\bf Step 1: the sketch.} By Theorem~\ref{thm:gaussian_kernel} with $r = 1$, accuracy $\epsilon'$, failure probability $\delta/2$ and the bound $\kappa \geq \kappa(Z)$, the event
    \begin{align}\label{eq:W_g_spectral}
        (1-\epsilon') \cdot G \preceq W_g^\top W_g \preceq (1+\epsilon') \cdot G
    \end{align}
    holds with probability at least $1-\delta/2$. We call it the sketching event, and we assume it in Steps 1, 2 and 3. For every $x \in \R^n$, we have $x^\top G x = \|Zx\|_2^2$ and $x^\top W_g^\top W_g x = \|W_g x\|_2^2$, so Eq.~\eqref{eq:W_g_spectral} gives
    \begin{align*}
        \sqrt{1-\epsilon'} \cdot \|Zx\|_2 \leq \|W_g x\|_2 \leq \sqrt{1+\epsilon'} \cdot \|Zx\|_2.
    \end{align*}
    Taking the minimum and the maximum over unit vectors $x$ gives $\sigma_{\min}(W_g) \geq \sqrt{1-\epsilon'} \cdot \sigma_{\min}(Z)$ and $\sigma_{\max}(W_g) \leq \sqrt{1+\epsilon'} \cdot \sigma_{\max}(Z)$. Since $\sigma_{\min}(Z)^2 = \lambda_{\min}(G) > 0$, the matrix $W_g \in \R^{m \times n}$ has full column rank, which forces $m \geq n$. Moreover,
    \begin{align*}
        \kappa(W_g) \leq \sqrt{\frac{1+\epsilon'}{1-\epsilon'}} \cdot \kappa(Z) \leq 2\kappa.
    \end{align*}

    {\bf Step 2: the solver.} Algorithm~\ref{alg:alg_2_in_swyz21} computes $\wh{x} = \textsc{FastPSDRegression}(W_g, y, m, n, \epsilon/2, \delta/2)$. We apply Lemma~\ref{lem:regression_solver} with the matrix $W_g \in \R^{m \times n}$ in place of $A$, that is, with $(n, d)$ replaced by $(m, n)$. Its hypothesis holds because $W_g$ has full column rank. The solver draws fresh random sketches, so the lemma applies to it for every realization of $W_g$ in the sketching event. By the lemma, whenever the solver stops, its output satisfies
    \begin{align}\label{eq:sketched_residual}
        \|W_g^\top W_g \wh{x} - y\|_2 \leq \frac{\epsilon}{2} \cdot \|y\|_2,
    \end{align}
    and with probability at least $1-\delta/2$, the solver stops within time
    \begin{align*}
        O((mn + n^3) \cdot \log(2\kappa(W_g)/\epsilon) \cdot \log^2(2m/\delta)) = O((mn + n^3) \cdot \log(\kappa/\epsilon) \cdot \log^2(m/\delta)),
    \end{align*}
    where we used $\kappa(W_g) \leq 2\kappa$ from Step 1 and $\kappa/\epsilon > 10$.

    {\bf Step 3: from the sketched kernel to $G$.} Since $G$ is positive definite, it has a positive definite square root $G^{1/2}$, with $\|G^{1/2}\| = \lambda_{\max}(G)^{1/2} = \sigma_{\max}(Z)$ and $\|G^{-1/2}\| = \lambda_{\min}(G)^{-1/2} = \sigma_{\min}(Z)^{-1}$. Define the symmetric matrix
    \begin{align*}
        E := G^{-1/2} W_g^\top W_g G^{-1/2} - I.
    \end{align*}
    Multiplying Eq.~\eqref{eq:W_g_spectral} by $G^{-1/2}$ on both sides gives $(1-\epsilon') \cdot I \preceq I + E \preceq (1+\epsilon') \cdot I$. Hence $\|E\| \leq \epsilon'$, and every eigenvalue of $I + E$ is at least $1 - \epsilon' > 0$, so $I + E$ is invertible with $\|(I+E)^{-1}\| \leq (1-\epsilon')^{-1}$. By the definition of $E$, we have $W_g^\top W_g = G^{1/2} (I+E) G^{1/2}$, and hence $W_g^\top W_g - G = G^{1/2} E G^{1/2}$. Therefore
    \begin{align}\label{eq:residual_split}
        G\wh{x} - y = (W_g^\top W_g \wh{x} - y) - G^{1/2} E G^{1/2} \wh{x}.
    \end{align}
    We bound the second term. From $W_g^\top W_g \wh{x} = G^{1/2} (I+E) G^{1/2} \wh{x}$, we get
    \begin{align*}
        G^{1/2}\wh{x} = (I+E)^{-1} G^{-1/2} W_g^\top W_g \wh{x}.
    \end{align*}
    By Eq.~\eqref{eq:sketched_residual} and the triangle inequality, $\|W_g^\top W_g \wh{x}\|_2 \leq \|y\|_2 + \|W_g^\top W_g \wh{x} - y\|_2 \leq (1+\epsilon/2) \cdot \|y\|_2$. So
    \begin{align*}
        \|G^{1/2}\wh{x}\|_2 \leq \|(I+E)^{-1}\| \cdot \|G^{-1/2}\| \cdot \|W_g^\top W_g \wh{x}\|_2 \leq \frac{1+\epsilon/2}{(1-\epsilon') \cdot \sigma_{\min}(Z)} \cdot \|y\|_2.
    \end{align*}
    Then
    \begin{align*}
        \|G^{1/2} E G^{1/2} \wh{x}\|_2
        \leq & ~ \|G^{1/2}\| \cdot \|E\| \cdot \|G^{1/2}\wh{x}\|_2 \\
        \leq & ~ \sigma_{\max}(Z) \cdot \epsilon' \cdot \frac{1+\epsilon/2}{(1-\epsilon') \cdot \sigma_{\min}(Z)} \cdot \|y\|_2 \\
        = & ~ \epsilon' \kappa(Z) \cdot \frac{1+\epsilon/2}{1-\epsilon'} \cdot \|y\|_2 \\
        \leq & ~ 2 \epsilon' \kappa \cdot \|y\|_2 \\
        = & ~ \frac{\epsilon}{2} \cdot \|y\|_2,
    \end{align*}
    where the first step follows from the definition of the spectral norm, the second step follows from the bounds above, the third step follows from $\kappa(Z) = \sigma_{\max}(Z)/\sigma_{\min}(Z)$, the fourth step follows from $\kappa(Z) \leq \kappa$ and $(1+\epsilon/2)/(1-\epsilon') \leq 1.05/0.975 < 2$, and the last step follows from $\epsilon' = \epsilon/(4\kappa)$. Plugging Eq.~\eqref{eq:sketched_residual} and this bound into Eq.~\eqref{eq:residual_split}, we obtain
    \begin{align*}
        \|G\wh{x} - y\|_2 \leq \frac{\epsilon}{2} \cdot \|y\|_2 + \frac{\epsilon}{2} \cdot \|y\|_2 = \epsilon \|y\|_2.
    \end{align*}

    {\bf Step 4: running time, failure probability and sketch dimension.} Computing $\epsilon'$ and $q$ takes $O(1)$ time. With $r = 1$ and accuracy $\epsilon'$, we have $\log_2(4n\kappa^2/\epsilon') = \log_2(16n\kappa^3/\epsilon)$, and $n\kappa/\epsilon > 10$, so the $q$ of Theorem~\ref{thm:gaussian_kernel} satisfies $q = \Theta(\log(n\kappa/\epsilon))$, and in particular $q = O(n\kappa/\epsilon)$. By Theorem~\ref{thm:gaussian_kernel}, computing $W_g$ takes $O(mn + q \cdot nd\log(2ndq/(\epsilon'\delta)))$ time, and $\log(2ndq/(\epsilon'\delta)) = \log(8ndq\kappa/(\epsilon\delta)) = O(\log(nd\kappa/(\epsilon\delta)))$. Since $q = O(\log(n\kappa/\epsilon))$, this time is
    \begin{align*}
        O(mn + nd\log(n\kappa/\epsilon)\log(nd\kappa/(\epsilon\delta))).
    \end{align*}
    Adding the time of the solver from Step 2 gives the stated time. The sketching event fails with probability at most $\delta/2$, and on the sketching event, the solver fails to stop within its time with probability at most $\delta/2$. By a union bound, with probability at least $1-\delta$, both events hold, and then $\wh{x}$ is computed within the stated time and satisfies $\|G\wh{x} - y\|_2 \leq \epsilon\|y\|_2$ by Step 3.

    Finally, by Theorem~\ref{thm:gaussian_kernel} with accuracy $\epsilon'$ and failure probability $\delta/2$, we have
    \begin{align*}
        m = \Theta(\epsilon'^{-2} n q^3 \log^2(2ndq/(\epsilon'\delta)) \log(2nq/\delta)).
    \end{align*}
    Here $\epsilon'^{-2} = 16\kappa^2/\epsilon^2$ and $q^3 = O(\log^3(n\kappa/\epsilon))$. Since $q = O(n\kappa/\epsilon)$, both $\log(2ndq/(\epsilon'\delta))$ and $\log(2nq/\delta)$ are $O(\log(nd\kappa/(\epsilon\delta)))$. Hence $m = O(\epsilon^{-2}\kappa^2 n \log^3(n\kappa/\epsilon) \log^3(nd\kappa/(\epsilon\delta)))$.
\end{proof}

\begin{remark}
The factor $\kappa^2$ in $m$ comes from Step 3, and Step 3 cannot be improved using only Eq.~\eqref{eq:W_g_spectral}. For $n \geq 2$, let $u_{\max}, u_{\min}$ be orthogonal unit eigenvectors of $G$ for $\lambda_{\max}(G)$ and $\lambda_{\min}(G)$, let $E := \epsilon'(u_{\max}u_{\min}^\top + u_{\min}u_{\max}^\top)$, so $\|E\| = \epsilon'$, and let $W_g^\top W_g := G^{1/2}(I+E)G^{1/2}$, which satisfies Eq.~\eqref{eq:W_g_spectral}. For $y := u_{\min}$, the exact solution $\wh{x}$ of $W_g^\top W_g \wh{x} = y$ has $\|G\wh{x} - y\|_2 = \frac{\epsilon'}{1-\epsilon'^2}\sqrt{\kappa(Z)^2+\epsilon'^2} \geq \epsilon' \kappa(Z) \|y\|_2$.
\end{remark}

\subsection{Sketch Dimension Below \texorpdfstring{$n$}{n}}\label{sec:kernel_below_n}

The algorithm of Theorem~\ref{thm:formal_exp} uses a sketch with $m \geq n$ rows. For positive definite $G$, this cannot be avoided by a sketch $W$ with $(1-\epsilon)G \preceq W^\top W$ (Remark~\ref{rem:kernel_rank_n}). Moreover, an algorithm whose output lies in a subspace of dimension $m \leq n/2$ chosen before $b$ is seen cannot guarantee $\|Gx'-b\|_2 \leq \epsilon\|b\|_2$ for every $b$ (Remark~\ref{rem:below_n_needed}). In this subsection, we sketch $G$ plus a ridge term $\lambda I_n$ instead. The resulting sketch can have far fewer than $n$ rows, and the guarantee becomes relative to $\OPT_K(b)$, the best residual among solutions of norm at most $(K/n)\|b\|_2$. Throughout this subsection, $K \geq 1$ is a scalar parameter given as input, and the kernel matrix is denoted by $G$.

We need the following tools from~\cite{akk+20}.

\begin{definition}[Statistical dimension, Definition 1 in~\cite{akk+20}]\label{def:stat_dim}
For $\nu>0$ and a PSD matrix $P\in\R^{n\times n}$, the $\nu$-statistical dimension of $P$ is $s_\nu(P):=\tr(P(P+\nu I_n)^{-1})$, and $s_0(P):=\rank(P)$.
\end{definition}
If $\nu>0$ and $P$ has eigenvalues $t_1,\ldots,t_n\geq 0$, then $s_\nu(P)=\sum_i t_i/(t_i+\nu)$. Each term is $0$ when $t_i=0$, at most $1$, and at most $t_i/\nu$. Hence
\begin{align}\label{eq:sdim_bounds}
s_\nu(P)\leq\min\{\rank(P),\ \tr(P)/\nu\}.
\end{align}

\begin{definition}[Oblivious subspace embedding, Definition 2 in~\cite{akk+20}]\label{def:ose_ridge}
Given $\epsilon,\delta,\mu>0$ and integers $D,n\geq 1$, an $(\epsilon,\delta,\mu,D,n)$-oblivious subspace embedding is a distribution over matrices $\Pi\in\R^{m\times D}$ such that for every $\lambda'\geq 0$ and every $C\in\R^{D\times n}$ with $s_{\lambda'}(C^\top C)\leq\mu$,
\begin{align*}
\Pr_{\Pi}\big[(1-\epsilon)(C^\top C+\lambda' I_n)\preceq(\Pi C)^\top\Pi C+\lambda' I_n\preceq(1+\epsilon)(C^\top C+\lambda' I_n)\big]\geq 1-\delta.
\end{align*}
\end{definition}

\begin{theorem}[Theorem 3 in~\cite{akk+20}]\label{thm:akk20}
For all positive integers $p,d,n$ and all $\epsilon,\mu>0$, there is a distribution over linear sketches $\Pi^p\in\R^{m\times d^p}$ which is an $(\epsilon,1/\poly(n),\mu,d^p,n)$-oblivious subspace embedding (Definition~\ref{def:ose_ridge}), provided that $m=\wt\Omega(p^4\mu/\epsilon^2)$. Moreover, for every $X\in\R^{d\times n}$, the matrix $\Pi^pX^{\otimes p}$ can be computed in time $\wt O(pnm+p^5\epsilon^{-2}\nnz(X))$. Here $\wt O$ and $\wt\Omega$ hide $\poly\log(nd/\epsilon)$ factors.
\end{theorem}

The sketch $W$ of Algorithm~\ref{alg:ridge_sketch} follows the construction in the proof of Theorem~5 of~\cite{akk+20} for the Gaussian kernel, which truncates the series $\exp(X^\top X)=\sum_{l\geq 0}(X^{\otimes l})^\top X^{\otimes l}/l!$, sketches each degree with Theorem~\ref{thm:akk20} and stacks the degrees with weights $1/\sqrt{l!}$. We apply it to $\exp(AA^\top)$ without their diagonal scaling, keep the degree-$0$ term exactly, and redo the recombination with the ridge $l!\,\lambda/(q+1)$ for degree $l$, so that, after the weights $1/l!$, the ridges add up to exactly $\lambda$, and the statistical dimension of degree $l$ is at most $\min\{\beta,4(q+1)K/(\epsilon\cdot l!)\}$ (Step 3 below). What is new here is this per-degree bound and the analysis of the ridge solution against $\OPT_K(b)$ for the unregularized objective (Step 5). Our result is the following. Algorithm~\ref{alg:ridge_sketch} is stated in Section~\ref{sec:main_result}.

\begin{theorem}[Attention kernel regression with sketch dimension below $n$]\label{thm:below_n}
Let $A\in\R^{n\times d}$ have rows $a_1,\ldots,a_n$ with $\|a_i\|_2\leq 1$ for all $i\in[n]$, let $G:=\exp(AA^\top)$ be the attention kernel matrix (Definition~\ref{def:attention_kernel}), and let $b\in\R^n$. Let $\epsilon,\delta\in(0,0.1)$, and let $K\geq 1$ be given as input. Let $q=\Theta(\log(K/\epsilon))$ be as in Algorithm~\ref{alg:ridge_sketch}, and let $\beta\in[1,n]$ satisfy
\begin{align*}
\beta\geq\max_{1\leq l\leq q}\rank(A^{\otimes l}),
\end{align*}
where $A^{\otimes l}\in\R^{n\times d^l}$ is the matrix whose $i$-th row is $a_i^{\otimes l}$. The choice $\beta=n$ is always allowed. Define
\begin{align*}
\OPT_K(b):=\inf\Big\{\|Gx-b\|_2:\ x\in\R^n,\ \|x\|_2\leq \frac{K}{n}\|b\|_2\Big\}.
\end{align*}
Then Algorithm~\ref{alg:ridge_sketch} uses a sketch $W\in\R^{m\times n}$ with
\begin{align*}
m=\wt O\big(\min\{\epsilon^{-2}K\beta,\ \epsilon^{-3}K^2\}\big)
\end{align*}
rows, where $\wt O$ hides $\poly\log(ndK/\epsilon)$ factors. With probability at least $1-\delta-q/\poly(n)$, where $1/\poly(n)$ is the failure probability of Theorem~\ref{thm:akk20}, it stops within time
\begin{align*}
O\big((nm+\min\{m,n\}^3)\cdot\log(K/\epsilon)\cdot\log^2((n+m)/\delta)\big)+\wt O\big(nm+\epsilon^{-2}K\cdot\nnz(A)\big)
\end{align*}
and outputs $x'\in\R^n$ such that
\begin{align*}
\|Gx'-b\|_2\leq \epsilon\|b\|_2+(1+\sqrt{\epsilon})\cdot\OPT_K(b).
\end{align*}
In particular, the following hold.
\begin{itemize}
\item[(i)] If $b=Gx^*$ for some $x^*\in\R^n$ with $\|x^*\|_2\leq (K/n)\|b\|_2$, then $\|Gx'-b\|_2\leq\epsilon\|b\|_2$. If $G$ is positive definite, this holds whenever $K\geq \lambda_{\max}(G)\cdot\|G^{-1}b\|_2/\|b\|_2$.
\item[(ii)] Let $U$ be the span of the eigenvectors of $G$ with eigenvalue at least $n/K$. Then $\OPT_K(b)\leq \dist(b,U)$ and $\dim U\leq\min\{eK,\,1+q\beta\}$. Hence $\|Gx'-b\|_2\leq\epsilon\|b\|_2+(1+\sqrt\epsilon)\cdot\dist(b,U)$, and $\|Gx'-b\|_2\leq\epsilon\|b\|_2$ for every $b\in U$.
\item[(iii)] If $G$ is positive definite and $K\geq n/\lambda_{\min}(G)$, then $\OPT_K(b)=0$ for every $b\in\R^n$. One may take $K=\kappa^2$ for any $\kappa\geq\kappa(Z)$, where $G=Z^\top Z$, and then $m=\wt O(\epsilon^{-2}\kappa^2 n)$, the order of Theorem~\ref{thm:formal_exp}. Since $\lambda_{\min}(G)\leq e$, this regime needs $K\geq n/e$, and in it $m\geq n$ whenever $q/\poly(n)<1$.
\item[(iv)] Let $U_\epsilon$ be the span of the eigenvectors of $G$ with eigenvalue at least $\epsilon n/K$. Then $\dim U_\epsilon\leq\min\{eK/\epsilon,\,1+q\beta\}$ and $\OPT_K(b)\geq\dist(b,U_\epsilon)-\epsilon\|b\|_2$ for every $b\in\R^n$.
\end{itemize}
So $m<n$ whenever $n$ exceeds $\min\{\epsilon^{-2}K\beta,\ \epsilon^{-3}K^2\}$ by a large enough $\poly\log(ndK/\epsilon)$ factor.
\end{theorem}

\begin{proof}[Proof of Theorem~\ref{thm:below_n}]
{\bf Notation and constants.} Let $X:=A^\top$, so the $i$-th column of $X$ is $x_i=a_i$ and $\|x_i\|_2\leq 1$. For $l\geq 0$, let $M_l:=(X^{\otimes l})^\top X^{\otimes l}\in\R^{n\times n}$. Then $M_l\succeq 0$, $(M_l)_{i,j}=\langle x_i,x_j\rangle^l$, $M_0={\bf 1}_n{\bf 1}_n^\top$, and $\rank(M_l)=\rank(X^{\otimes l})=\rank(A^{\otimes l})$, since $A^{\otimes l}=(X^{\otimes l})^\top$. As in the proof of Theorem~\ref{thm:gaussian_kernel}, $G=\sum_{l\geq 0}M_l/l!$. Let
\begin{align*}
Q:=\sum_{l=0}^q\frac{M_l}{l!},\quad T:=G-Q,\quad G_\lambda:=G+\lambda I_n,\quad H_\lambda:=W^\top W+\lambda I_n,\quad \epsilon_0:=2\epsilon_s=\frac{\epsilon}{8\sqrt{eK}}.
\end{align*}
Since $K\geq 1$ and $\epsilon<0.1$, we have $\epsilon_s<0.004$, $\epsilon_0<0.008$ and $K/\epsilon>10$. By the construction of $W$, $W^\top W=\sum_{l=0}^q Z_l^\top Z_l/l!$ and $Z_0^\top Z_0=M_0$.

{\bf Step 1: the size of $G$.} The matrix $G$ is PSD as a sum of PSD terms. We have $\tr(M_l)=\sum_i\|x_i\|_2^{2l}\leq n$, so $\tr(G)=\sum_i\exp(\|x_i\|_2^2)\leq en$ and $\|G\|\leq\tr(G)\leq en$. Since $G\succeq M_0={\bf 1}_n{\bf 1}_n^\top$, we have $\lambda_{\max}(G)\geq n$. If $G$ is positive definite, then $\lambda_{\min}(G)\leq G_{1,1}=\exp(\|x_1\|_2^2)\leq e$.

{\bf Step 2: truncation.} The matrix $T=\sum_{l>q}M_l/l!$ is PSD, so $Q\preceq G$. Each $M_l$ satisfies $\|M_l\|\leq\tr(M_l)\leq n$. For $l>q$, we have $1/l!\leq (q+2)^{-(l-q-1)}/(q+1)!\leq 2^{-(l-q-1)}/(q+1)!$, so $\|T\|\leq\sum_{l>q}n/l!\leq 2n/(q+1)!$. By the choice of $q$, $2^{q+1}\geq 212K^{3/2}/\epsilon^2>21200$, so $q+1\geq 15$ and $(q+1)!\geq 2^{q+1}$. Therefore
\begin{align*}
\|T\|\leq\frac{2n\epsilon^2}{212K^{3/2}}=\frac{n\epsilon^2}{106K^{3/2}}\leq\frac{n\epsilon^2}{64\sqrt{e}\,K^{3/2}}=\epsilon_s\lambda,
\end{align*}
where we used $64\sqrt{e}<105.6$ and $\epsilon_s\lambda=\frac{\epsilon}{16\sqrt{eK}}\cdot\frac{\epsilon n}{4K}$.

{\bf Step 3: the sketch event.} Let $\mathcal{E}_Q$ be the event
\begin{align}\label{eq:EQ}
(1-\epsilon_s)(Q+\lambda I_n)\preceq W^\top W+\lambda I_n\preceq(1+\epsilon_s)(Q+\lambda I_n).
\end{align}
We show $\Pr[\mathcal{E}_Q]\geq 1-q/\poly(n)$. Fix $l\in[q]$ and let $\lambda_l:=l!\,\lambda/(q+1)>0$. By Eq.~\eqref{eq:sdim_bounds}, $s_{\lambda_l}(M_l)\leq\rank(M_l)\leq\beta$, and
\begin{align*}
s_{\lambda_l}(M_l)\leq\frac{\tr(M_l)}{\lambda_l}\leq\frac{n(q+1)}{l!\,\lambda}=\frac{4(q+1)K}{\epsilon\cdot l!}.
\end{align*}
Hence $s_{\lambda_l}(M_l)\leq\mu_l$. The matrix $X^{\otimes l}$ is fixed by the input, and $\Pi^l$ is drawn, independently of the other degrees, from an oblivious subspace embedding with parameters $(\epsilon_s,1/\poly(n),\mu_l,d^l,n)$ (Theorem~\ref{thm:akk20}). Applying Definition~\ref{def:ose_ridge} with $\lambda'=\lambda_l$ and $C=X^{\otimes l}$, we get, with probability at least $1-1/\poly(n)$,
\begin{align*}
(1-\epsilon_s)(M_l+\lambda_l I_n)\preceq Z_l^\top Z_l+\lambda_l I_n\preceq(1+\epsilon_s)(M_l+\lambda_l I_n).
\end{align*}
For $l=0$, with $\lambda_0:=\lambda/(q+1)$, the same holds with probability $1$, since $Z_0^\top Z_0=M_0$. Divide the $l$-th relation by $l!$. Since $\lambda_l/l!=\lambda/(q+1)$ for every $0\leq l\leq q$, summing over $l=0,\ldots,q$ gives Eq.~\eqref{eq:EQ}, because $\sum_l Z_l^\top Z_l/l!=W^\top W$, $\sum_l M_l/l!=Q$ and the ridge terms add up to exactly $\lambda$. A union bound over the $q$ random degrees gives $\Pr[\mathcal{E}_Q]\geq 1-q/\poly(n)$.

{\bf Step 4: ridge approximation of $G$.} Assume $\mathcal{E}_Q$. We show
\begin{align}\label{eq:ridge_approx}
(1-\epsilon_0)G_\lambda\preceq H_\lambda\preceq(1+\epsilon_0)G_\lambda \quad\text{and}\quad \|W\|^2\leq 1.02\,en.
\end{align}
For the upper bound, $H_\lambda\preceq(1+\epsilon_s)(Q+\lambda I_n)\preceq(1+\epsilon_s)G_\lambda$, by $Q\preceq G$ (Step 2). For the lower bound, Step 2 gives $Q+\lambda I_n=G-T+\lambda I_n\succeq G+(1-\epsilon_s)\lambda I_n\succeq(1-\epsilon_s)G_\lambda$, so $H_\lambda\succeq(1-\epsilon_s)^2G_\lambda\succeq(1-2\epsilon_s)G_\lambda$. Finally, $W^\top W=H_\lambda-\lambda I_n\preceq(1+\epsilon_s)G+\epsilon_s\lambda I_n$. With Step 1, this gives $\|W\|^2\leq 1.004\,en+0.004\cdot\epsilon n/(4K)\leq 1.02\,en$.

{\bf Step 5: the exact sketched solution.} Assume $\mathcal{E}_Q$ and let $\wh x:=H_\lambda^{-1}b$. We show that for every $\bar x\in\R^n$ with $\|\bar x\|_2\leq(K/n)\|b\|_2$,
\begin{align}\label{eq:exact_sketched}
\|G\wh x-b\|_2\leq(1+0.28\sqrt{\epsilon})\cdot\|G\bar x-b\|_2+0.38\,\epsilon\|b\|_2.
\end{align}
Let $E:=G_\lambda^{-1/2}H_\lambda G_\lambda^{-1/2}-I_n$. Multiplying Eq.~\eqref{eq:ridge_approx} by $G_\lambda^{-1/2}$ on both sides gives $-\epsilon_0 I_n\preceq E\preceq\epsilon_0 I_n$, and $H_\lambda^{-1}=G_\lambda^{-1/2}(I_n+E)^{-1}G_\lambda^{-1/2}$. Let $x_\lambda:=G_\lambda^{-1}b$. Then $Gx_\lambda-b=(G-G_\lambda)G_\lambda^{-1}b=-\lambda G_\lambda^{-1}b$, and
\begin{align*}
G\wh x-b=(Gx_\lambda-b)+G(\wh x-x_\lambda)=-\lambda G_\lambda^{-1}b+GG_\lambda^{-1/2}\big[(I_n+E)^{-1}-I_n\big]G_\lambda^{-1/2}b.
\end{align*}
The eigenvalues of $(I_n+E)^{-1}-I_n$ are $-f/(1+f)$ for the eigenvalues $f\in[-\epsilon_0,\epsilon_0]$ of $E$, so its norm is at most $\epsilon_0/(1-\epsilon_0)\leq 1.02\,\epsilon_0$. The matrices $G$ and $G_\lambda$ commute, and $GG_\lambda^{-1/2}$ has eigenvalues $t/\sqrt{t+\lambda}\leq\sqrt t$ over the eigenvalues $t$ of $G$, so $\|GG_\lambda^{-1/2}\|\leq\|G\|^{1/2}\leq\sqrt{en}$. Hence
\begin{align}\label{eq:two_terms}
\|G\wh x-b\|_2\leq\lambda\|G_\lambda^{-1}b\|_2+1.02\,\epsilon_0\sqrt{en}\,\|G_\lambda^{-1/2}b\|_2.
\end{align}
Now fix $\bar x$ with $\|\bar x\|_2\leq(K/n)\|b\|_2$, and let $\rho:=G\bar x-b$, so $b=G\bar x-\rho$.

First term. We have $\lambda G_\lambda^{-1}b=\lambda G_\lambda^{-1}G\bar x-\lambda G_\lambda^{-1}\rho$, with $\|G_\lambda^{-1}G\|\leq 1$ and $\|\lambda G_\lambda^{-1}\|\leq 1$. So
\begin{align*}
\lambda\|G_\lambda^{-1}b\|_2\leq\lambda\|\bar x\|_2+\|\rho\|_2\leq\frac{\lambda K}{n}\|b\|_2+\|\rho\|_2=\frac{\epsilon}{4}\|b\|_2+\|\rho\|_2.
\end{align*}
Second term. We have $\|G_\lambda^{-1/2}b\|_2\leq\|G_\lambda^{-1/2}G\bar x\|_2+\|G_\lambda^{-1/2}\rho\|_2$. Since $G_\lambda^{-1/2}G=(G_\lambda^{-1/2}G^{1/2})G^{1/2}$ and $\|G_\lambda^{-1/2}G^{1/2}\|\leq 1$, we get $\|G_\lambda^{-1/2}G\bar x\|_2\leq(\bar x^\top G\bar x)^{1/2}$. Moreover, $\bar x^\top G\bar x=\bar x^\top(b+\rho)\leq\frac{K}{n}\|b\|_2(\|b\|_2+\|\rho\|_2)\leq\frac{K}{n}(\|b\|_2+\|\rho\|_2/2)^2$, using $a(a+c)\leq(a+c/2)^2$. Also $\|G_\lambda^{-1/2}\rho\|_2\leq\lambda^{-1/2}\|\rho\|_2=2\sqrt{K/(\epsilon n)}\,\|\rho\|_2$. Using $\epsilon_0\sqrt{eK}=\epsilon/8$, we get
\begin{align*}
1.02\,\epsilon_0\sqrt{en}\,\|G_\lambda^{-1/2}b\|_2
\leq & ~ 1.02\,\epsilon_0\sqrt{eK}\Big(\|b\|_2+\frac{\|\rho\|_2}{2}+\frac{2\|\rho\|_2}{\sqrt{\epsilon}}\Big) \\
= & ~ 0.1275\,\epsilon\|b\|_2+\big(0.06375\,\epsilon+0.255\sqrt{\epsilon}\big)\|\rho\|_2 \\
\leq & ~ 0.1275\,\epsilon\|b\|_2+0.28\sqrt{\epsilon}\,\|\rho\|_2,
\end{align*}
where the last step uses $\epsilon\leq\sqrt{0.1}\sqrt{\epsilon}$, so $0.06375\,\epsilon\leq 0.021\sqrt\epsilon$. Plugging both terms into Eq.~\eqref{eq:two_terms} gives Eq.~\eqref{eq:exact_sketched}, since $0.25+0.1275\leq 0.38$.

{\bf Step 6: the solver.} Assume $\mathcal{E}_Q$, and fix $W$. We show that over the solver's own fresh randomness, with probability at least $1-\delta$, it stops within time $O((nm+\min\{m,n\}^3)\cdot\log(K/\epsilon)\cdot\log^2((n+m)/\delta))$, and that whenever it stops, $\|G(x'-\wh x)\|_2\leq 0.25\,\epsilon\|b\|_2$.

{\em Case $m<n$.} The matrix $C=W^\top\oplus\sqrt\lambda I_m$ has full column rank because of its $\sqrt\lambda I_m$ block, and $C^\top C=WW^\top+\lambda I_m$ has eigenvalues in $[\lambda,\|W\|^2+\lambda]$. By Eq.~\eqref{eq:ridge_approx}, $\kappa(C)^2\leq 1+\|W\|^2/\lambda\leq 1+4.08\,eK/\epsilon\leq 12K/\epsilon$, using $K/\epsilon>10$. We apply Lemma~\ref{lem:regression_solver} with $C\in\R^{(n+m)\times m}$ in place of $A$, $b_2=u$, $\epsilon_2=\eta<0.1$ and $\delta_2=\delta$. With probability at least $1-\delta$, it stops within time $O(((n+m)m+m^3)\cdot\log(\kappa(C)/\eta)\cdot\log^2((n+m)/\delta))$, and $\log(\kappa(C)/\eta)\leq\log(\sqrt{12K/\epsilon}\cdot 500K^2/\epsilon^3)=O(\log(K/\epsilon))$. Whenever it stops, $r:=(WW^\top+\lambda I_m)\wt v-u$ satisfies $\|r\|_2\leq\eta\|u\|_2\leq\eta\|W\|\,\|b\|_2$. By the push-through identity $(W^\top W+\lambda I_n)W^\top=W^\top(WW^\top+\lambda I_m)$, and since $x'=(b-W^\top\wt v)/\lambda$,
\begin{align*}
H_\lambda x'-b
= & ~ \frac{1}{\lambda}\big(W^\top Wb+\lambda b-W^\top(WW^\top+\lambda I_m)\wt v\big)-b \\
= & ~ \frac{1}{\lambda}W^\top\big(u-(WW^\top+\lambda I_m)\wt v\big) \\
= & ~ -\frac{1}{\lambda}W^\top r.
\end{align*}
(With $r=0$ this is the Woodbury formula, and $x'=\wh x$.) Hence $\|H_\lambda x'-b\|_2\leq\eta\|W\|^2\|b\|_2/\lambda\leq(4.08\,eK\eta/\epsilon)\|b\|_2$.

{\em Case $m\geq n$.} The matrix $B=W\oplus\sqrt\lambda I_n\in\R^{(m+n)\times n}$ has full column rank, $B^\top B=H_\lambda$ and $\kappa(B)^2\leq 12K/\epsilon$. Lemma~\ref{lem:regression_solver} gives the stated time and, whenever the solver stops, $\|H_\lambda x'-b\|_2\leq\eta\|b\|_2\leq(4.08\,eK\eta/\epsilon)\|b\|_2$.

{\em Both cases.} We have $x'-\wh x=H_\lambda^{-1}(H_\lambda x'-b)$, and $H_\lambda\succeq\lambda I_n$ gives $\|GH_\lambda^{-1}\|\leq\|G\|/\lambda\leq en\cdot 4K/(\epsilon n)=4eK/\epsilon$. Therefore
\begin{align*}
\|G(x'-\wh x)\|_2\leq\frac{4eK}{\epsilon}\cdot\frac{4.08\,eK\eta}{\epsilon}\|b\|_2=\frac{16.32\,e^2K^2\eta}{\epsilon^2}\|b\|_2\leq\frac{120.6\,\epsilon}{500}\|b\|_2\leq 0.25\,\epsilon\|b\|_2.
\end{align*}

{\bf Step 7: the guarantee and the failure probability.} The event $\mathcal{E}_Q$ fails with probability at most $q/\poly(n)$ (Step 3). For every $W$ in $\mathcal{E}_Q$, the solver fails to stop within its time bound with probability at most $\delta$ (Step 6). By a union bound, both succeed with probability at least $1-\delta-q/\poly(n)$. Then, by Steps 5 and 6, for every $\bar x$ with $\|\bar x\|_2\leq(K/n)\|b\|_2$,
\begin{align*}
\|Gx'-b\|_2\leq\|G\wh x-b\|_2+\|G(x'-\wh x)\|_2\leq(1+0.28\sqrt\epsilon)\|G\bar x-b\|_2+0.63\,\epsilon\|b\|_2.
\end{align*}
Taking the infimum over such $\bar x$ gives $\|Gx'-b\|_2\leq\epsilon\|b\|_2+(1+\sqrt\epsilon)\OPT_K(b)$.

{\bf Step 8: sketch dimension.} By Theorem~\ref{thm:akk20}, $m_l=\wt\Theta(l^4\mu_l/\epsilon_s^2)=\wt\Theta(l^4\mu_lK/\epsilon^2)$, since $\epsilon_s^{-2}=256eK/\epsilon^2$ and $\log(nd/\epsilon_s)=O(\log(ndK/\epsilon))$. On the one hand, $\mu_l\leq\beta$, so $m=1+\sum_{l=1}^q m_l=\wt O(q^5K\beta/\epsilon^2)$. On the other hand, $\mu_l\leq 1+4(q+1)K/(\epsilon\cdot l!)$, and $\sum_{l\geq 0}l^4/l!=15e$ by Dobinski's formula for the Bell number $B_4=15$. So $\sum_{l=1}^q l^4\mu_l\leq q^5+60e(q+1)K/\epsilon$, and $m=\wt O(q^5K/\epsilon^2+qK^2/\epsilon^3)$. Since $q=O(\log(K/\epsilon))$ and $K/\epsilon\geq 1$, we get $m=\wt O(\min\{\epsilon^{-2}K\beta,\ \epsilon^{-3}K^2\})$.

{\bf Step 9: running time.} Computing $\lambda,\epsilon_s,\eta,q$ and the $\mu_l$ takes $O(q)$ time. By Theorem~\ref{thm:akk20}, the time to compute $Z_l$ is $\wt O(lnm_l+l^5\epsilon_s^{-2}\nnz(X))$. Summed over $l\in[q]$, this is
\begin{align*}
\wt O(qnm+q^6K\nnz(A)/\epsilon^2)=\wt O(nm+\epsilon^{-2}K\cdot\nnz(A)).
\end{align*} Forming $u=Wb$, $W^\top\wt v$ and $x'$ takes $O(nm)$ time. Adding the solver time from Step 6 gives the stated time.

{\bf Step 10: consequences.}
(i) Take $\bar x=x^*$. Then $\OPT_K(b)=0$. If $G$ is positive definite and $K\geq\lambda_{\max}(G)\|G^{-1}b\|_2/\|b\|_2$, then $x^*:=G^{-1}b$ satisfies $\|x^*\|_2\leq K\|b\|_2/\lambda_{\max}(G)\leq(K/n)\|b\|_2$, by Step 1.

(ii) Let $P_U$ be the orthogonal projection onto $U$, and let $\bar x:=G^\dagger P_Ub$. Then $G\bar x=P_Ub$ and $\|\bar x\|_2\leq(K/n)\|P_Ub\|_2\leq(K/n)\|b\|_2$, since $G^\dagger$ acts on $U$ with eigenvalues at most $K/n$. Hence $\OPT_K(b)\leq\|P_Ub-b\|_2=\dist(b,U)$. By Step 1, the number of eigenvalues of $G$ that are at least $n/K$ is at most $\tr(G)/(n/K)\leq eK$. Moreover, $\rank(Q)\leq\sum_{l=0}^q\rank(M_l)\leq 1+q\beta=:k$. If $k<n$, Weyl's inequality and Step 2 give $\lambda_{k+1}(G)\leq\lambda_{k+1}(Q)+\|T\|\leq\epsilon_s\lambda=\epsilon_s\epsilon n/(4K)<n/K$, so at most $k$ eigenvalues of $G$ are at least $n/K$. Hence $\dim U\leq\min\{eK,1+q\beta\}$.

(iii) If $G$ is positive definite and $K\geq n/\lambda_{\min}(G)$, then for every $b$, $\bar x:=G^{-1}b$ satisfies $\|\bar x\|_2\leq\|b\|_2/\lambda_{\min}(G)\leq(K/n)\|b\|_2$ and $G\bar x-b=0$, so $\OPT_K(b)=0$. For $\kappa\geq\kappa(Z)$, Step 1 gives $n/\lambda_{\min}(G)\leq\lambda_{\max}(G)/\lambda_{\min}(G)=\kappa(Z)^2\leq\kappa^2$, so $K=\kappa^2$ is allowed, and Step 8 with $\beta\leq n$ gives $m=\wt O(\epsilon^{-2}\kappa^2n)$. Since $\lambda_{\min}(G)\leq e$ (Step 1), this regime needs $K\geq n/e$. In it, $W$ is injective on $\mathcal{E}_Q$. Indeed, suppose $Wx=0$ for a unit vector $x$. Then $x^\top H_\lambda x=\lambda$, and Eq.~\eqref{eq:ridge_approx} gives $\lambda\geq(1-\epsilon_0)(\lambda_{\min}(G)+\lambda)$, that is, $\lambda\geq(1-\epsilon_0)\lambda_{\min}(G)/\epsilon_0>100\,\lambda_{\min}(G)$. But $\lambda=\epsilon n/(4K)\leq\epsilon\lambda_{\min}(G)/4<\lambda_{\min}(G)$, a contradiction. Hence $m\geq n$ on $\mathcal{E}_Q$. Since $m=\sum_{l=0}^q m_l$ is fixed by the parameters and does not depend on the randomness, and $\Pr[\mathcal{E}_Q]\geq 1-q/\poly(n)>0$ when $q/\poly(n)<1$ (Step 3), this gives $m\geq n$ whenever $q/\poly(n)<1$.

(iv) The dimension bound is proved as in (ii), with $\epsilon n/K$ in place of $n/K$, using $\|T\|\leq\epsilon_s\lambda<\epsilon n/K$. Let $P_\perp$ be the orthogonal projection onto $U_\epsilon^\perp$. It commutes with $G$, and $\|GP_\perp\|<\epsilon n/K$. For every $x$ with $\|x\|_2\leq(K/n)\|b\|_2$, $\|Gx-b\|_2\geq\|GP_\perp x-P_\perp b\|_2\geq\|P_\perp b\|_2-\epsilon\|b\|_2=\dist(b,U_\epsilon)-\epsilon\|b\|_2$. Taking the infimum over such $x$ gives the bound on $\OPT_K(b)$.
\end{proof}

\begin{remark}[An obstruction for outputs in a fixed subspace of dimension below $n$]\label{rem:below_n_needed}
Consider a randomized algorithm whose output lies in a subspace $V$ of dimension at most $m$, where $V$ may depend on $A$ and on the algorithm's randomness but not on $b$. This is the case, for example, when the output has the form $S^\top z$ for a sketch $S\in\R^{m\times n}$ drawn before $b$ is seen. Let $m\leq n/2$, fix a realization of $V$, and let $b$ be uniform on the unit sphere. Every output $x'\in V$ has $\|Gx'-b\|_2\geq\dist(b,GV)$, and $\E[\dist(b,GV)^2]=1-\dim(GV)/n\geq 1/2$. Since $\dist(b,GV)^2\leq 1$, we get $\Pr[\dist(b,GV)^2\geq 1/4]\geq(1/2-1/4)/(1-1/4)=1/3$, so the residual is at least $\|b\|_2/2>\epsilon\|b\|_2$ with probability at least $1/3$. Averaging over the algorithm's randomness gives a fixed $b$ on which it fails with probability at least $1/3>\delta$. So no such algorithm guarantees $\|Gx'-b\|_2\leq\epsilon\|b\|_2$ for every $b$. This holds even for inputs with $\beta=1$: take $a_i=t_iu$ for a unit vector $u$ and distinct $t_i\in(0,1]$. Then every $A^{\otimes l}$ has rank $1$, and $G$ is positive definite because the rows of $A$ are distinct (see the proof of Theorem~\ref{thm:formal_exp}). Algorithm~\ref{alg:ridge_sketch} is not of this form, since its output $x'=(b-W^\top\wt v)/\lambda$ contains $b$ itself, but with $m<n$ it does not guarantee $\|Gx'-b\|_2\leq\epsilon\|b\|_2$ for every $b$ either. Indeed, let $m<n$ and $q/\poly(n)<1$. As in the proof of part (iii) of Theorem~\ref{thm:below_n}, some $W$ in $\mathcal{E}_Q$ has a unit vector in its kernel, which gives $\lambda_{\min}(G)\leq\epsilon_0\lambda/(1-\epsilon_0)<0.01\lambda$. Let $b$ be a unit eigenvector of $G$ for $\lambda_{\min}(G)$. Then $\lambda\|G_\lambda^{-1}b\|_2>0.99$, and the second term of the identity before Eq.~\eqref{eq:two_terms} has norm at most $1.02\,\epsilon_0\sqrt{en}\,\lambda^{-1/2}=0.255\sqrt{\epsilon}$, so with Step 6, $\|Gx'-b\|_2\geq 0.99-0.255\sqrt{\epsilon}-0.25\epsilon>0.88$ whenever the algorithm succeeds. This agrees with Theorem~\ref{thm:below_n}, since $\OPT_K(b)=1-\lambda_{\min}(G)K/n>0.99$. What makes the theorem possible is its weaker guarantee, not the presence of $b$ in the output: the best $x$ in the span of the eigenvectors of $G$ with eigenvalue at least $\epsilon n/K$, a subspace of dimension at most $eK/\epsilon$ fixed before $b$ is seen, has residual at most $\epsilon\|b\|_2+\OPT_K(b)$. Part (iii) of Theorem~\ref{thm:below_n} shows that when $G$ is positive definite and $K\geq n/\lambda_{\min}(G)$, which makes $\OPT_K(b)=0$ for every $b$, the sketch has $m\geq n$ rows whenever $q/\poly(n)<1$.
\end{remark}

\ifdefined\isarxiv
\bibliographystyle{alpha}
\bibliography{ref}
\else

\fi




\end{document}